\documentclass[twoside,11pt]{article}

\usepackage{jmlr2e}
\usepackage{graphics}
\usepackage{graphicx}
\usepackage{epsfig}
\usepackage{lineno}
\usepackage{graphicx,epsfig} 
\usepackage{hyperref}    
\usepackage{epstopdf}
\usepackage{ifpdf}
\usepackage{paralist}
\usepackage{amsmath,amssymb,bm} 
\usepackage{algorithmic}
\usepackage{fancyhdr}
\usepackage{proof}
\usepackage[ruled]{algorithm2e}
\usepackage{multirow}
\usepackage{paralist}

\jmlrheading{1}{2018}{1-100}{10/18}{10/18}{ol-survey-2018}{Hoi, Sahoo, Lu, Zhao}

\ShortHeadings{Online Learning: A Comprehensive Survey}{Hoi, Sahoo, Lu and Zhao}
\firstpageno{1}

\begin{document}

\title{Online Learning: A Comprehensive Survey}

\author{\name Steven C. H. Hoi \email chhoi@smu.edu.sg\\
       \addr School of Information Systems, Singapore Management University, Singapore\\
       \name  Doyen Sahoo \email doyens@smu.edu.sg \\
       \addr School of Information Systems, Singapore Management University, Singapore\\
        \name Jing Lu \email lvjing12@jd.com \\
       \addr JD.com\\
        \name Peilin Zhao \email masonzhao@tencent.com\\
       \addr Tencent AI Lab\\
   }

\editor{XYZ}

\maketitle

\begin{abstract}

{\it Online learning} represents a family of machine learning methods, where a learner attempts to tackle some predictive (or any type of decision-making) task by learning from a sequence of data instances one by one at each time. The goal of online learning is to maximize the accuracy/correctness for the sequence of predictions/decisions made by the online learner given the knowledge of correct answers to previous prediction/learning tasks and possibly additional information. This is in contrast to traditional {\it batch} or {\it offline} machine learning methods that are often designed to learn a model from the entire training data set at once. Online learning has become a promising technique for learning from continuous streams of data in many real-world applications.
This survey aims to provide a comprehensive survey of the online machine learning literature through a systematic review of basic ideas and key principles and a proper categorization of different algorithms and techniques. Generally speaking, according to the types of learning tasks and the forms of feedback information, the existing online learning works can be classified into three major categories: (i) {\it online supervised learning} where full feedback information is always available, (ii) {\it online learning with limited feedback}, and (iii) {\it online unsupervised learning} where no feedback is available. Due to space limitation, the survey will be mainly focused on the first category, but also briefly cover some basics of the other two categories. Finally, we also discuss some open issues and attempt to shed light on potential future research directions in this field.
\end{abstract}


\section{Introduction}

Machine learning plays a crucial role in modern data analytics and artificial intelligence (AI) applications. Traditional machine learning paradigms often work in a batch learning or offline learning fashion (especially for supervised learning), where a model is trained by some learning algorithm from an entire training data set at once, and then the model is deployed for inference without (or seldom) performing any update afterwards. Such learning methods suffer from expensive re-training cost when dealing with new training data, and thus are poorly scalable for real-world applications. In the era of big data, traditional batch learning paradigms become more and more restricted, especially when live data grows and evolves rapidly. Making machine learning scalable and practical especially for learning from continuous data streams has become an open grand challenge in machine learning and AI.

Unlike traditional machine learning, {\it online learning} is a subfield of machine learning and includes an important family of learning techniques which are devised to learn models incrementally from data in a sequential manner.
Online learning overcomes the drawbacks of traditional batch learning in that the model can be updated instantly and efficiently by an online learner when new training data arrives. Besides, online learning algorithms are often easy to understand, simple to implement, and often founded on solid theory with rigorous regret bounds. Along with urgent need of making machine learning practical for real big data analytics, online learning has attracted increasing interest in recent years.

This survey aims to give a comprehensive survey of {\it online learning} literature. Online learning \footnote{The term of ``online learning" in this survey is {\it not} related to ``e-learning" in the online education field.} has been extensively studied across different fields, ranging from machine learning, data mining, statistics, optimization and applied math, to artificial intelligence and data science. This survey aims to distill the core ideas of online learning methodologies and applications in literature. This survey is written mainly for machine learning audiences, and assumes readers with basic knowledge in machine learning. While trying our best to make the survey as comprehensive as possible, it is very difficult to cover every detail since online learning research has been evolving rapidly in recent years. We apologize in advance for any missing papers or inaccuracies in description, and encourage readers to provide feedback, comments or suggestions. Finally, as a supplemental document to this survey, readers may check our updated version online at: \url{http://libol.stevenhoi.org/survey}.

\subsection{What is Online Learning?}


Traditional machine learning paradigm often runs in a batch learning fashion\index{Batch learning}, e.g., a supervised learning task, where a collection of training data is given in advance to train a model by following some learning algorithm. Such a paradigm requires the entire training data set to be made available prior to the learning task, and the training process is often done in an offline environment due to the expensive training cost. Traditional batch learning methods suffer from some critical drawbacks: (i) low efficiency in both time and space costs; and (ii) poor scalability for large-scale applications because the model often has to be re-trained from scratch for new training data.


In contrast to batch learning algorithms, online learning \index{Online learning} is a method of machine learning for data arriving in a sequential order, where a learner aims to learn and update the best predictor for future data at every step. Online learning is able to overcome the drawbacks of batch learning in that the predictive model can be updated instantly for any new data instances. Thus, online learning algorithms are far more efficient and scalable for large-scale machine learning tasks in real-world data analytics applications where data are not only large in size, but also arriving at a high velocity.




\subsection{Tasks and Applications}

Similar to traditional (batch) machine learning methods, online learning techniques can be applied to solve a variety of tasks in a wide range of real-world application domains.
Examples of online learning tasks include the following:

{\it Supervised learning tasks:} Online learning algorithms can be derived for supervised learning tasks. One of the most common tasks is classification, aiming to predict the categories for a new data instance belongs to, on the basis of observing past training data instances whose category labels are given. For example, a commonly studied task in online learning is online binary classification (e.g., spam email filtering) which only involves two categories (``spam" vs ``benign" emails); other types of supervised classification tasks include multi-class classification, multi-label classification, and multiple-instance classification, etc.

In addition to classification tasks, another common supervised learning task is regression analysis, which refers to the learning process for estimating the relationships among variables (typically between a dependent variable and one or more independent variables). Online learning techniques are naturally applied for regression analysis tasks, e.g., time series analysis in financial markets where data instances naturally arrive in a sequential way. Besides, another application for online learning with financial time series data is online portfolio section where an online learner aims to find a good (e.g., profitable and low-risk) strategy for making a sequence of decisions for portfolio selection.

{\it Bandit learning tasks:} Bandit online learning algorithms, also known as Multi-armed bandits (MAB), have been extensively used for many online recommender systems, such as online advertising for internet monetization, product recommendation in e-commerce, movie recommendation for entertainment, and other personalized recommendation, etc.

{\it Unsupervised learning tasks:} Online learning algorithms can be applied for unsupervised learning tasks. Examples include clustering or cluster analysis --- a process of grouping objects such that objects in the same group (``cluster") are more similar to each other than to objects in other clusters. Online clustering aims to perform incremental cluster analysis on a sequence of instances, which is common for mining data streams.



{\it Other learning tasks:} Online learning can also be used for other kinds of machine learning tasks, such as learning for recommender systems, learning to rank, or reinforcement learning. For example, collaborative filtering with online learning can be applied to enhance the performance of recommender systems by learning to improve collaborative filtering tasks sequentially from continuous streams of ratings/feedback information from users.

Last but not least, we note that online learning techniques are often used in two major scenarios. One is to improve efficiency and scalability of existing machine learning methodologies for batch machine learning tasks where a full collection of training data must be made available before the learning tasks. For example, Support Vector Machines (SVM) is a well-known supervised learning method for batch classification tasks, in which classical SVM algorithms (e.g., QP or SMO solvers~\citep{platt1999fast}) suffer from poor scalability for very large-scale applications. In literature, various online learning algorithms have been explored for training SVM in an online (or stochastic) learning manner~\citep{poggio2001incremental,shalev2011pegasos}, making it more efficient and scalable than conventional batch SVMs. The other scenario is to apply online learning algorithms to directly tackle online streaming data analytics tasks where data instances naturally arrive in a sequential manner and the target concepts may be drifting or evolving over time. Examples include time series regression, such as stock price prediction, where data arrives periodically and the learner has to make decisions immediately before getting the next instance.

\begin{figure*}[ht!]
	\centering
	\includegraphics[width=\textwidth]{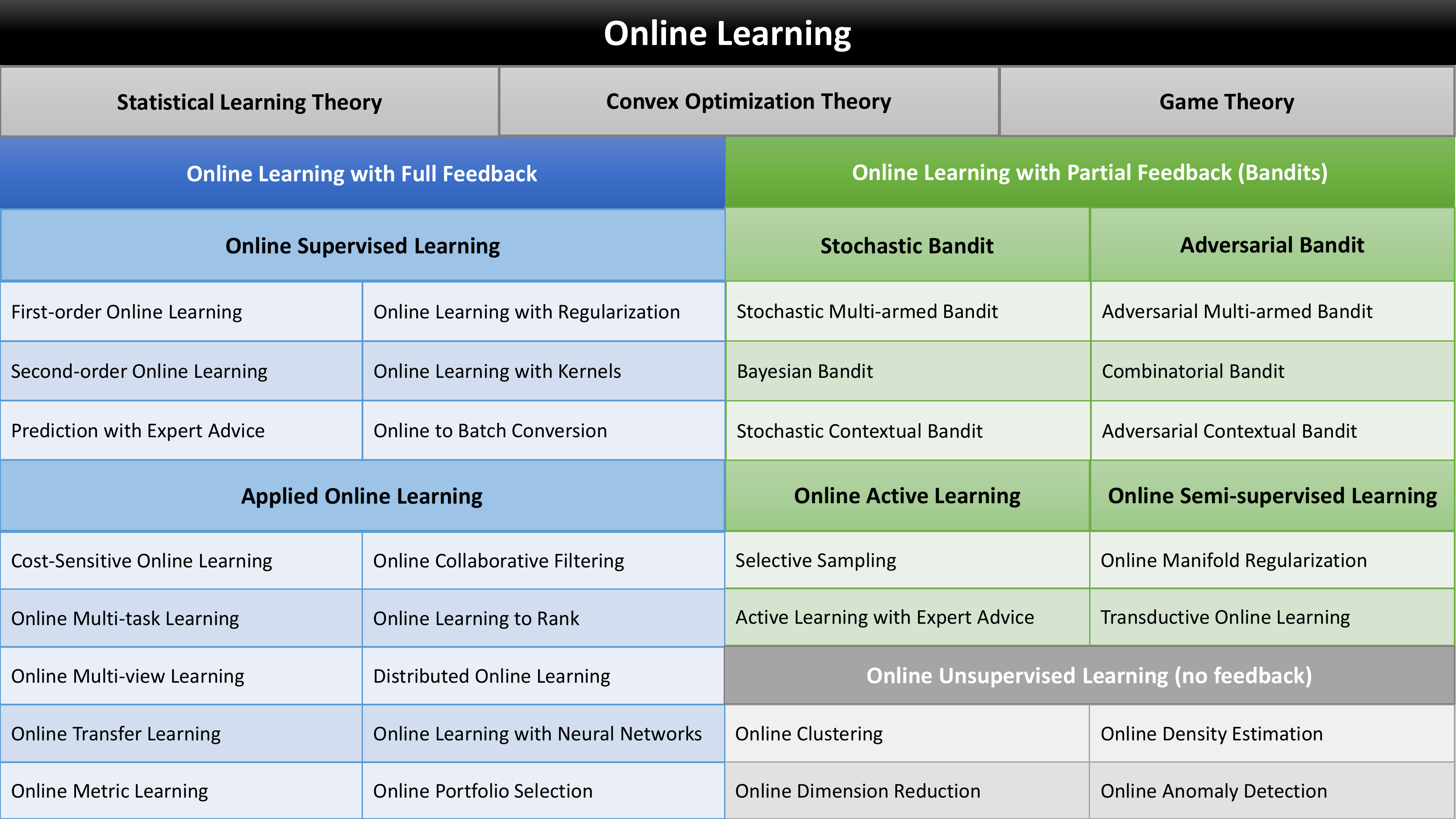}
	\caption{Taxonomy of Online Learning Techniques\index{Taxonomy}}\label{fig:ol-taxonomy}
\end{figure*}

\subsection{Taxonomy}

To help readers better understand the online learning literature as a whole, we attempt to construct a taxonomy of online learning methods and techniques, as summarized in Figure~\ref{fig:ol-taxonomy}. In general, from a theoretical perspective, online learning methodologies are founded based on theory and principles from three major theory communities: learning theory, optimization theory, and game theory. From the perspective of specific algorithms, we can further group the existing online learning techniques into different categories according to their specific learning principles and problem settings. Specifically, according to the types of feedback information and the types of supervision in the learning tasks, online learning techniques can be classified into the following three major categories:

\begin{itemize}
\item{\bf Online supervised learning:} This is concerned with supervised learning tasks where full feedback information is always revealed to a learner at the end of each online learning round. It can be further divided into two groups of studies: (i) ``Online Supervised Learning" which forms the fundamental approaches and principles for Online Supervised Learning; and (ii) ``Applied Online Learning" which constitute more non-traditional online supervised learning, where the fundamental approaches cannot be directly applied, and algorithms have been appropriately tailored to suit the non-traditional online learning setting. 


\item{\bf Online learning with limited feedback:} This is concerned with tasks where an online learner receives partial feedback information from the environment during the online learning process. For example, consider an online multi-class classification task, at a particular round, the learner makes a prediction of class label for an incoming instance, and then receives the partial feedback indicating whether the prediction is correct or not, instead of the particular true class label explicitly. For such tasks, the online learner often has to make the online updates or decisions by attempting to achieve some tradeoff between the exploitation of disclosed knowledge and the exploration of unknown information with the environment.

\item{\bf Online unsupervised learning:} This is concerned with online learning tasks where the online learner only receives the sequence of data instances without any additional feedback (e.g., true class label) during the online learning tasks. Unsupervised online learning can be considered as a natural extension of traditional unsupervised learning for dealing with data streams, which is typically studied in batch learning fashion. Examples of unsupervised online learning include online clustering, online dimension reduction, and online anomaly detection tasks, etc. Unsupervised online learning has less restricted assumptions about data without requiring explicit feedback or label information which could be difficult or expensive to acquire.
\end{itemize}




This article will conduct a systematic review of existing work for online learning, especially for online supervised learning and online learning with partial feedback. Finally, we note that it is always very challenging to make a precise categorization of all the existing online learning work, and it is likely that the above proposed taxonomy may not fully cover all the existing online learning work in literature, though we have tried our best to cover as much as possible.

\subsection{Related Work and Further Reading}

This paper attempts to make a comprehensive survey of online learning research work. In literature, there are some related books, PHD theses, and articles published over the past years dedicated to online learning~\citep{bianchi-2006-prediction,shalev2011online}, in which many of them also include rich discussions on related work on online learning. For example, the book titled ``Prediction, Learning, and Games"~\citep{bianchi-2006-prediction} gave a nice introduction about some niche subjects of online learning, particularly for online prediction with expert advice and online learning with partial feedback. Another work titled ``Online Learning and Online Convex Optimization"~\citep{shalev2011online} gave a nice tutorial about basics of online learning and foundations of online convex optimization. In addition, there are also quite a few PHD theses dedicated to addressing different subjects of online learning~\citep{kleinberg2005online,shalev2007online,okl2013zhao,binli2013}. Readers are also encouraged to read some older related books, surveys and tutorial notes about online learning and online algorithms~\citep{fiat1998online,bottou1998online,tr2008,blum1998line,albers2003online}. Finally, readers who are interested in applied online learning can explore some open-source toolboxes, including LIBOL \citep{hoi2014libol,wu2016sol} and Vowpal Wabbit \citep{langford2007vowpal}.



\section{Problem Formulations and Related Theory}
\def \c {\mathbf{c}}
\def \r {\mathbf{r}}
\def \u {\mathbf{u}}
\def \v {\mathbf{v}}
\def \w {\mathbf{w}}
\def \x {\mathbf{x}}
\def \hy {\widehat{y}}

\def \A {\mathcal{A}}
\def \E {\mathbb{E}}
\def \H {\mathcal{H}}
\def \I {\mathbb{I}}
\def \L {\mathcal{L}}
\def \N {\mathbb{N}}
\def \B {\mathcal{B}}
\def \R {\mathbb{R}}
\def \S {\mathcal{S}}

\def \Tr {\mathrm{Tr}}

\def \bqs {\begin{eqnarray*}}
\def \eqs {\end{eqnarray*}}

\def \sign {\mathrm{sign}}

\def \Reg {\mathrm{Reg}}
\def \bqs {\begin{eqnarray*}}
\def \eqs {\end{eqnarray*}}

\def \v {\mathbf{v}}
\def \w {\mathbf{w}}
\def \hw {\hat{\w}}
\def \R {\mathbb{R}}

\def \A {\mathcal{A}}
\def \Z {\mathcal{Z}}

Without loss of generality, we first give a formal formulation of a classic online learning problem, i.e., online binary classification, and then introduce basics of statistical learning theory, online convex optimization and game theory as the theoretical foundations for online learning techniques.

\subsection{Problem Settings}

Consider an online binary classification task, online learning takes place in a sequential way. On each round, a learner receives a data instance, and then makes a prediction of the instance, e.g., classifying it into some predefined categories. After making the prediction, the learner receives the true answer about the instance from the environment as a feedback. Based on the feedback, the learner can measure the loss suffered, depending on the difference between the prediction and the answer. Finally, the learner updates its prediction model by some strategy so as to improve predictive performance on future received instances.

Consider spam email detection as a running example of online binary classification, where the learner answers every question in binary: yes or no. The task is supervised binary classification from a machine learning perspective. More formally, we can formulate the problem as follows: consider a sequence of instances/objects represented in a vector space, $\x_t\in \R^d$, where $t$ denotes the $t$-th round and $d$ is the dimensionality, and we use $y_t \in \{+1, -1\}$ to denote the true class label of the instance. The online binary classification takes place sequentially. On the $t$-th round, an instance $\x_t$ is received by the learner, which then employs a binary classifier $\w_t$ to make a prediction on $\x_t$, e.g., $\hat{y}_t=\sign(\w_t^\top\x_t)$ that outputs $\hat{y}_t=+1$ if $\w_t^\top\x_t \geq 0$ and $\hat{y}_t=-1$ otherwise. After making the prediction, the learner receives the true class label $y_t$ and thus can measure the suffered loss, e.g., using the hinge-loss $\ell_t(\w_t)=\max\left(0,1-y_t\w_t^\top\x_t\right)$. Whenever the loss is nonzero, the learner updates the prediction model from $\w_t$ to $\w_{t+1}$ by applying some strategy on the training example $(\x_t, y_t)$. The procedure of Online Binary Classification is summarized in Algorithm~\ref{alg:online-binary-classificaition}.

\begin{algorithm}[hptb]
\caption{Online Binary Classification process. }\label{alg:online-binary-classificaition}
\begin{algorithmic}
\STATE Initialize the prediction function as $\w_1$;
\FOR{ $t=1,2,\ldots,T$}
\STATE Receive instance: $\x_t\in \R^d$;
\STATE Predict $\hy_t=\sign(\w_t^\top\x_t)$ as the label of $\x_t$;
\STATE Receive the true class label: $y_t\in\{-1,+1\}$;
\STATE Suffer loss: $\ell_t(\w_t)$ which is a convex loss function on both $\w_t^\top\x_t$ and  $y_t$;
\STATE Update the prediction model $\w_t$ to $\w_{t+1}$;
\ENDFOR
\end{algorithmic}
\end{algorithm}
By running online learning over a sequence of $T$ rounds, the number of mistakes made by the online learner can be measured as $M_T = \sum^T_{t=1}\I(\hy_t\not=y_t)$.
In general, the classic goal of an online learning task is to minimize the regret of the online learner's predictions against the best fixed model in hindsight, defined as
\begin{eqnarray}
R_T=\sum^T_{t=1}\ell_t(\w_t)-\min_\w\sum^T_{t=1}\ell_t(\w)
\end{eqnarray}
where the second term is the loss suffered by the optimal model $\w^*$ that can only be known in hindsight after seeing all the instances and their class labels.
From the theoretical perspective of regret minimization, if an online algorithm guarantees that its regret is sublinear as a function of $T$, i.e., $R_T=o(T)$, it implies that $\lim_{T\rightarrow\infty} R(T)/T = 0$ and thus on average the learner performs almost as well as the best fixed model in hindsight.

\subsection{Statistical Learning Theory}
Statistical learning theory, first introduced in the late 1960's, is one of key foundations for theoretical analysis of machine learning problems, especially for supervised learning.
In literature, there are many comprehensive survey articles and books \citep{vapnik1998statistical,vapnik1999overview}. In the following, we introduce some basic concept and framework.

\subsubsection{Empirical Error Minimization}
Assume instance $\x_t$ is generated randomly from a fixed but unknown distribution $P(\x)$ and its class label $y$ is also generated with a fixed but unknown distribution $P(y|\x)$. The joint distribution of labeled data is $P(\x,y)=P(\x)P(y|\x)$. The goal of a learning problem is to find a prediction function $f(\x)$ that minimizes the expected value of the loss function:
\bqs
R(f)=\int \ell(y,f(\x))\text d P(x,y)\\
\eqs
which is also termed as the {\it True Risk} function. The solution $f^*=\arg\min R(f)$ is the optimal predictor. In general, the true risk function $R(f)$ cannot be computed directly because of the unknown distribution $P(x,y)$.
In practice, we approximate the true risk by estimating the risk over a finite collection of instances $(\x_1,y_1),...,(\x_N,y_N)$ drawn i.i.d., which is called the ``Empirical Risk" or ``Empirical Error"
\bqs
R_{emp}(f)=\frac{1}{N}\sum_{n=1}^N \ell(y_n,f(\x_n))
\eqs
The problem of learning via the Empirical Error Minimization (ERM) is to find a hypothesis $f$ over a hypothesis space $\mathcal F$ by minimizing the Empirical Error:
\bqs
\hat f_n=\arg\min_{f\in\mathcal F}R_{emp}(f)
\eqs
ERM is the theoretical base for many machine learning algorithms. For example, in the problem of binary classification, when assuming $\mathcal F$ is the set of linear classifiers and the hinge loss is used, the ERM principle indicates that the best linear model $\w$ can be trained by minimizing the following objective
\bqs
R_{emp}(\w)=\frac{1}{N}\sum_{n=1}^N \max(0,1-y_n\w^\top\x_n)
\eqs

\subsubsection{Error Decomposition}
The difference between the optimal predictor $f^*$ and the empirical best predictor $\hat f_n$ can be measured by the Excess Risk, which can be decomposed as follows:
$$
R(\hat f_n)-R(f^*)=\left( R(\hat f_n)-\inf_{f\in\mathcal F}R(f)\right)+\left( \inf_{f\in\mathcal F}R(f)-R(f^*)\right)
$$
where the first term is called the Estimation Error due to the finite amount of training samples that may not be enough to represent the unknown distribution, and the second term is called the Approximation Error due to the restriction of model class $\mathcal F$ that may not be flexible enough to include the optimal predictor $f^*$. In general, the estimation error will be reduced when increasing the amount of training data, while the approximation error can be reduced by increasing the model complexity/capacity. However, the estimation error often increases when the model complexity grows, making it challenging for model selection.


\subsection{Convex Optimization Theory}
Many online learning problems can  essentially be (re-)formulated as an Online Convex Optimization (OCO) task. In the following, we introduce some basics of OCO.

An online convex optimization task typically consists of two major elements: a convex set $\mathcal{S}$ and a convex cost function $\ell_t(\cdot)$.
At each time step $t$, the online algorithm decides to choose a weight vector $\mathbf{w}_t\in\mathcal{S}$; after that, it suffers a loss $\ell_t(\mathbf{w}_t)$, which is computed based on a convex cost function $\ell_t(\cdot)$ defined over $\mathcal{S}$. The goal of the online algorithm is to choose a sequence of decisions $\mathbf{w}_1,\mathbf{w}_2,\ldots$ such that the regret in hindsight can be minimized.

More formally, an online algorithm aims to achieve a low regret $R_T$ after $T$ rounds, where the regret $R_T$ is defined as:
\begin{equation}
R_T=\sum_{t=1}^T{\ell_t(\mathbf{w}_t)} - \inf_{\mathbf{w}^*\in\mathcal{S}}{\sum_{t=1}^T{\ell_t(\mathbf{w}^*)}},
\end{equation}
where $\mathbf{w}^*$ is the solution that minimizes the convex objective function $\sum_{t=1}^T \ell_t(\mathbf{w})$ over $\mathcal{S}$.


For example, consider an online binary classification task for training online Support Vector Machines (SVM) from a sequence of labeled instances $(\mathbf{x}_t,y_t), t=1,\ldots, T$, where $\mathbf{x}_t\in\mathcal{R}^d$ and $y_t\times\{+1,-1\}$. One can define the loss function $\ell(\cdot)$ as $\ell_t(\mathbf{w}_t) = \max(0,1-y_t \mathbf{w}^{\top}\mathbf{x})$ and the convex set $S$ as $\{\forall \mathbf{w}\in\mathcal{R}^d | \|\mathbf{w}\|\leq C\}$ for some constant parameter $C$. There are a variety of algorithms to solve this problem.

For a comprehensive treatment of this subject, readers are referred to the books in~\citep{shalev2011online,hazan2016introduction}.
Below we briefly review three major families of online convex optimization (OCO) methods, including first-order algorithms, second-order algorithms, and regularization based approaches.

\subsubsection{First-order Methods}

First order methods aim to optimize the objective function using the first order (sub) gradient information. { Online Gradient Descent (OGD)}\citep{DBLP:conf/icml/Zinkevich03} can be viewed as an online version of Stochastic Gradient Descent (SGD) in convex optimization, and is one of the simplest and most popular methods for convex optimization.

At every iteration, based on the loss suffered on instance $\x_t$, the algorithm takes a step from the current model to update to a new model, in the direction of the gradient of the current loss function. This update gives us $\u = \w_t-\eta_t\nabla \ell_t(\w_t)$. The resulting update may push the model to lie outside the feasible domain. Thus, the algorithm projects the model onto the feasible domain, i.e., $\Pi_{\S}(\u)=\arg\min_{\w\in\S}\|\w-\u\|$ (where $\Pi_{\S}$ denotes the projection operation). OGD is simple and easy to implement, but the projection step sometimes may be computationally intensive which depends on specific tasks. In theory~\citep{DBLP:conf/icml/Zinkevich03}, OGD achieves sublinear regret $O(\sqrt{T})$ for an arbitrary sequence of $T$ convex cost functions (of bounded gradients), with respect to the best single decision in hindsight.

\subsubsection{Second-order Methods.}
Second-order methods aim to exploit second order information to speed up the convergence of the optimization. A popular approach is the Online Newton Step Algorithm. The Online Newton Step~\citep{DBLP:journals/ml/HazanAK07} can be viewed as an online analogue of the Newton-Raphson method in batch optimization. Like OGD, ONS also performs an update by subtracting a vector from the current model in each online iteration. While the vector subtracted by OGD is the gradient of the current loss function based on the current model, in ONS the subtracted vector is the inverse Hessian multiplied by the gradient, i.e., $A_t^{-1}\nabla \ell_t(\w_t)$ where $A_t$ is related to the Hessian. $A_t$ is also updated in each iteration as $A_t= A_{t-1}+\nabla \ell_t(\w_t)\nabla \ell_t(\w_t)^\top$. The updated model is projected back to the feasible domain as $\w_{t+1}=\Pi_{\S}^{A_t}(\w_t-\eta A_t^{-1}\nabla \ell_t(\w_t))$, where $\Pi^{A}_\S(\u)=\arg\min_{\w\in\S}(\w-\u)^\top A(\w-\u)$.
Different from OGD where the projection is made under the Euclidean norm, ONS projects under the norm induced by the matrix $A_t$. Although ONS's time complexity $O(n^2)$ is higher than OGD's $O(n)$, it guarantees a logarithmic regret $O(\log T)$ under relatively weaker assumptions of exp-concave cost functions.

\subsubsection{Regularization}

Unlike traditional convex optimization, the aim of OCO is to optimize the regret. Traditional approaches (termed as Follow the Leader (FTL)) can be unstable, leading to high regret (e.g. linear regret) in the worst case \citep{hazan2016introduction}. This motivates the need to stabilize the approaches through regularzation. Here we discuss the common regularization approaches.

\emph{Follow-the-Regularized-Leader (FTRL).} The idea of Follow-the-Regularized-Leader (FTRL)~\citep{abernethy2008competing,shalev2007primal} is to stablize the prediction of the Follow-the-Leader (FTL)~\citep{DBLP:journals/jcss/KalaiV05,hannan1957approximation} by adding a regularization term $R(\w)$ which is strongly convex, smooth and twice differentiable. The idea is to solve the following optimization problem in each iteration:
\bqs
	\w_{t+1}=\arg\min_{\w\in\S}\left[\eta \sum^t_{s=1}\nabla\ell_s(\w_s)^\top\w+R(\w)\right]
\eqs
where $\S$ is the feasible convex set and $\eta$ is the learning rate. In theory, the FTRL algorithm in general achieves a sublinear regret bound $O(\sqrt{T})$.

\emph{Online Mirror Descent (OMD)}.
OMD is an online version of the Mirror Descent (MD) method~\citep{citeulike:13330757,DBLP:conf/colt/DuchiSST10} in batch convex optimization. The OMD algorithm behaves like OGD, in that it updates the model using a simple gradient rule. However, it generalizes OGD as it performs updates in the dual space. This duality is induced by the choice of the regularizer: the gradient of the regularization serves as a mapping from $\R^d$ to itself. Due to this transformation by the regularizer, OMD is able to obtain better bounds in terms of the geometry of the space.

In general, OMD has two variants of algorithms: lazy OMD and active OMD. The lazy version keeps track of a point in Euclidean space and projects it onto the convex feasible domain only when making prediction, while the active version keeps a feasible model all the time, which is a direct generalization of OGD. Unlike OGD, the projection step in OMD is based on the Bregman Divergence $\B_R$, i.e.,  $\w_{t+1}=\arg\min_{\w\in\S}\B_R(\w\|\v_{t+1})$, where $\v_{t+1}$ is the updated model after the gradient step. In general, the lazy OMD has the same regret bound as FTRL. The active OMD also has a similar regret bound. When $R(\w)=\frac{1}{2}\|\w\|_2^2$, OMD recovers OGD. If we use other functions as $R$, we can also recover some other interesting algorithms, such as the Exponential Gradient (EG) algorithm below.

{\it Exponential Gradient (EG).}
Let $R(\w)=\w\ln \w$ be the negative entropy function and the feasible convex domain be the simplex $\S=\Delta_d=\{\w\in\R^d_+|\sum_i w_i=1\}$, then OMD will recover the Exponential Gradient (EG) algorithm~\citep{DBLP:conf/stoc/KivinenW95}. In this special case, the induced projection is the normalization by the $L1$ norm, which indicates
\bqs
w_{t+1,i}=\frac{w_{t,i}\exp[-\eta (\nabla \ell_t(\w_t))_i]}{\sum_j w_{t,j}\exp[-\eta (\nabla \ell_t(\w_t))_j]}
\eqs
As a special case of OMD, the regret of EG is bounded by $O(\sqrt{T})$.

\emph{Adaptive (Sub)-Gradient Methods.}
In the previous algorithms, the regularization function $R$ is always fixed and data independent, during the whole  learning process. Adaptive (Sub)-Gradient (AdaGrad) algorithm~\citep{DBLP:journals/jmlr/DuchiHS11} is an algorithm that can be considered as online mirror descent with adaptive regularization, i.e., the regularization function $R$ can change over time. The regularizer $R$ at the $t$-th step, is actually the function $R(\w)=\frac{1}{2}\|\w\|^2_{A_t^{1/2}}=\frac{1}{2}\w^\top A_t^{1/2} \w $, which is constructed from the (sub)-gradients received before (and including) the $t$-th step. In each iteration the model is updated as:
\bqs
	\w_{t+1}=\arg\min_{\w\in\S}\left\|\w-[\w_t-\eta A_t^{-\frac{1}{2}}\nabla \ell_t(\w_t)]\right\|_{A_t^{\frac{1}{2}}}^2
\eqs
where $A_t$ is updated as:
\bqs
	A_t= A_{t-1}+ \nabla \ell_t(\w_t) \nabla \ell_t(\w_t)^\top
\eqs

We also note that there are also other emerging online convex optimization methods, such as Online Convex Optimization with long term constraints \citep{jenatton2015adaptive}, which assumes that the constraints are only required to be satisfied in long term, and
Online ADMM~\citep{DBLP:conf/icml/WangB12} which is an online version for the Alternating Direction Method of Multipliers (ADMM)~\citep{gabay1976dual,boyd2011distributed} and is particularly suitable for distributed optimization applications. The RESCALEDEXP algorithm \citep{cutkosky2016online}, proposed recently, does not use any prior knowledge about the loss functions and does not require the tuning of learning rate.

\subsection{Game Theory}

Game theory is closely related to online learning. In general, an online prediction task can be formulated as a problem of learning to play a repeated game between a learner and an environment \citep{freund1999adaptive}. Consider online classification as an example, during each iteration, the algorithm chooses one class from a finite number of classes and the environment reveals the true class label. Assume the environment is stable (e.g., i.i.d), i.e., not played by an adversary. The algorithm aims to perform as well as the best fixed strategy. The classic online classification problem thus can be modeled by the game theory under the simplest assumption, full feedback and a stable environment. More generally, various settings in game theory can be related to many other types of online learning problems. For example, the feedback may be partly observed, or the environment is not i.i.d. or can be operated by an adversary who aims to maximize the loss of the predictor. In this section, we will introduce some basic concepts about game theory and some fundamental theory of learning in games. We will focus on regret-based minimization procedures and limit our attention to finite strategic or normal form games. A more comprehensive study on this subject can be found in \citep{bianchi-2006-prediction,nisan2007algorithmic}.

\subsubsection{Game Playing and Nash Equilibrium}

{\bf $K$-Player Normal-Form Games.} Consider a game with $K$ players ($1<K<\infty$), where each player $k\in\{1,...,K \}$ can take $N_k$ possible actions. The players' actions can be represented by a vector $\mathbf i=(i_1,...,i_K)$, where $i_k\in\{1,...,N_k\}$ denotes the action of player $k$. The loss suffered by the player $k$ is denoted by $\ell^{(k)}(\mathbf i)$ since the loss is related to not only the action of player $k$ but the action of all the other players. During each iteration of the game, each player tries to take actions in order to minimize its own loss.

Using a mixed strategy, player $k$ takes actions based on a probability distribution $\mathbf p^{(k)}=(p_{1}^{(k)},\ldots,p_{N_k}^{(k)})$ over the set of $\{1,\ldots,N_k\}$ actions. In particular, the actions played by all the $K$ players can be denoted as a random vector $\mathbf I=(I_1,...,I_K)$, where $I_k$ is the action played by player $k$ which is a random variable taking value over the set of $\{1,\ldots,N_k\}$ actions distributed according to $\mathbf p^{(k)}$. The expected loss of player $k$ can be computed as
\bqs
\E \ell^{(k)}(\mathbf I) =\sum_{i_1=1}^{N_1}\cdot\cdot\cdot \sum_{i_K=1}^{N_K} p_{i_1}^{(1)} \times \cdot \cdot\cdot \times p_{i_K}^{(K)} \ell^{(k)}(i_1,...,i_K)
\eqs

{\bf Nash Equilibrium.}  This is an important notion in game theory. In particular, a collective strategy of all players $\mathbf p^{(1)}\times \cdot\cdot\cdot \times \mathbf p^{(K)}$ is called a {\it Nash equilibrium} if any mixed strategy among the $K$ players $\mathbf p^{(k)}$ is replaced by any new mixed strategy $\mathbf q^{(k)}$ while all other $K-1$ players' mixed strategies make no change, we have
\bqs
\E \ell^{(k)}(\mathbf I)\leq \E \ell^{(k)}(\mathbf I')
\eqs
where $\mathbf I'$ denotes the actions played by the $K$ players using the new strategies. This definition means that in a Nash Equilibrium, no player can achieve a lower loss by only changing its own strategy if other players do not change. In a Nash Equilibrium, each player gets its own optimal strategy and has no incentive of changing its strategy. One can prove that every finite game has at least one Nash equilibrium, but a game may have multiple Nash equilibria depending on the structure of the game and the loss functions.

\subsubsection{Repeated Two-player Zero-Sum Games}
A simple but important special class of $K$-Player Normal Form Games is the class of two-player zero-sum games where only one player plays against one opponent, i.e., $K=2$. {\it Zero-sum} means that for any action, the sum of losses of all players is zero.
This indicates that the game is purely competitive and a player's loss results in another player's gain. In such games, the first player is often called the row player, and the second player is called the column player whose goal is to maximize the loss of the first player. To simplify notation, we consider the row player has $N$ possible actions and the column player has $M$ possible actions. We denote by $L\in[0,1]^{N\times M}$ where $L(i,j)$ is the loss of the row player taking action $i$ while the column player chooses action $j$, and the mixed strategies for the row and column players denoted by $\mathbf p=(p_,\ldots,p_N)$ and $\mathbf q=(q_,\ldots,q_N)$, respectively. For the two mixed strategies $\mathbf p$ and $\mathbf q$, the expected loss for the row player (which is equivalent to the expected gain of the column player) can be computed by
\bqs
L(\mathbf p,\mathbf q) =\sum_{i=1}^N\sum_{j=1}^M p(i)q(j)L(i,j)
\eqs
A pair of mixed strategies $(\mathbf p,\mathbf q)$ is a Nash equilibrium if and only if
\bqs
L(\mathbf p,\mathbf q') \leq L(\mathbf p,\mathbf q) \leq L(\mathbf p',\mathbf q), \quad \forall \mathbf p', \forall \mathbf q'
\eqs
One natural solution to the two-player zero-sum games is to follow the minimax solution. In particular, for the row player using some strategy $\mathbf p$, the worst-case loss is at most $\max_{\mathbf q} L(\mathbf p, \mathbf q)$ if the column player makes the decision after seeing $\mathbf p$. Therefore, the worst-case optimal strategy (also called the minimax optimal strategy) for the row player is $\mathbf p ^* = \arg\min_{\mathbf p}\max_{\mathbf q} L(\mathbf p, \mathbf q)$. Similarly, the maximin optimal strategy for the column player is $\mathbf q ^* = \arg\max_{\mathbf q}\min_{\mathbf p} L(\mathbf p, \mathbf q)$. The pair of $(\mathbf p ^*,\mathbf q ^*)$ is called a minimax solution of the game. Surprisingly there is no difference between $\min_{\mathbf p}\max_{\mathbf q} L(\mathbf p, \mathbf q)$ and $\max_{\mathbf q}\min_{\mathbf p} L(\mathbf p, \mathbf q)$, which is known as the von Neumann's minimax theorem, a fundamental result of game theory.
\begin{theorem}(von Neumann's minimax theorem)
In a two-player zero-sum game, when two players follow the strategies of the minimax solution, they reach the same optimal value
\bqs
V^* = \min_{\mathbf p}\max_{\mathbf q} L(\mathbf p, \mathbf q) = \max_{\mathbf q} \min_{\mathbf p} L(\mathbf p, \mathbf q)
\eqs
\end{theorem}
$V^*$ is called the value of the game which is unique for a two-player zero-sum game. A pair of mixed strategies $(\mathbf p,\mathbf q)$ is a Nash equilibrium if and only if it achieves the value of game.

We can now relate game theory to online learning as the problem of learning to play repeated two-player zero-sum games. In the context of online learning, the row player is also called the {\it learner} and the column player is called the {\it environment}. The repeated game playing between the row player and the column player is treated as a sequence of $T$ rounds of interactions between the learner and the environment. On each round $t=1,\ldots,T$,
\begin{itemize}
\item the leaner chooses a mixed strategy $\mathbf p_t$;
\item the environment chooses a mixed strategy $\mathbf q_t$ (may be chosen by the knowledge $\mathbf p_t$);
\item the learner observes the losses $L(i,\mathbf q_t)$ \quad $\forall i\in[N]$
\end{itemize}
In general, the goal of the learner is to minimize the cumulative loss, i.e., $\sum_{t=1}^T L(\mathbf p_t,\mathbf q_t)$.

\section{Online Supervised Learning}
\def \w {\mathbf{w}}
\def \R {\mathbb{R}}
\def \D {\mathbf{D}}
\def \x {\mathbf{x}}
\def \C {\mathbf{C}}
\def \W {\mathbf{W}}
\def \y {\mathbf{y}}
\def \S {\mathbf{S}}
\def \A {\mathcal{A}}
\def \Ab {\bar{\A}}
\def \Kt {\widetilde{K}}
\def \k {\mathbf{k}}
\def \K {\mathbf{K}}
\def \Se {\mathcal{S}}
\def \E {\mathrm{E}}
\def \Rh {\widehat{R}}
\def \x {\mathbf{x}}
\def \p {\mathbf{p}}
\def \a {\mathbf{a}}
\def \diag {\mbox{diag}}
\def \b {\mathbf{b}}
\def \tr {\mbox{tr}}
\def \d {\mathbf{d}}
\def \z {\mathbf{z}}
\def \bh {\widehat{b}}
\def \y {\mathbf{y}}
\def \u {\mathbf{u}}
\def \L {\mathcal{L}}
\def \H {\mathcal{H}}
\def \g {\mathbf{g}}
\def \F {\mathcal{F}}
\def \Dh {\widehat{\D}}
\def \xh {\widehat{\x}}
\def \yh {\widehat{y}}
\def \c {\mathbf{c}}
\def \X {\mathcal{X}}
\def \sign {\mbox{sgn}}
\def \I {\mathbf{I}}
\def \w {\mathbf{w}}
\def \R {\mathbb{R}}
\def \D {\mathbf{D}}
\def \x {\mathbf{x}}
\def \H {\mathcal{H}}
\def \C {\mathbf{C}}
\def \W {\mathbf{W}}
\def \y {\mathbf{y}}
\def \S {\mathbf{S}}
\def \A {\mathcal{A}}
\def \Ab {\bar{\A}}
\def \Kt {\widetilde{K}}
\def \k {\mathbf{k}}
\def \K {\mathbf{K}}
\def \Se {\mathcal{S}}
\def \E {\mathrm{E}}
\def \Rh {\widehat{R}}
\def \x {\mathbf{x}}
\def \p {\mathbf{p}}
\def \a {\mathbf{a}}
\def \diag {\mbox{diag}}
\def \b {\mathbf{b}}
\def \tr {\mbox{tr}}
\def \d {\mathbf{d}}
\def \z {\mathbf{z}}
\def \bh {\widehat{b}}
\def \y {\mathbf{y}}
\def \u {\mathbf{u}}
\def \L {\mathcal{L}}
\def \H {\mathcal{H}}
\def \g {\mathbf{g}}
\def \F {\mathcal{F}}
\def \Dh {\widehat{\D}}
\def \xh {\widehat{\x}}
\def \yh {\widehat{y}}
\def \Kh {\widehat{K}}
\def \fh {\widehat{f}}
\def \f {\mathbf{f}}
\def \N {\mathcal{N}}
\def \Y  {\mathcal{Y}}
\def \Yh {\widehat{Y}}
\def \yh {\widehat{y}}
\def \S {\mathcal{S}}
\def \Hk {\H_{\kappa}}
\def \e {\mathbf{e}}
\def \P {\mathcal{P}}
\def \v {\mathbf{v}}
\def \V {\mathbf{V}}
\def \c {\mathbf{c}}
\def \ft {\widehat{f}}
\def \X {\mathcal{X}}
\def \sign {\mbox{sgn}}
\newtheorem{thm}{Theorem}
\newcommand{\argmin}{\operatornamewithlimits{argmin}}

\subsection{Overview}

In this section, we survey a family of ``online supervised learning" algorithms which define the fundamental approaches and principles for online learning methodologies toward supervised learning tasks \cite{shalev2011online,rakhlin2010online}.

We first discuss linear online learning methods, where a target model is a linear function. More formally, consider an input domain $\mathcal{X}$ and an output domain $\mathcal{Y}$ for a learning task, we aim to learn a hypothesis $f:\mathcal{X}\mapsto\mathcal{Y},$ where the target model $f$ is linear. For example, consider a typical linear binary classification task, our goal is to learn a linear classifier $f:\mathcal{X}\mapsto\{+1,-1\}$ as follows:
$f(\mathbf{x}_t;\mathbf{w}) = sgn(\mathbf{w} \cdot \mathbf{x}_t)$, where $\mathcal{X}$ is typically a $d$-dimensional vector space $\R^d$, $\mathbf{w}\in\mathcal{X}$ is a weight vector specified for the classifier to be learned, and $sgn(z)$ is an indicator function that outputs +1 when $z>0$ and -1 otherwise.
We review two major types of linear online learning algorithms: first-order online learning and second-order online learning algorithms. Following this, we discuss Prediction with expert advice, and Online Learning with Regularization. This is followed by reviewing nonlinear online learning using kernel based methods. We discuss a variety of kernel-based online learning approaches, their computational challenges, and several approximation strategies for efficient learning. We end this section by discussing the theory for converting using online learning algorithms to learn a batch model that can generalize well.

\subsection{First-order Online Learning}

In the following, we survey a family of first-order linear online learning algorithms, which exploit the first order information of the model during learning process.

\subsubsection{Perceptron}

Perceptron \citep{rosenblatt1958perceptron,Agmon1954,Novikoff1962} is the oldest algorithm for online learning. Algorithm \ref{alg:perceptron} gives the Perceptron algorithm for online binary classification.
\begin{algorithm}[hptb]
\label{alg:perceptron}
\caption{Perceptron}
\begin{algorithmic}
\STATE \textbf{INIT}: $\mathbf{w}_1=0$\\
\FOR{ $t=1,2,\ldots,T$}
    \STATE Given an incoming instance $\mathbf{x}_t$, predict \\$\hat{y}_t = f_t(\mathbf{x}_t)=sign(\mathbf{w}_t \cdot \mathbf{x}_t)$;
    \STATE Receive the true class label $y_t \in \{+1, -1\}$;
    \IF {$\hat{y}_t \neq y_t$}
    \STATE  $\mathbf{w}_{t+1} \leftarrow \mathbf{w}_{t} + y_t\mathbf{x}_t$;
    \ENDIF
\ENDFOR
\end{algorithmic}
\end{algorithm}

In theory, by assuming the data is separable with some margin $\gamma$, the Perceptron algorithm makes at most $\big(\frac{R}{\gamma}\big)^2$ mistakes, where the margin $\gamma$ is defined as $\gamma = \min_{t\in[T]}{|\mathbf{x}_t \cdot \mathbf{w}^*|}$ and $R$ is a constant such that $\forall t\in[T], \|\mathbf{x}_t\|\leq R$. The larger the margin $\gamma$ is, the tighter the mistake bound will be.

In literature, many variants of Perceptron algorithms have been proposed. One simple modification is the ``normalized Perceptron" algorithm that differs only in the updating rule as follows:
$$\mathbf{w}_{t+1} = \mathbf{w}_{t} + y_t\frac{\mathbf{x}_t}{\|\mathbf{x}_t\|}$$
The mistake bound of the ``normalized Perceptron" algorithm can be improved from $\big(\frac{R}{\gamma}\big)^2$ to $\big(\frac{1}{\gamma}\big)^2$ for the separable case due to the normalization effect.

\subsubsection{Winnow}
Unlike the Perceptron algorithm that uses additive updates, Winnow \citep{littlestone1988learning} employs multiplicative updates. The problem setting is slightly different from the Perceptron: $\mathcal{X}=\{0,1\}^d$ and $y\in\{0, 1\}$. The goal is to learn a classifier $f(x_1,\dots,x_n)=x_{i_1}\vee...\vee x_{i_k}$ called monotone disjunction, where $i_k\in{}1,\dots, d$. The separating hyperplane for this classifier is given by $x_{i_1}+...+x_{i_k}$. The Winnow algorithm is outlined in Algorithm \ref{alg:winnow}.

\begin{algorithm}[hptb]
\label{alg:winnow}
\caption{Winnow}
\begin{algorithmic}
\STATE \textbf{INIT}: $\mathbf{w}_1={\mathbf{1}}^d$, constant $\alpha>1$ (e.g.,$\alpha=2$)\\
\FOR{ $t=1,2,\ldots,T$}
    \STATE Given an instance $\mathbf{x}_t$, predict $\hat{y}_t =\mathbb{I}_{\w_t\cdot\x_t\geq\theta}$ (outputs 1 if statement holds and 0 otherwise);
    \STATE Receive the true class label $y_t \in \{1, 0\}$;
    \IF {$\hat{y}_t=1, y_t=0$}
        \STATE set $w_i=0$ for all $x_{t,i}=1$ (``elimination" or ``demotion"),
    \ENDIF
    \IF{$\hat{y}_t=0, y_t=1$}
        \STATE set $w_i=\alpha w_i$ for all $x_{t,i}=1$ (``promotion").
    \ENDIF
\ENDFOR
\end{algorithmic}
\end{algorithm}

The Winnow algorithm has a mistake bound $\alpha k(\log_\alpha \theta +1)+n/\theta$ where $\alpha>1$ and $\theta\geq 1/\alpha$ and the target function is a $k$-literal monotone disjunction.

\subsubsection{Passive-Aggressive Online Learning (PA)}

This is a popular family of first-order online learning algorithms which generally follows the principle of margin-based learning~\citep{crammer2006online}. Specifically, given an instance $\x_t$ at round $t$, PA formulates the updating optimization as follows:
\begin{eqnarray}
\w_{t+1}=\arg\min_{\w\in\R^d}\frac{1}{2}||\w-\w_t||^2~~~~~s.t. ~~\ell_t(\w)=0
\end{eqnarray}
where $\ell_t(\w)=\max(0,1-y_t\w\cdot\x_t)$ is the hinge loss. The above resulting update is passive whenever the hinge loss is zero, i.e., $\w_{t+1} = \w_t$ whenever $\ell=0$. In contrast, whenever the loss is
nonzero, the approach will force $\w_{t+1}$ aggressively to satisfy the constraint regardless of any step-size; the algorithm is thus named as ``Passive-Aggressive" (PA)~\citep{crammer2006online}.
Specifically, PA aims to keep the updated classifier $\w_{t+1}$ stay close to the previous classifier (``passiveness") and ensure every incoming instance to be classified correctly by the updated classifier (``aggressiveness"). The regular PA algorithm assumes training data is always separable, which may not be true for noisy training data in real-world applications. To overcome such limitation, two variants of PA relax the assumption as:
\begin{eqnarray}\begin{aligned}
&\mathrm{PA-I}:\w_{t+1}=\arg\min_{\w\in\R^d}\frac{1}{2}||\w-\w_t||^2+C\xi\\&\text{subject to}\quad \ell_t(\w)\leq\xi ~~\text{and}~~ \xi\geq0\\
&\mathrm{PA-II}:\w_{t+1}=\arg\min_{\w\in\R^d}\frac{1}{2}||\w-\w_t||^2+C\xi^2\\&\text{subject to}\quad \ell_t(\w)\leq\xi
\end{aligned}\end{eqnarray}
where $C$ is a positive parameter to balance the tradeoff between ``passiveness" (first regularization term) and ``aggressiveness" (second slack-variable term). By solving the three optimization tasks, we can derive the closed-form updating rules of three PA algorithms:
$$\w_{t+1}=\w_t+\tau_ty_t\x_t, \quad \tau_t=
   \begin{cases}
   \ell_t/||\x_t||^2& (\text{PA})\\
   \min\{C, \ell_t/||\x_t||^2\}& (\text{PA-I})\\
   \frac{\ell_t}{||\x_t||^2+\frac{1}{2C}}& (\text{PA-II})\\
   \end{cases}
$$
It is important to note a major difference between PA and Perceptron algorithms. Perceptron makes an update only when there is a classification mistake. However, PA algorithms aggressively make an update whenever the loss is nonzero (even if the classification is correct). In theory~\citep{crammer2006online}, PA algorithms have comparable mistake bounds as the Perceptron algorithms, but empirically PA algorithms often outperform Perceptron significantly. The PA algorithms are outlined in Algorithm \ref{alg:pa}.

\begin{algorithm}[hptb]
	\label{alg:pa}
	\caption{Passive Aggressive Algorithms}
	\begin{algorithmic}
		\STATE  \textbf{INIT}: $\w_1$, Aggressiveness Parameter $C$;
		\FOR{ $t=1,2,\ldots,T$}
		\STATE Receive $\x_t \in \R^d$, predict $\hat{y_t}$ using $\w_t$;
		\STATE Suffer loss $\ell_t(\w_t)$;
		\STATE Set $\tau = \begin{cases}
		\ell_t/||\x_t||^2& (\text{PA})\\
		\min\{C, \ell_t/||\x_t||^2\}& (\text{PA-I})\\
		\frac{\ell_t}{||\x_t||^2+\frac{1}{2C}}& (\text{PA-II})\\
		\end{cases}$
		\STATE Update $\w_{t+1} = \w_t+\tau_ty_t\x_t$;
		\ENDFOR
	\end{algorithmic}
\end{algorithm}

\subsubsection{Online Gradient Descent (OGD)}
Many online learning problems can be formulated as an online convex optimization task, which can be solved by applying the OGD algorithm. Consider the online binary classification as an example,
where we use the hinge loss function, i.e., $\ell_t(\w)=\max(0,1-y_t\w\cdot\x_t)$. By applying the OGD algorithm, we can derive the updating rule as follows:
\begin{eqnarray}
\w_{t+1} =  \w_t + \eta_t y_t \x_t
\end{eqnarray}
where $\eta_t$ is the learning rate (or step size) parameter. The OGD algorithm is outlined in Algorithm \ref{alg:ogd}, where any generic convex loss function can be used. $\Pi_\S$ is the projection function to constrain the updated model to lie in the feasible domain.

\begin{algorithm}[hptb]
	\label{alg:ogd}
	\caption{Online Gradient Descent}
	\begin{algorithmic}
		\STATE  \textbf{INIT}: $\w_1$, convex set $\S$, step size $\eta_t$;
		\FOR{ $t=1,2,\ldots,T$}
		\STATE Receive $\x_t \in \R^d$, predict $\hat{y_t}$ using $\w_t$;
		\STATE Suffer loss $\ell_t(\w_t)$;
		\STATE Update $\w_{t+1} = \Pi_\S (\w_t - \eta_t \nabla \ell_t(\w_t))$
		\ENDFOR
	\end{algorithmic}
\end{algorithm}
OGD and PA share similar updating rules but differ in that OGD often employs some predefined learning rate scheme while PA chooses the optimal learning rate $\tau_t$ at each round (but subject to a predefined cost parameter $C$). In literature, different OGD variants have been proposed to improve either theoretical bounds or practical issues, such as adaptive OGD~\citep{hazan2007adaptive}, and mini-batch OGD~\citep{dekel2012optimal}, amongst others.

\subsubsection{Other first-order algorithms}

In literature, there are also some other first-order online learning algorithms, such as Approximate Large Margin Algorithms (ALMA) \citep{DBLP:journals/jmlr/Gentile01} which is a large margin variant of the p-norm Perceptron algorithm, and the Relaxed Online Maximum Margin Algorithm (ROMMA) \citep{DBLP:journals/ml/LiL02}. Many of these algorithms often follow the principle of large margin learning. The metaGrad algorithm \citep{van2016metagrad} tries to adapt the learning rate automatically for faster convergence.

\subsection{Second-Order Online Learning}

Unlike the first-order online learning algorithms that only exploit the first order derivative information of the gradient for the online optimization tasks, second-order online learning algorithms exploit both first-order and second-order information in order to accelerate the optimization convergence. Despite the better learning performance, second-order online learning algorithms often fall short in higher computational complexity. In the following we present a family of popular second-order online learning algorithms.

\subsubsection{Second Order Perceptron (SOP)}
SOP algorithm~\citep{cesa2005second} is able to exploit certain geometrical properties
of the data which are missed by the first-order algorithms.

For better understanding, we first introduce the whitened Perceptron algorithm, which strictly speaking, is not an online learning method. Assuming that the instances $\x_1,...,\x_T$ are preliminarily available, we can get the correlation matrix $M=\sum_{t=1}^T\x_t\x_t^\top$. The whitened Perceptron algorithm is simply the standard Perceptron run on the transformed sequence $(M^{-1/2}\x_1,y_1),...,(M^{-1/2}\x_T,y_T)$. By reducing the correlation matrix of the transformed instances, the whitened Perceptron algorithm can achieve significantly better mistake bound.

SOP can be viewed as an online variant of the whitened Perceptron algorithm. In online setting the correlation matrix $M$ can be approximated by the previously seen instances. SOP is outlined in Algorithm \ref{alg:sop}\\

\begin{algorithm}[hptb]
\label{alg:sop}
\caption{SOP}
\begin{algorithmic}
\STATE \textbf{INIT}: $\mathbf{w}_1=0$, $X_0$=[], $\mathbf{v}_0=0$, $k=1$\\
\FOR{ $t=1,2,\ldots,T$}
    \STATE Given an incoming instance $\mathbf{x}_t$, set $S_t=[X_{k-1}~\x_t]$,
    \STATE predict $\hat{y}_t = f_t(\mathbf{x}_t)=sign(\mathbf{w}_t \cdot \mathbf{x}_t)$, where $\w_t=(aI_n+S_tS_t^\top)^{-1}\mathbf{v}_{k-1}$
    \STATE Receive the true class label $y_t \in \{+1, -1\}$;
    \IF {$\hat{y}_t \neq y_t$}
     \STATE $\mathbf{v}_k=\mathbf{v}_{k-1}+y_t\x_t$, $X_k=S_t$, $k=k+1$.
     \ENDIF
\ENDFOR
\end{algorithmic}
\end{algorithm}

Here $a\in\R^+$ is a parameter that guarantees the existence of the matrix inverse.

\subsubsection{Confidence Weighted Learning (CW)}
The CW algorithm \citep{dredze2008confidence} is motivated by the following observation: the frequency of occurrence of different features may differ a lot in an online learning task.  (For example) The parameters of binary features are only updated when the features occur. Thus, the frequent features typically receive more updates and are estimated more accurately compared to rare features. However, no distinction is made between these feature types in most online algorithms. This indicates that the lack of second order information about the frequency or confidence of the features can hurt the learning.

In the CW setting, we model the linear classifier with a Gaussian distribution, i.e., $\w\sim \mathcal{N}(\bm\mu ,\Sigma)$, where $\bm\mu \in \R^{d}$ is the mean vector and $\Sigma\in \R^{d\times d}$ is the covariance matrix. When making a prediction, the prediction confidence $M=\w\cdot \x$ also follows a Gaussian distribution: $M\sim \mathcal{N}(\bm\mu_{M} ,\Sigma_{M})$, where $\mu_{M}=\bm\mu \cdot \x$ and $\Sigma_{M}=\x^{\top}\Sigma \x$.

Similar to the PA update strategy, the update rule in round $t$ can be obtained by solving the following convex optimization problem:
\begin{equation}
\begin{aligned}
(\bm\mu_{t+1},\Sigma_{t+1})=\arg\min_{\bm\mu\in\R^d} \text{D}_{\text{KL}}\left(\mathcal{N}(\bm\mu ,\Sigma)||\mathcal{N}(\bm\mu_t ,\Sigma_t)\right)\quad s.t. ~\Pr[y_tM_t\geq 0]\geq\eta
\end{aligned}
\end{equation}
The objective function means that the new distribution should stay close to the previous distribution so that the classifier does not forget the information learnt from previous instances, where the distance between the two distributions is measured by the KL divergence. The constraint means that the new classifier should classify the new instance $\x_t$ correctly with probability higher than a predefined threshold parameter $\eta\in(0,1)$.

Note that this is only the basic form of confidence weighted algorithms and has several drawbacks. 1) Similar to the hard margin PA algorithm, the constraint forces the new instance to be correctly classified, which makes this algorithm very sensitive to noise. 2) The constraint is in a probability form. It is easy to solve a problem with the constraint $g(\bm\mu_M,\Sigma_M)<0$. However, a problem with a probability form constraint is only solvable when the distribution is known. Thus, this method faces difficulty in generalizing to other online learning tasks where the constraint does not follow a Gaussian distribution.

\subsubsection{Adaptive Regularization of Weight Vectors (AROW)}

AROW \citep{crammer2009adaptive} is a variant of CW that is designed for non-separable data. This algorithm adopts the same Gaussian distribution assumption on classifier vector $\w$ while the optimization problem is different. By recasting the CW constraint as  regularizers, the optimization problem can be formulated as:
\begin{equation}\begin{aligned}
\mathcal{C}(\bm\mu,\Sigma)=\text{D}_{\text{KL}}\left(\mathcal{N}(\bm\mu ,\Sigma)||\mathcal{N}(\bm\mu_t ,\Sigma_t)\right) +\lambda_1\ell(y_t,\bm\mu\cdot\x_t)+\lambda_2\x_t^\top\Sigma\x_t
\end{aligned}\end{equation}
where $\ell(y_t,\bm\mu\cdot\x_t)=(\max(0,1-y_t\bm\mu\cdot\x_t))^2$ is the squared-hinge loss. During each iteration, the update rule is obtained by solving the optimization problem:
$$(\bm\mu_{t+1},\Sigma_{t+1})=\arg\min_{\bm\mu\in\R^d}(\mathcal{C}(\bm\mu,\Sigma))$$
which balances the three desires. First, the parameters should not change radically on
each round, since the current parameters contain information about previous examples (first
term). Second, the new mean parameters should predict the current example with low loss
(second term). Finally, as we see more examples, our confidence in the parameters should
generally grow (third term). $\lambda_1$ and $\lambda_2$ are two positive parameters that control the weight of the three desires.

Besides the robustness to noisy data, another important advantage of AROW is its ability to be easily generalized to other online learning tasks, such as Confidence Weighted Online Collaborative Filtering algorithm \citep{lu2013second} and Second-Order Online Feature Selection \citep{wu2014massive}.

\subsubsection{Soft Confidence weighted Learning (SCW)}
This is a variant of CW learning in order to deal with non-separable data \citep{wang2012exact,wang2016soft}. Different from AROW which directly adds loss and confidence regularization, and thus loses the adaptive margin property, SCW exploits adaptive margin by assigning different margins for different instances via a probability formulation. Consequently, SCW tends to be more efficient and effective.

Specifically, the constraint of CW can be rewritten as $y_t(\bm\mu\cdot\x_t)\geq\phi\sqrt{\x_t^\top\Sigma\x_t}$. Thus, the loss function can be defined as:$\ell(\mathcal{N}(\bm\mu,\Sigma);(\x_t,y_t))=\max(0,\phi\sqrt{\x_t^\top\Sigma\x_t}-y_t(\bm\mu\cdot\x_t))$. The original CW optimization can be rewritten as:
\begin{eqnarray*}
(\bm\mu_{t+1},\Sigma_{t+1})=\arg\min_{\bm\mu\in\R^d} \text{D}_{\text{KL}}\left(\mathcal{N}(\bm\mu ,\Sigma)||\mathcal{N}(\bm\mu_t ,\Sigma_t)\right)\\
\text{subject to} ~\ell(\mathcal{N}(\bm\mu,\Sigma);(\x_t,y_t))=0
\end{eqnarray*}
Inspired by soft-margin PA variants, SCW generalized CW into two soft-margin formulations:
\begin{eqnarray*}
(\bm\mu_{t+1},\Sigma_{t+1})=\arg\min_{\bm\mu\in\R^d} \text{D}_{\text{KL}}\left(\mathcal{N}(\bm\mu ,\Sigma)||\mathcal{N}(\bm\mu_t ,\Sigma_t)\right) + C\ell(\mathcal{N}(\bm\mu,\Sigma);(\x_t,y_t))\\
(\bm\mu_{t+1},\Sigma_{t+1})=\arg\min_{\bm\mu\in\R^d} \text{D}_{\text{KL}}\left(\mathcal{N}(\bm\mu ,\Sigma)||\mathcal{N}(\bm\mu_t ,\Sigma_t)\right) + C\ell^2(\mathcal{N}(\bm\mu,\Sigma);(\x_t,y_t))
\end{eqnarray*}
where $C\in\R^+$ is a parameter controls the aggressiveness of this algorithm, similar to the $C$ in PA algorithm. The two algorithms are termed ``SCW-I'' and ``SCW-II''.

\subsubsection{Other second-order algorithms}
The confidence weighted idea also works for other online learning tasks such as multi-class classification \citep{crammer2009multi}, active learning \citep{dredze2008active} and structured-prediction \citep{mejer2010confidence}. There are many other online learning algorithms that adopt second order information: IELLIP \citep{yang2009online} assumes the objective classifier $\w$ lies in an ellipsoid and incrementally updates the ellipsoid based on the current received instance. Other approaches include New variant of Adaptive Regularization (NAROW) \citep{orabona2010new} and the Normal Herding method via Gaussian Herding (NHERD) \citep{crammer2010learning}. Recently, Sketched Online Newton \citep{luo2016efficient} made significant improvements to speed-up second order online learning.

\subsection{Prediction with Expert Advice}

This is an important online learning subject \citep{roughgarden2017online} with many applications. A general setting is as follows. A learner has $N$ experts to choose from, denoted by integers $1,\dots,N$. At each time step $t$, the learner decides on a distribution $\p_t$ over the experts, where $p_{t,i}\ge 0$ is the weight of each expert $i$, and $\sum^N_{i=1}p_{t,i}=1$. Each expert $i$ then suffers some loss $\ell_{t,i}$ according to the environment. The overall loss suffered by the learner is $\sum^N_{i=1}p_{t,i}\ell_{t,i}=\p_t^\top\mathbf{\ell}_t$, i.e., the weighted average loss of the experts with respect to the distribution chosen by the learner.

Typically we assume that the loss suffered by any expert is bounded. Specifically, we assume $\ell_{t,i}\in[0,1]$ without loss of generality. Besides this condition, no assumption is made on the form of the loss, or about how they are generated. Suppose the cumulative losses experienced by each expert and the forecaster are calculated respectively as follows:
\bqs
L_{t,i}=\sum^t_{s=1}\ell_{s,i},\quad L_t=\sum^t_{s=1}\p_t^\top\mathbf{\ell}_t.
\eqs
The loss difference between the forecaster and the expert is known as the ``regret'', i.e.,
\bqs
R_{t,i} = L_t - L_{t,i},\quad i=1,\ldots,N.
\eqs
The goal of learning the forecaster is to make the regret with respect to each expert as small as possible, which is equivalent to minimizing the overall regret, i.e.,
\bqs
R_T=\max_{1\le i\le N} R_{T,i}=L_T-\min_{1\le i\le N} L_{T,i}
\eqs
In general, online prediction with expert advice aims to find an ideal forecaster to achieve a vanishing per-round regret, a property known as the {\it Hannan-consistency}~\citep{hannan1957approximation},i.e.,
\bqs
R_T = o(T) \Leftrightarrow \lim_{T\rightarrow\infty}\frac{1}{T}\left(L_T - \min_{1\le i\le N}L_{T,i}\right)
\eqs
An online learner satisfying the above is called a Hannan-consistent forecaster~\citep{bianchi-2006-prediction}. Next we review some representative algorithms for prediction with expert advice.

\subsubsection{Weighted Majority Algorithms} The weighted majority algorithm (WM) is a simple but widely studied algorithm that makes a binary prediction based on a series of expert advices \citep{DBLP:journals/iandc/LittlestoneW94,littlestone1989weighted}. The simplest version is shown in Algorithm \ref{alg:wmm}, where $\beta \in (0,1)$ is a user specified discount rate parameter.
\begin{algorithm}[hptb]
	\label{alg:wmm}
	\caption{Weighted Majority}
	\begin{algorithmic}
		\STATE \textbf{INIT}: Initialize the weights $p_1, p_2,...p_N$ of all experts to $1/N$. \\
		\FOR{ $t=1,2,\ldots,T$}
		\STATE Get the prediction $x_1,...,x_N$ from $N$ experts.
		\STATE Output 1 if $\sum_{i:x_i=1}p_i\geq\sum_{i:x_i=0}p_i$; otherwise output 0.
		\STATE Receive the true value; if the $i$-th expert made a mistake, then $p_i=p_i*\beta$
		\ENDFOR
	\end{algorithmic}
\end{algorithm}

\subsubsection{Randomized Multiplicative Weights Algorithms}
This algorithm works under the same assumption that the expert advices are all binary \citep{arora2012multiplicative}. While the prediction is random, the algorithm gives the prediction 1 with probability of $\gamma=\frac{\sum_{i:x_i=1}p_i}{\sum_i^N p_i}$ and 0 with probability of $1-\gamma$.

\subsubsection{Hedge Algorithm}

The Hedge algorithm \citep{DBLP:journals/jcss/FreundS97} is perhaps the most well-known approach for online prediction with expert advice, which can be viewed as a direct generalization of Littlestone and Warmuth's weighted majority algorithm~\citep{DBLP:journals/iandc/LittlestoneW94,littlestone1989weighted}. The working of Hedge algorithm is shown in Algorithm \ref{alg:hedge}.
\begin{algorithm}[hptb]
	\label{alg:hedge}
	\caption{Hedge Algorithm}
	\begin{algorithmic}
		\STATE \textbf{INIT}: $\beta\in[0,1]$, initial weight vector $\w_1\in[0,1]^N$ with $\sum^N_{i=1}w_{1,i}=1$\\
		\FOR{ $t=1,2,\ldots,T$}
		\STATE set distribution $\p_t=\frac{\w_t}{\sum^N_{i=1}w_{t,i}}$;
		\STATE Receive loss $\mathbf{\ell}_t\in[0,1]^N$ from environment;
		\STATE Suffer loss $\p_t^\top\mathbf{\ell}_t$;
		\STATE Update the new weight vector to $w_{t+1,i}=w_{t,i}\beta^{\ell_{t,i}}$\\
		\ENDFOR
	\end{algorithmic}
\end{algorithm}
The algorithm maintains a weight vector whose value at time $t$ is denoted $\w_t=(w_{t,1},\ldots,w_{t,N})$. At all times, all weights are nonnegative. All of the weights of the initial weight vector $\w_1$ must be nonnegative and sum to one, which can be considered as a prior over the set of experts. If it is believed that one expert performs the best, it is better to assign it more weight. If no prior is known, it is better to set all the initial weights equally, i.e., $w_{1,i}=1/N$ for all $i$.  The algorithm uses the normalized distribution to make prediction, i.e., $\p_t={\w_t}/{\sum^N_{i=1}w_{t,i}}$. After the loss $\ell_t$ is disclosed, the weight vector $\w_t$ is updated using a multiplicative rule $w_{t+1,i}=w_{t,i}\beta^{\ell_{t,i}},\quad \beta\in[0,1]$, which implies that the weight of expert $i$ will exponentially decrease with the loss $\ell_{t,i}$.
In theory, the Hedge algorithm is proved to be Hannan consistent.

\subsubsection{EWAF Algorithms}

Besides Hedge, there are some other algorithms for online prediction with expert advice under more challenging settings, including exponentially weighted average forecaster (EWAF) and Greedy Forecaster (GF)~\citep{bianchi-2006-prediction}. We will mainly discuss EWAF, which is shown in Algorithm \ref{alg:ewaf}

\begin{algorithm}[hptb]
	\caption{EWAF}
	\label{alg:ewaf}
	\begin{algorithmic}
		\STATE \textbf{INIT}: a poll of experts $f_i,\quad i=1,\ldots,N$ and $L_{0,i}=0,\ i=1,\ldots,N$, and learning rate $\eta$ \\
		\FOR{ $t=1,2,\ldots,T$}
		\STATE The environment chooses the next outcome $y_t$ and the expert advice $\{f_{t,i}\}$;
		\STATE The expert advice is revealed to the forecaster
		\STATE The forecaster chooses the prediction $\hat{p}_t=\frac{\sum^N_{i=1}\exp(-\eta L_{t-1,i})f_{t,i}}{\sum^N_{i=1}\exp(-\eta L_{t-1,i})}$
		\STATE The environment reveals the outcome $y_t$;
		\STATE The forecaster incurs loss $\ell(\hat{p}_t,y_t)$ and;
		\STATE Each expert incurs loss $\ell(f_{t,i},y_t)$
		\STATE The forecaster update the cumulative loss $L_{t,i}=L_{t-1,i}+\ell(f_{t,i},y_t)$
		\ENDFOR
	\end{algorithmic}
\end{algorithm}

The difference between EWAF and Hedge is that the loss in Hedge is the inner product between the distribution and the loss suffered by each expert, while for EWAF, the loss is between the prediction and the true label, which can be much more complex.

\subsubsection{Parameter-free Online Learning}
A category of prediction with expert advice deals with learning without user specified learning rate. It is a difficult task to set a learning rate prior to the learning procedure. To address this issue, parameter-free online algorithms were proposed. Among the early efforts, \cite{chaudhuri2009parameter} proposed a variant of Hedge Algorithm without the use of a learning rate. The proposed method achieved optimal regret matching the best bounds of all the previous algorithms (with optimally-tuned parameters). \cite{chernov2010prediction} improved this bound, but did not have a closed-form solution. There were further extensions which derived data-dependent bounds too \citep{luo2015achieving,koolen2015second}.

\subsection{Online Learning with Regularization}

Traditional online learning methods learn a classifier $\w\in\R^d$ where the magnitude of each element $|\w^j|$ weights the importance of each feature, which are often non-zero. When dealing with high dimensional data, traditional online learning methods suffer from expensive computational time and space costs. This drawback is often addressed using regularization by performing Sparse online learning, which aims to exploit the sparsity property with real-world high-dimensional data. Specifically, a batch sparse learning problem can be formalized as:
$$
P(\w)= \frac{1}{n}\sum_{i=1}^n\ell_t(\w) + \phi_s(\w)
$$
where $\phi_s$ is a sparsity-inducing regularizer. For example, when choosing $\phi_s = \lambda||\w||_0$, it is equivalent to imposing a hard constraint on the number of nonzero elements in $\w$. Instead of choosing  $\ell_0$-norm which is hard to be optimized, a more commonly used regularizer is $\ell_1$-norm, i.e., $\phi_s = \lambda||\w||_1$, which can induce sparsity of the weight vector but does not explicitly constrain the number of nonzero elements. The following reviews some popular sparse online learning methods.

\subsubsection{Truncated Gradient Descent}
A straightforward idea to sparse online learning is to modify Online Gradient Descent and round small coefficients of the weight vector to $0$ after every $K$ iterations:
\begin{eqnarray*}
	\w_{t+1}=T_0(\w_t-\eta\nabla\ell_t(\w_t),\theta)
\end{eqnarray*}
where the function $T_0(\mathbf{v},\theta)$ performs an element-wise rounding on the input vector: if the $j$-th element $v^j$ is smaller than the threshold $\theta$, set $v^j=0$. Despite its simplicity, this method struggles to provide satisfactory performance because the aggressive rounding strategy may ignore many useful weights which may be very small due to low frequency of appearance.

Motivated by addressing the above limitation, the Truncated Gradient Descent (TGD) method \citep{langford2009sparse} explores a less aggressive version of the truncation function:
\begin{eqnarray*}\begin{aligned}
		&\w_{t+1}=T_1(\w_t-\eta\nabla\ell_t(\w_t),\eta g_i,\theta)~~~\\
		&\mathrm{where}~~T_1(v^j,\alpha,\theta)=
		\begin{cases}
			\max(0,v^j-\alpha)& \text{if  } v^j\in[0,\theta]\\
			\min(0,v^j+\alpha)& \text{if  } v^j\in[-\theta,0]\\
			v^j&\text{otherwise}
	\end{cases}\end{aligned}
\end{eqnarray*}
where $g_i>0$ is a parameter that controls the level of aggressiveness of the truncation. By exploiting sparsity, TGD achieves efficient time and space complexity that is linear with respect to the number of nonzero features and independent of the dimensionality $d$. In addition, it is proven to enjoy a regret bound of $O(\sqrt{T})$ for convex loss functions when setting $\eta=O(1/\sqrt{T})$.

\subsubsection{Forward Looking Subgradients (FOBOS)}

Consider the objective function in the $t$-th iteration of a sparse online learning task as $\ell_t(\w)+r(\w)$, FOBOS \citep{duchi2009efficient} assumes $f_t$ is a convex loss function (differentiable), and $r$ is a sparsity-inducing regularizer (non-differentiable). FOBOS updates the classifier in the following two steps:
\begin{itemize}
	\item[] (1)~Perform Online Gradient Descent: $\w_{t+\frac{1}{2}}=\w_t-\eta_t\nabla \ell_t(\w_t)$
	\item[] (2)~Project the solution in (i) such that the projection stays close to the interim vector $\w_{t+\frac{1}{2}}$ and (ii) has a low complexity due to  $r$: $$\w_{t+1}=\arg\min_{\w}\{\frac{1}{2}||\w-\w_{\frac{1}{2}}||^2+\eta_{t+\frac{1}{2}}r(\w)\}$$
\end{itemize}
When choosing $\ell_1$-norm as the regularizer, the above optimization can be solved with the closed-form solution for each coordinate:
$$
w_{t+1}^j=\sign(w_{t+\frac{1}{2}}^j)\left[|w_{t+\frac{1}{2}}^j|-\eta_{t+\frac{1}{2}}\right]_+
$$
The FOBOS algorithm with $\ell_1$-norm regularizer can be viewed as a special case of TGD, where the truncation threshold $\theta=\infty$, and the truncation frequency $K=1$. When $\eta_{t+\frac{1}{2}}=\eta_{t+1}$ and $\eta_t=O(1/\sqrt{t})$, this algorithm also achieves $O(\sqrt{T})$ regret bound.

\subsubsection{Regularized Dual Averaging (RDA)} Motivated by the theory of dual-averaging techniques \citep{nesterov2009primal}, the RDA algorithm \citep{xiao2009dual}  updates the classifier by:
$$
\w_{t+1}=\arg\min_{\w}\left\{\bar{\mathbf{g}_t}^\top\w+\Psi(\w)+\frac{\beta_t}{t}h(\w)\right\}
$$
where $\Psi(\w)$ is the original sparsity-inducing regularizer, i.e., $\Psi(\w)=\lambda||\w||_1$; $h(\w)=\frac{1}{2}||\w||^2$ is an auxiliary strongly convex function and $\bar{\mathbf{g}_t}$ is the averaged gradients of all previous iterations, i.e., $\bar{\mathbf{g}}=\frac{1}{t}\sum_{\tau=1}^t\nabla \ell_\tau(\w_\tau)$. Setting the step size $\beta_t=\gamma\sqrt{t}$, one can derive the closed-form solution:
$$
w_{t+1}^j=
\begin{cases}
0 &\text{if  } |\bar{g_t}^j|<\lambda\\
-\frac{\sqrt{t}}{\gamma}(\bar{g_t}^j-\lambda\sign(\bar{g_t}^j))&\text{otherwise}
\end{cases}
$$
To further pinpoint the differences between RDA and FOBOS, we rewrite FOBOS in the same notation as RDA:
$$
\w_{t+1}=\arg\min_{\w}\left\{\mathbf{g}_t^\top\w+\Psi(\w)+\frac{1}{2\alpha_t}||\w-\w_t||^2_2\right\}
$$
Specifically, RDA differs from FOBOS in several aspects. First, RDA uses the averaged gradient instead of the current gradient. Second, $h(\w)$ is a global proximal function instead of its local Bregman divergence. Third, the coefficient for $h(\w)$ is $\beta_t/t=\gamma/\sqrt{t}$ which is $1/\alpha_t=O(\sqrt{t})$ in FOBOS. Fourth, the truncation of RDA is a constant $\lambda$, while the truncation in FOBOS $\eta_{t+\frac{1}{2}}$ decrease with a factor $\sqrt{t}$. Clearly, RDA uses a more aggressive truncation threshold, thus usually generates significantly more sparse solutions. RDA also ensures the $O(\sqrt{T})$ regret bound.

\subsubsection{Adaptive Regularization}
One major issue with both FOBOS and RDA is that the auxiliary strongly convex function $h(\w)$ may not fully exploit the geometry information of underlying data distribution.
Instead of choosing $h(\w)$ as an $\ell_2$-norm $\frac{1}{2}||\w||^2$ in RDA or a Mahalanobis norm $||\cdot||_{H_t}$ in FOBOS, \citep{duchi2011adaptive} proposed a data-driven adaptive regularization for $h(\w)$, i.e.,
$$
h_t(\w)=\frac{1}{2}\w^\top H_t\w
$$
where $H_t=(\sum_{\tau=1}^t\mathbf{g}_\tau\mathbf{g}_\tau^\top)^\frac{1}{2}$ accumulates the second order info from the previous instances over time. Replacing the previous $h(\w)$ in both RDA and FOBOS by the temporal adaptation function $h_t(\w)$, \citep{duchi2011adaptive} derived two generalized algorithms (Ada-RDA and Ada-FOBOS) with the solutions as follows respectively.

Ada-RDA:
$$
{w_{t+1}^j} =
\begin{cases}
0 &\text{if  } |\bar{g_t}^j|<\lambda\\
-\frac{t}{\beta H_{t,jj}}(\bar{g_t}^j-\lambda\sign(\bar{g_t}^j))&\text{otherwise}
\end{cases}
$$
Ada-FOBOS:
\begin{equation}
{w_{t+1}^j}=\sign(w_t^i-\frac{\alpha_t}{H_{t,jj}}g_t^j)\left[|w_t^i-\frac{\alpha_t}{H_{t,jj}}g_t^j| -\frac{\alpha_t\lambda}{H_{t,ii}} \right]_+
\end{equation}
In the above, $H_t$ is approximated by a diagonal matrix since computing the root of a matrix is computationally impractical in high-dimensional data.

\subsubsection{Online Feature Selection}
Online feature selection \citep{hoi2012online,wang2014onlineFS,kale2017adaptive,wu2017large} is closely related to sparse online learning in that they both aim to learn an efficient classifier for very high dimensional data. However, the sparse learning algorithms aim to minimize the $\ell$-1 regularized loss, while the feature selection algorithms are motivated to explicitly address the feature selection issue and thus impose a hard constraint on the number of non-zero elements in classifier. Because of these similarities, they share some common strategies such as truncation and projection.

\subsubsection{Others} Two stochastic methods were proposed in \citep{shalev2011stochastic} for $\ell_1$-regularized loss minimization. The Stochastic Coordinate Descent (SCD) algorithm randomly selects one coordinate from $d$ dimensions and update this single coordinate with the gradient of the total loss of all instances. The Stochastic Mirror Descent Made Sparse (SMIDAS) algorithm combines the idea of truncating the gradient with mirror descent algorithm, i.e., truncation is performed on the vector in dual space. The disadvantage of the two algorithms is that their computational complexity depends on the dimensionality $d$. Besides, the two algorithms are designed in batch learning setting, i.e., they assume all instances are known prior to the learning task. Besides, there are also some recent sparse online learning algorithms proposed \citep{wanghigh,wang2015framework}, which combine the ideas of sparse learning, second order online learning, and cost-sensitive classification together to make the online algorithms scalable for high-dimensional class-imbalanced learning tasks.

\subsection{Online Learning with Kernels}

We now survey a family of ``Kernel-based Online Learning" algorithms for learning a nonlinear target function, where the nonlinearity is induced by kernels. We take the typical nonlinear binary classification task as an example. Our goal is to learn a nonlinear classifier $f: \R^d \rightarrow \R$ from a sequence of labeled instances $(\x_t, y_t), t=1,...,T$, where $\x_t\in \R^d$ and $y_t\in\{+1,-1\}$. We build the classification rule as: $\yh_t=\sign(f(\x_t))$, where $\yh_t$ is the predicted class label. We measure the classification confidence of certain instance $\x_t$ by $|f(\x_t)|$. Similar to the linear case, for an online classification task, one can define the hinge loss function $\ell(\cdot)$ for the $t$-th instance using the classifier at the $t$-th iteration:
$$\ell((\x_t,y_t);f_t) = \max(0,1-y_tf_t(\x_t))$$
Formally speaking, an online nonlinear learner aims to achieve the lowest regret $R(T)$ after time $T$, where the regret function $R(T)$ is defined as follows:
\begin{equation}
R(T)=\sum_{t=1}^T{\ell_t(f_t)} - \inf_{f}{\sum_{t=1}^T{\ell_t(f)}},
\end{equation}
where $\ell_t(\cdot)$ is the loss for the classification of instance $(\x_t,y_t)$, which is short for $\ell((\x_t,y_t);\cdot)$. We denote by $f^*$ the optimal solution of the second term, i.e., $f^*=\arg\min_{f}\sum_{t=1}^T{\ell_t(f)}$

In the following, we first introduce online kernel methods and then survey a family of scalable online kernel learning algorithms organized into two major categories: (i) budget online kernel learning using {\it budget maintenance} strategies and (ii) budget online kernel learning using {\it functional approximation} strategies. Then we briefly introduce some approaches for online learning with multiple kernels. Without loss of generality, we will adopt the above online binary classification setting for the discussions in this section.


\subsubsection{Online Kernel Methods}
We refer to the output $f$ of the learning algorithm as a \emph{hypothesis} and denote the set of all possible hypotheses by $\H=\{f | f:\R^d \rightarrow \R\}$. Here $\H$ a Reproducing Kernel Hilbert Space (\textbf{RKHS}) endowed with a kernel function $\kappa(\cdot, \cdot):\R^d\times\R^d \rightarrow \R$ \citep{vapnik1998statistical} implementing the inner product$\langle\cdot,\cdot\rangle$ such that: 1) $\kappa$ has the reproducing property $\langle f, \kappa(\x,\cdot)\rangle=f(\x)$ for $\x\in \R^d$; 2) $\H$ is the closure of the span of all $\kappa(\x,\cdot)$ with $\x\in \R^d$, that is, $\kappa(\x,\cdot)\in\H$ $\forall\x\in\mathcal{X}$. The inner product $\langle\cdot,\cdot\rangle$ induces a norm on $f\in \H$ in the usual way: $\|f\|_{\H}:=\langle f,f\rangle^{\frac{1}{2}}$. We denote by $\H_\kappa$ an RKHS with explicit dependence on kernel $\kappa$. Throughout the analysis, we assume $\kappa(\x,\x)\le X^2$, $\forall\x\in \R^d$, $X \in \R^+$ is a constant.

The task of training the model of SVM $f(\x)$ in batch is formulated as the optimization:
\begin{eqnarray}
\min_{f\in\Hk} \frac{\lambda}{2}\|f\|_{\H}^2+\frac{1}{T}\sum_{t=1}^T\ell(f(\x_t); y_t)
\end{eqnarray}
where $\lambda>0$ is a regularization parameter used to control model complexity. According to the Representer Theorem \citep{DBLP:conf/colt/ScholkopfHS01}, the optimal solution of the above convex optimization problem lies in the span of $T$ kernels, i.e., those centered on the training points. Consequently, the goal of a typical online kernel learning algorithm is to learn the kernel-based predictive model $f(\x)$ for classifying a new instance $\x \in \R^d$ as follows: $f(\x) = \sum_{t=1}^{T}{\alpha_t \kappa(\x_t,\x)}, $
where $T$ is the number of processed instances, $\alpha_t$ denotes the coefficient of the $t$-th instance, and $\kappa(\cdot,\cdot)$ denotes the kernel function. We define support vector (SV) as  the instance whose coefficient $\alpha$ is nonzero. Thus, we rewrite the previous classifier as $f(\x) = \sum_{i\in\mathcal{SV}}{\alpha_i \kappa(\x_i,\x)}$, where $\mathcal{SV}$ is the set of SV's and $i$ is its index. We use the notation $|\mathcal{SV}|$ to denote the SV set size. In literature, different online kernel methods have been proposed. We begin by introducing the simplest one, that is, the kernelized Perceptron algorithm.

\paragraph{Kernelized Perceptron.} This extends the Perceptron algorithm using the kernel trick. Algorithm \ref{alg:kernelizedPerceptron} outlines the Kernelized Perceptron algorithm \citep{LMP99}.

\begin{algorithm}[hptb]
\label{alg:kernelizedPerceptron}
\caption{Kernelized Perceptron}
\begin{algorithmic}
\STATE \textbf{INIT}: $f_0=0$\\
\FOR{ $t=1,2,\ldots,T$}
    \STATE Given an incoming instance $\mathbf{x}_t$, predict $\hat{y}_t =\sign(f_t(\x_t))$;
    \STATE Receive the true class label $y_t \in \{+1, -1\}$;
    \IF{$\hat{y}_t \neq y_t$}
    \STATE $\mathcal{SV}_{t+1}=\mathcal{SV}_{t}\cup{(\x_t,y_t)}$, $f_{t+1}=f_t+y_t\kappa(\x_t,\cdot)$;
    \ENDIF
\ENDFOR
\end{algorithmic}
\end{algorithm}
The algorithm works similar to the standard Perceptron algorithm, except that the inner product, i.e., $f_t(\x_t)=\sum_i\alpha_i\x_i^\top\x_t$, is replaced by a kernel function in the kernel Percetron, i.e., $f_t(\x_t)=\sum_i\alpha_i \kappa(\x_i^\top,\x_t)$.
\paragraph{Kernelized OGD.} This extends the OGD algorithm with kernels \citep{kivinen2004online}, as shown in Algorithm \ref{alg:kernelOGD}. Here, $\eta_t>0$ is the learning rate parameter, and $\ell'_t$ is used to denote the derivative of loss function with respect to the classification score $f_t(\x_t)$.
\begin{algorithm}[hptb]
\label{alg:kernelOGD}
\caption{Kernelized OGD}
\begin{algorithmic}
\STATE \textbf{INIT}: $f_0=0$\\
\FOR{ $t=1,2,\ldots,T$}
    \STATE Given an incoming instance $\mathbf{x}_t$, predict $\hat{y}_t =\sign(f_t(\x_t))$;
    \STATE Receive the true class label $y_t \in \{+1, -1\}$;
    \IF {$\ell_t(f_t)>0$}
    \STATE
    $\mathcal{SV}_{t+1}=\mathcal{SV}_{t}\cup{(\x_t,y_t)}$, $f_{t+1}=f_t-\eta_t\nabla\ell_t(f_t(\x_t)) = f_t-\eta_t\ell'_t\kappa(\x_t,\cdot)$;
    \ENDIF
\ENDFOR
\end{algorithmic}
\end{algorithm}

\paragraph{Other Related Work.}
The kernel trick implies that the inner product between any two instances can be replaced by a kernel function, i.e., $\kappa(\x_i,\x_j)=\Phi(\x_i)^\top\Phi(\x_j), \forall i,j$, where $\Phi(\x_t)\in R^D$ denotes the feature mapping from the original space to a new $D$-dimensional space which can be infinite. Using the kernel trick, many existing linear online learning algorithms can be easily extended to their kernelized variants,
such as the kernelized Perceptron and kernelized OGD as well as kernel PA variants \citep{crammer2006online}. However, some algorithms that use complex update rules are non-trivial to be converted into kernelized versions, such as Confidence-Weighted algorithms \citep{dredze2008confidence}. Moreover, some online kernel learning methods also attempt to make more effective updates at each iteration. For example, Double Updating Online Learning (DUOL) \citep{zhao2009duol,zhao2011double,zhao2012bduol} improves the efficacy of traditional online kernel learning methods by not only updating the weight of the newly added SV, but also the weight for one existing SV. Finally, we note one major challenge of online kernel method is the computational efficiency and scalability due to the curse of kernelization \citep{wang2012breaking}. In the following, we will discuss two types of techniques to scale up kernel-based online learning methods.


\subsubsection{Scalable Online Kernel Learning via Budget Maintenance}
Despite enjoying the clear advantage of accuracy performance over linear models, online kernel learning falls short in some critical drawbacks, in which one critical issue is the growing unbounded number of support vectors with increasing computational and space complexity over time. To address this challenge, a family of algorithms, termed ``budget online kernel learning", have been proposed to bound the number of SV's with a fixed budget $B=|\mathcal{SV}|$ using diverse budget maintenance strategies whenever the budget overflows. The general framework for budgeting strategies is shown in Algorithm \ref{alg:budgetKernel}. Most existing budget online kernel methods maintain the budget by three strategies: (i) SV Removal, (ii) SV Projection, and (iii) SV Merging. We briefly review each of them below.

\begin{algorithm}[hptb]
	\label{alg:budgetKernel}
	\caption{Budget Online Kernel Learning}
	\begin{algorithmic}
		\STATE \textbf{INIT}: $f_0=0$\\
		\FOR{ $t=1,2,\ldots,T$}
		\STATE Given an incoming instance $\mathbf{x}_t$, predict $\hat{y}_t =\sign(f_t(\x_t))$;
		\STATE Receive the true class label $y_t \in \{+1, -1\}$;
		\IF {update is needed}
		\STATE update the classifier from $f_t$ to $f_{t+\frac{1}{2}}$and $\mathcal{SV}_{t+\frac{1}{2}}=\mathcal{SV}_{t}\cup{(\x_t,y_t)}$
		\ENDIF
		\IF {$|\mathcal{SV}_{t+\frac{1}{2}}|>B$}
		\STATE Update Support Vector Set from $\mathcal{SV}_{t+\frac{1}{2}}$ to $\mathcal{SV}_{t+1}$ such that $|\mathcal{SV}_{t+1}| = B$ \\
		\STATE Update the classifier from $f_{t+\frac{1}{2}}$ to $f_{t+1}$
		\ENDIF
		\ENDFOR
	\end{algorithmic}
\end{algorithm}

\paragraph{SV Removal.}
This strategy maintains the budget by a simple and efficient way: 1) update the classifier by adding a new SV whenever necessary (depending on the prediction mistake/loss); 2) if the SV size exceeds the budget, discard one of existing SV's and update the classifier accordingly.
To achieve this, we need to address the following concerns: (i) how to update the classifier  and (ii) how to choose one of existing SV's for removal.

The first step of updating classifiers depends on which online learning method is used. For example, the Perceptron algorithm has been used in RBP \citep{cavallanti2007tracking}, Forgetron \citep{DBLP:conf/nips/DekelSS05}, and Budget Perceptron \citep{crammer2003online}. The OGD algorithm has been adopted by BOGD \citep{DBLP:conf/icml/HoiWZJW12} and BSGD+ removal \citep{wang2012breaking}, while PA has been used by BPA-S \citep{DBLP:journals/jmlr/WangV10}.

The second step of SV removal is to find one of existing SV's, denoted as $(\x_{del},y_t)$, to be removed by minimizing the impact of the resulting classifier. One simple way is to randomly discard one of existing SV's uniformly with probability $\frac{1}{B}$, as adopted by RBP \citep{cavallanti2007tracking} and BOGD \citep{DBLP:conf/icml/HoiWZJW12}. Instead of choosing randomly, another way as used in ``Forgetron"~\citep{DBLP:conf/nips/DekelSS05} is to discard the oldest SV by assuming an older SV is less representative for the distribution of fresh training data streams. Despite enjoying the merits of simplicity and high efficiency, these methods are often too simple to achieve satisfactory learning results.

To optimize the performance, some approaches have tried to perform exhaustive search in deciding the best SV for removal. For instance, the Budget Perceptron algorithm \citep{crammer2003online} searches for one SV that is classified with high confidence by the classifier:
$$y_{del}(f_{t+\frac{1}{2}}(\x_{del})-\alpha_{del}\kappa(\x_{del},\x_{del}))>\beta$$
where $\beta>0$ is a fixed tolerance parameter. BPA-S shares the similar idea of exhaustive search. For every $r\in [B]$, a candidate classifier $f^r=f_{t+\frac{1}{2}}-\alpha_r\kappa(\x_r,\cdot)$ is generated by discarding the $r$-th SV from $f_{t+\frac{1}{2}}$. By comparing the $B$ candidate classifiers, the algorithm selects the one that minimizes the current objective function of PA:
$$f_{t+1}=\argmin_{r\in [B]}\frac{1}{2}||f^r-f_t||_\H^2+C\ell_t(f^r)$$
where $C>0$ is the regularization parameter of PA to balance aggressiveness and passiveness. Comparing the principles of different SV removal strategies, we observe that a simple rule may not always generate satisfactory accuracy, while an exhaustive search often incurs non-trivial computational overhead, which again may limit the application to large-scale problems. When deploying a solution in practice, one would need to balance the trade-off between effectiveness and efficiency.

\paragraph{SV Projection.}
SV Projection strategy first appeared in \citep{orabona2009bounded} where two new algorithms, Projectron and Projectron++, were proposed, which significantly outperformed the previous SV removal based algorithms such RBP and Forgetron. The SV projection method follows the setting of SV removal and identifies a support vector for removal during the update of the model.
It then chooses a subset of $\mathcal{SV}$ as the projection base, which will be denoted by $\mathcal{P}$. Following this, a linear combination of kernels in $\mathcal{P}$ is used to approximate the removed SV. The procedure of finding the optimal linear combination can be formulated as a convex optimization of minimizing the projection error:
$$\mathbf{\beta}=\argmin_{\beta\in \R^|\mathcal{P}|}E_{proj}=\argmin_{\beta\in \R^|\mathcal{P}|}||\alpha_{del}\kappa(\x_{del},\cdot)-\sum_{i\in\mathcal{P}}\beta_i\kappa(\x_i,\cdot)||^2_{\H}$$\\
Finally, the classifier is updated by combining this result with the original classifier:
$$f_{t+1}=f_{t+\frac{1}{2}}-\alpha_{del}\kappa(\x_{del},\cdot)+\sum_{i\in\mathcal{P}}\beta_i\kappa(\x_i,\cdot)$$

There are several algorithms adopting the projection strategy, for example Projectron, Projectron++, BPA-P, BPA-NN \citep{DBLP:journals/jmlr/WangV10} and BSGD+Project \citep{wang2012breaking}. These methods differ in a few aspects. First, the update rules are based on different online learning algorithms. Generally speaking, PA based and OGD based tend to outperform Perceptron based algorithms because of their effective update. Second, the choice of discarded SV is different. Since projection itself is relative slow,  exhaustive search based algorithms (BPA-NN, BPA-P) are extremely time consuming. Thus algorithms with simple selecting rules are prefered (Projectron, Projectron++, BSGD+Project). Third, the choice of projection base set $\mathcal{P}$ is different. In Projectron, Projectron++, BPA-P and BSGD+Project, the discarded SV is projected onto the whole SV set, i.e. $\mathcal{P}=\mathcal{SV}$. While in BPA-NN, $\mathcal{P}$ is only a small subset of $\mathcal{SV}$, made up of the nearest neighbors of the discarded SV $(\x_{del},y_{del})$. In general, a larger projection base set implies a more complicated optimization problem and thus more time costs. The research direction of SV projection based budget learning is to find a proper way of selecting $\mathcal{P}$ so that the algorithm achieves the minimized projection error with a relative small projection base set.

\paragraph{SV Merging.}
\cite{wang2012breaking} proposed a SV merging method called ``BSGD+Merge" which replaces the sum of two SV's $\alpha_m\kappa(\x_m,\cdot)+\alpha_n\kappa(\x_n,\cdot)$ by a newly created SV $\alpha_z\kappa(\mathbf{z},\cdot)$, where $\alpha_m$, $\alpha_n$ and $\alpha_z$ are the corresponding coefficients of $\x_m$, $\x_n$ and $\mathbf{z}$. Following the previous discussion, the goal of online budget learning through SV merging strategy is to find the optimal $\alpha_z\in \R$ and $\mathbf{z}\in\R^d$ that minimizes the gap between $f_{t+1}$ and $f_{t+\frac{1}{2}}$.

As it is relatively complicated to optimize the two terms simultaneously, the optimization is divided into two steps. First, assuming the coefficient of $\mathbf{z}$ is $\alpha_m+\alpha_n$, this algorithm tries to create the optimal SV that minimizes the merging error as follows
$$\min_\mathbf{z} ||(\alpha_m+\alpha_n)\kappa(\mathbf{z},\cdot)-(\alpha_m\kappa(\x_m,\cdot)+\alpha_n\kappa(\x_n,\cdot))||$$
The solution is $\mathbf{z}=h\x_m+(1-h)x_n$, where $0<h<1$ is a real number that can be found by a line search method. This solution indicates that the optimal created SV lies on the line connecting $\x_m$ and $\x_n$. After obtaining the optimal created SV $\mathbf{z}$, the next step is to find the optimal coefficient $\alpha_z$, which can be formulated as $$\min_{\alpha_z} ||(\alpha_z\kappa(\mathbf{z},\cdot)-(\alpha_m\kappa(\x_m,\cdot)+\alpha_n\kappa(\x_n,\cdot))||.$$ The solution becomes $\alpha_z=\alpha_m\kappa(\x_m,\mathbf{z})+\alpha_n\kappa(\x_n,\mathbf{z})$. The remaining problem is which two SV's $\x_m$ and $\x_n$ should be merged. The ideal solution is to find the optimal pair with the minimal merging error through exhaustive search, which however requires $O(B^2)$ time complexity. How to find the optimal pair efficiently remains an open challenge.

\paragraph{Summary.} Among various algorithms of budget online kernel learning using budget maintenance, the key differences are their updating rules and budget maintenance strategies.
Table~\ref{table:algorithm} gives a summary of different algorithms and their properties.
\begin{table*}[htb]
\scriptsize
\begin{center}
\caption{Comparisons of different budget online kernel learning algorithms.\label{table:algorithm}}{
\begin{tabular}{lllll}        \hline
Algorithms &Update Strategy& Budget Strategy & Update Time & Space\\
\hline
Stoptron \citep{orabona2009bounded}&Perceptron&Stop&$O(1)$&$O(B)$\\
Tighter Perceptron \citep{weston2005online}&Perceptron&Removal&$O(B^2)$&$O(B)$\\
Tightest Perceptron \citep{wang2009tighter}&Perceptron&Removal&$O(B^2)$&$O(B)$\\
Budget Perceptron \citep{crammer2003online}&Perceptron&Removal&$O(B^2)$&$O(B)$\\
RBP \citep{cavallanti2007tracking}&Perceptron&Removal&$O(B)$&$O(B)$\\
Forgetron\citep{DBLP:conf/nips/DekelSS05}&Perceptron&Removal&$O(B)$&$O(B)$\\
BOGD\citep{DBLP:conf/icml/HoiWZJW12}&OGD&Removal&$O(B)$&$O(B)$\\
BPA-S \citep{DBLP:journals/jmlr/WangV10}&PA& Removal&$O(B)$&$O(B)$\\
BSGD+removal \citep{wang2012breaking}&OGD& Removal& $O(B)$&$O(B)$\\
Projectron \citep{orabona2009bounded}&Perceptron& Projection&$O(B^2)$&$O(B^2)$\\
Projectron++ \citep{orabona2009bounded}&Perceptron& Projection&$O(B^2)$&$O(B^2)$\\
BPA-P \citep{DBLP:journals/jmlr/WangV10}&PA& Projection &$O(B^3)$&$O(B^2)$\\
BPA-NN \citep{DBLP:journals/jmlr/WangV10}&PA& Projection&$O(B)$&$O(B)$\\
BSGD+projection \citep{wang2012breaking}&OGD&Projection&$O(B^2)$&$O(B^2)$\\
BSGD+merging \citep{wang2012breaking}&OGD&Merging&$O(B)$&$O(B)$\\
 \hline
\end{tabular}
}
\vspace{-0.15in}
\end{center}
\end{table*}
In addition to the previous budget kernel learning algorithms, there are also some other works in online kernel learning. For example, some studies \citep{LZH16,zhang2012efficient} introduce the idea of sparse kernel learning to reduce the number of SV's in the online-to-batch-conversion problem, where an online algorithm can be used to train a kernel model efficiently for the batch setting (See Section \ref{sec:onlineToBatch}).

\subsubsection{Scalable Online Kernel Learning via Functional Approximation}

In contrast to the previous budget online kernel learning methods using budget maintenance strategies to guarantee efficiency and scalability, another emerging and promising strategy is to explore functional approximation techniques for achieving scalable online kernel learning \citep{wang2013large,Lu2015}.

The key idea is to construct a kernel-induced feature representation $\z(\x)\in\R^D$ such that the inner product of instances in the new feature space can effectively approximate the kernel function:
$$\kappa(\x_i, \x_j) \approx  \z(\x_i)^{\top} \z(\x_j)$$
Using the above approximation, the predictive model with kernels can be rewritten as follows:
$$f(\x) = \sum_{i=1}^{B}{\alpha_i \kappa(\x_i,\x)} \approx
\sum_{i=1}^{B}{\alpha_i \z(\x_i)^{\top} \z(\x)} = \w^{\top} \z(\x)$$
where $\w =  \sum_{i=1}^{B}{\alpha_i \z(\x_i)}$ denotes the weight vector to be learned in the new feature space.

As a consequence, solving a regular online kernel classification task can be turned into a linear online classification task on the new feature space derived from the kernel approximation. For example, the methods of online kernel learning with kernel approximation in \citep{wang2013large,Lu2015} integrate some existing online learning algorithms (e.g., OGD) with kernel approximation techniques \citep{lin2011efficient,sonnenburg2010coffin,chang2010training} to derive scalable online kernel learning algorithms, including Fourier Online Gradient Descent (FOGD) that explores random Fourier features for kernel functional approximation \citep{rahimi2007random}, and Nystr\"{o}m Online Gradient Descent (NOGD) that explores Nystr\"{o}m low-rank matrix approximation methods for approximating large-scale kernel matrix \citep{NIPS2000_1866}. A recent work, Dual Space Gradient Descent \citep{le2016dual,nguyen2017large} updates the model as the RBP algorithm, but also builds an FOGD model using the discarded SV's. The final prediction is the combination of the two models.

\subsubsection{Online Multiple Kernel Learning}
Traditional online kernel methods usually assume a predefined good kernel is given prior to the online learning task. Such approaches could be restricted since it is often hard to choose a good kernel prior to the learning task. To overcome the drawback, Online Multiple Kernel Learning (OMKL) aims to combine multiple kernels automatically for online learning tasks without fixing any predefined kernel. In the following, we begin by introducing some basics of batch Multiple Kernel Learning (MKL) \citep{LSMKL06}.

Given a training set $\mathcal{D} = \{(\x_t, y_t), t=1, \ldots, T\}$ where $\x_t\in\mathbb{R}^d$, $y_t \in \{-1, +1\}$, and a set of $m$ kernel functions $\mathcal{K}=\{\kappa_i:\mathcal{X}\times\mathcal{X} \rightarrow \mathbb{R}, i=1, \ldots, m\}$. MKL learns a kernel-based prediction function by identifying an optimal combination of the $m$ kernels, denoted by $\theta = (\theta_1,
\ldots, \theta_m)$, to minimize the margin-based classification error, which can be cast into the optimization below:
\begin{eqnarray}
\min\limits_{\theta \in \Delta} \; \min\limits_{f \in
\mathcal{H}_{K(\theta)}} \frac{1}{2}|f|^2_{\mathcal{H}_{K(\theta)}}
+ C\sum_{t=1}^n \ell(f(\x_t), y_t)
\label{eqn:opt-mkl-batch}
\end{eqnarray}
where $\Delta = \{ \theta \in \mathbb{R}_+^m | \theta^{\top}\mathbf{1}_m = 1\},\;
K(\theta)(\cdot, \cdot) = \sum_{i=1}^m \theta_i \kappa_i(\cdot,
\cdot), \;\ell(f(\x_t), y_t) = \max(0, 1 - y_tf(\x_t)).
$
In the above, we use notation $\mathbf{1}_T$ to represent a vector of $T$ dimensions with all its elements being $1$. It can
also be cast into the following mini-max optimization problem:
\begin{eqnarray}
\min\limits_{\theta \in \Delta} \max\limits_{\alpha \in \Xi}
\left\{\alpha^{\top}\mathbf{1}_T - \frac{1}{2}(\alpha \circ
\mathbf{y})^{\top}\left(\sum_{i=1}^m \theta_iK^i \right)(\alpha
\circ \mathbf{y}) \right\} \label{eqn:kernel-learning}
\end{eqnarray}
where $K^i \in \mathbb{R}^{T\times T}$ with $K^i_{j,l} = \kappa_i(\x_j, \x_l)$, $\Xi = \{\alpha | \alpha \in [0, C]^T\}$, and
$\circ$ defines the element-wise product between two vectors. The above batch MKL optimization has been extensively studied \citep{simpleMKL,xu-2008-level}, but an efficient solution remains an open challenge.

Some efforts of online MKL studies~\citep{OM2,DBLP:journals/jmlr/MartinsSXAF11} have attempted to solve batch MKL optimization via online learning. Unlike these approaches that are mainly concerned in optimizing the optimal kernel combination as regular MKL, another framework of {\it Online Multiple Kernel Learning} (OMKL) \citep{DBLP:conf/alt/JinHY10,DBLP:journals/ml/HoiJZY13,yang2012online} is focused on exploring effective online combination of multiple kernel classifiers via a significantly more efficient and scalable way. Specifically, the OMKL in \citep{DBLP:conf/alt/JinHY10,DBLP:journals/ml/HoiJZY13} learns a kernel-based prediction function by selecting a subset of predefined kernel functions in an online learning fashion, which is in general more challenging than typical online learning because both the kernel classifiers and the subset of selected kernels are unknown, and more importantly the solutions to the kernel classifiers and their combination weights are correlated. \citep{DBLP:journals/ml/HoiJZY13} proposed novel algorithms based on the fusion of two types of online learning algorithms, i.e., the {\it Perceptron} algorithm that learns a classifier for a given kernel, and the {\it Hedge} algorithm \citep{DBLP:journals/jcss/FreundS97} that combines classifiers by linear weights. Some stochastic selection strategies were also proposed by randomly selecting a subset of kernels for combination and model updating to further improve the efficiency. These methods were later extended for regression \citep{sahoo2014online}, learning from data with time-sensitive patterns  \citep{sahootemporal} and imbalanced data streams\citep{sahoo2016cost}. In addition, there have been budgeting approaches to make OMKL scalable \citep{lu2018sparse}.

\subsection{Online to Batch Conversion}
\label{sec:onlineToBatch}

Let us denote by $\A$ an online learning algorithm for the purpose of training a binary classifier from a sequence of training examples. On each round, the algorithm receives an instance $\mathbf{x}_t\in\mathbb{R}^d$, the algorithm chooses a vector $\w_t\in\S\subseteq\R^d$ to predict the class label of the instance, i.e., $\hat{y_t}=\sign(\w_t^{\top}\x_t)$. After that, the environment responds by disclosing the true label $y_t$ and some convex loss function $\ell(\w;(\x_t,y_t)$, and the algorithm suffers a loss $\ell_t(\w_t)$ at the end of this round. For such a setting, consider a sequence of $T$ rounds $(\x_1,y_1),\ldots,(\x_T,y_T)$, the online algorithm aims to minimize the following regret
\bqs
\Reg_\A(T)=\sum^T_{t=1}\ell(\w_t;(\x_t,y_t))-\min_{\w\in\S}\sum^T_{t=1}\ell(\w;(\x_t,y_t))
\eqs
However, for batch training setting, we are more interested in finding a model $\hw$ with good generalization ability, i.e., we want to achieve a small excess risk defined as
\bqs
R(\hw)-\min_{\w\in\S}R(\w)
\eqs
where the generalization risk is $R(\w)=\E_{(\x,y)}[\ell(w;(x,y))]$, and $(\x,y)$ satisfies a fixed unknown distribution.
Therefore, we would like to study the generalization performance of online algorithms through the Online to Batch Conversion \citep{DBLP:journals/tit/Cesa-BianchiCG04,wang2012generalization,kakade2009generalization,levy2017online}, where the conversion relates the regret of the online algorithm to its generalization performance.

\subsubsection{A General Conversion Theory}

We now consider the generalization ability of online learning under the assumption that the loss $\ell(\w;(\x,y))$ is strongly convex, which is reasonable as many loss functions (e.g., square loss) are strongly convex, and even if some loss (e.g., hinge loss) is not strongly convex, we can impose some regularization term (e.g., $\frac{1}{2}\|\cdot\|$) to achieve strong convexity. We denote the dual norm of $\|\cdot\|$ as $\|\cdot\|_*$, where $\|\v\|_*=\sup_{\|\w\|\le 1}\v^\top\w$. Let $Z=(\x,y)$ be a random variable taking values in some space $\Z$. Our goal is to minimize $R(\w)=\E_Z[\ell(\w;Z)]$ over $\w\in\S$. Specifically, we assume the loss $\ell: \S\times\Z\rightarrow [0, B]$ satisfies the following assumption:

{\bf Assumption LIST}(LIpschitz and STrongly convex assumption) For all $z\in\Z$, the function $\ell_z(\w)=\ell(\w;z)$ is convex in $\w$ and satisfies:
\begin{enumerate}
	\item $\ell_z$ has Lipschitz constant $L$ w.r.t. the norm $\|\cdot\|$, i.e., $|\ell_z(\w)-\ell_z(\w')\le L\|\w-\w'\|$.
	
	\item $\ell_z$ is $\lambda$-strongly convex w.r.t. $\|\cdot\|$, i.e., $\forall\theta\in[0,1]$, $\forall\w,\w'\in\S$,
	\bqs\begin{aligned}
		&\ell_z(\theta\w+(1-\theta)\w')  \le&\theta\ell_z(\w)+(1-\theta)\ell_z(\w')-\frac{\lambda}{2}\theta(1-\theta)\|\w-\w'\|^2.
	\end{aligned}\eqs
	
\end{enumerate}
For this kind of loss function, we consider an online learning setting where $Z_1,\ldots,Z_T$ are given sequentially in i.i.d. We then have
\bqs
\E[\ell(\w;Z_t)]=\E[\ell(\w,(\x_t,y_t))]:=R(\w),\ \forall t,\ \w\in\S.
\eqs
Now consider an online learning algorithm $\A$, which is initialized as $\w_1$. Whenever $Z_t$ is given, model $\w_t$ is updated to $\w_{t+1}$. Let $\E_t[\cdot]$ denote conditional expectation w.r.t. $Z_1,\ldots,Z_t$, we have $\E_t[\ell(\w_t;Z_t)]=R(\w_t)$. Using the above assumptions and the Freedman's inequality leads to the following theorem for the generalization ability of online learning
\begin{thm}(\cite{DBLP:journals/tit/Cesa-BianchiCG04})
	Under the assumption LIST, we have the following inequality, with probability at least $1-4\delta\ln T$,
	\bqs\begin{aligned}
		&\frac{1}{T}\sum^T_{t=1}R(\w_t)-R(\w_*)\le \frac{\Reg_\A(T)}{T} + &4\sqrt{\frac{L^2\ln\frac{1}{\delta}}{\lambda}}\frac{\sqrt{\Reg_\A(T)}}{T}+\max\big(\frac{16L^2}{\lambda},6B\big)\frac{\ln\frac{1}{\delta}}{T},
	\end{aligned}
	\eqs
	where $\w_*=\arg\min_{\w\in\S}R(\w)$. Further, using Jensen's inequality, $\frac{1}{T}\sum_t R(\w_t)$ can be replaced by $R(\bar{\w}_T)$ where $\bar{\w}_T=\frac{1}{T}\sum_t\w_t$.
\end{thm}

If the assumption LIST is satisfied by $\ell_z(\w)$, then the Online Gradient Descent (OGD) algorithm that generates $\w_1,\ldots,\w_T$ has the following regret
$ \Reg_\A(T) \le \frac{L^2}{2\lambda}(1+\ln T).$ Plugging this inequality back into the theorem and using $(1+\ln T)/(2T)\le \ln T/ T,\ \forall T\ge 3$ gives the following Corollary.
\begin{corollary}
Suppose assumption $LIST$ holds for $\ell_z(\w)$. Then the Online Gradient Descent (OGD) algorithm that generates $\w_1,\ldots,\w_T$ and finally outputs $\bar{\w}_T=\frac{1}{T}\sum_t\w_t$, satisfies the following inequality for its generalization ability, with probability at least $1-4\delta\ln T$,
	\bqs
	R(\bar{\w}_T)-R(\w_*)\le \frac{L^2\ln T}{\lambda T} + \sqrt{\ln\frac{1}{\delta}}\frac{4L^2\sqrt{\ln T}}{\lambda T}+\max\big(\frac{16L^2}{\lambda}, 6B\big)\frac{\ln\frac{1}{\delta}}{T},
	\eqs
	for any $T\ge 3$, where $\w_*=\arg\min_{\w\in\S}R(\w)$.
\end{corollary}

\subsubsection{Other Conversion Theories}
Online to batch conversion has been studied in literature \citep{DBLP:conf/colt/Littlestone89,DBLP:journals/tit/Cesa-BianchiCG04,DBLP:conf/colt/Zhang05,DBLP:journals/tit/Cesa-BianchiG08,wang2012generalization}. For general convex loss functions, \cite{DBLP:journals/tit/Cesa-BianchiCG04} proved the following generalization ability of online learning algorithm with probability at least $1-\delta$
\begin{eqnarray}
\hspace{-0.3in}R(\bar{\w}_T)\le \frac{1}{T}\sum^T_{t=1}\ell(\w_t;z_t)+\sqrt{\frac{2}{T}\ln\frac{1}{\delta}}= \frac{\Reg_\A(T)}{T}+\min_{\w\in\S}\frac{1}{T}\sum^T_{t=1}\ell(\w;z_t)+\sqrt{\frac{2}{T}\ln\frac{1}{\delta}},\nonumber
\end{eqnarray}
where the loss $\ell\leq 1$. \cite{DBLP:conf/colt/Zhang05} is another work that explicitly goes by the exponential moment method to drive sharper concentration results. In addition, \cite{DBLP:journals/tit/Cesa-BianchiG08} improved their initial generalization bounds using Bernstein's inequality by assuming $\ell(\cdot)\le 1$, and proves the following inequality with probability at least $1-\delta$
\bqs\begin{aligned}
	&R(\hw)\le \frac{1}{T}\sum^T_{t=1}\ell(\w_t;z_t)+O\left(\frac{\ln(T^2/\delta)}{T}+\sqrt{\frac{1}{T}\sum^T_{t=1}\ell(\w_t;z_t)\frac{\ln (T^2/\delta)}{T}}\right).
\end{aligned}\eqs
where $\hw$ is selected from $\w_1,\ldots,\w_T$, by minimizing a specifically designed penalized empirical risk. In particular, the generalization risk converges to $\frac{1}{T}\sum^T_{t=1}\ell(\w_t;z_t)$ at rate $O(\sqrt{\ln T^2/T})$ and vanishes at rate $O(\ln T^2/T)$ whenever the loss $\sum^T_{t=1}\ell(\w_t;z_t)$ is $O(1)$.



\section{Applied Online Learning for Supervised Learning}
\subsection{Overview}

In this section, we survey the most representative algorithms for a group of non-traditional online learning tasks, wherein the supervised online algorithms cannot be used directly. These algorithms are motivated by new problem settings and applications which follow the traditional online setting, where the data arrives in a sequential manner. However, there was a need to develop new algorithms which were suited to these scenarios. Our review includes
cost-sensitive online learning, online multi-task learning, online multi-view learning,  online transfer learning, online metric learning, online collaborative filtering, online learning structured prediction, distributed online learning, online learning with neural networks, and online portfolio selection.

\subsection{Cost-Sensitive Online Learning}

In a supervised classification task, traditional online learning methods are often designed to optimize mistake rate or equivalently classification accuracy. However, it is well-known that classification accuracy becomes a misleading metric when dealing with class-imbalanced data which is common for many real-world applications, such as anomaly detection, fraud detection, intrusion detection, etc. To address this issue, cost-sensitive online learning \citep{chen2017cstg} represents a family of online learning algorithms that are designed to take care of different misclassification costs of different classes in a class-imbalanced classification task. Next, we briefly survey these algorithms.



\paragraph{Perceptron Algorithms with Uneven Margin (PAUM)}

PAUM \citep{li2002perceptron} is a cost-sensitive extension of Perceptron \citep{rosenblatt1958perceptron} and the Perceptron with Margins (PAM) algorithms \citep{krauth1987learning}. Perceptron makes an update only when there is a mistake, while PAM tends to make more aggressive updates by checking the margin instead of mistake. PAM makes an update whenever $y_t\w_t^\top\x_t\leq\tau$, where $\tau\in\R^+$ is a fixed parameter controlling the aggressiveness. To deal with class imbalance, PAUM extends PAM via an uneven margin setting, i.e., employing different margin parameters for the two classes: $\tau_+$ and $\tau_-$. Consequently, the update becomes $y_t\w_t^\top\x_t\leq\tau_{y_t}$. By properly adjusting the two parameters, PAUM achieves cost-sensitive updating effects for different classes. One of major limitations with PAUM is that it does not directly optimize a predefined cost-sensitive measure, thus, it does not fully resolve the cost-sensitive challenge.


\paragraph{Cost-sensitive Passive Aggressive (CPA)}
CPA \citep{crammer2006online} was proposed as a cost-sensitive variant of the PA algorithms. It was originally designed for multi-class classification by the following prediction rule:
$\yh_t={\arg\max}_y(\w_t\Phi(\x_t,y))$, where $\Phi$ is a feature mapping function that maps $\x_t$ to a new feature according to the class $y$. For simplicity, we restrict the discussion on the binary classification setting. Using $\Phi(\x,y)=\frac{1}{2}y\x$, we will map the formulas to our setting. The prediction rule is: $\yh_t=\sign(\w_t^\top\x_t)$.
We define the cost-sensitive loss as $$\ell(\w,\x,y)=\w\cdot\Phi(\x,\yh)-\w\cdot\Phi(\x,y)+\sqrt{\rho(y,\yh)},$$
where $\rho(y_1,y_2)$ is the function define to distinguish the different cost of different kind misclassifications and we have assumed $\rho(y,y)=0$. When being converted to binary setting, the loss becomes
$$\ell(\w,\x,y)=
\begin{cases}
0& y_t=\yh\\
|\w^\top\x|+\sqrt{\rho(y,\yh)}&y_t\neq\yh
\end{cases}$$
The mistake depends on the prediction confidence and the loss type. We omit the detailed update steps since it follows the similar optimization as PA learning as discussed before.
Similar to PAUM, this algorithm also is limited in that it does not optimize a cost-sensitive measure directly.

\paragraph{Cost-Sensitive Online Gradient Descent (CSOGD)}
Unlike traditional OGD algorithms that often optimize accuracy, CSOGD \citep{wang2014cost,wang2012cost,zhao2015cost} applies OGD to directly optimize two cost-sensitive measures:
\begin{itemize}
\item[(1)] maximizing the weighted sum of $sensitivity$ and $specificity$, i.e, $sum = \eta_p \times sensitivity + \eta_n \times specificity$, where the two weights satisfy $0 \leq \eta_p,\eta_n \leq 1$ and $\eta_p+\eta_n=1$.
\item[(2)] minimizing the weighted $misclassification~cost$, i.e., $cost = c_p \times M_p + c_n \times M_n$, where $M_p$ and $M_n$ are the number of false negatives and false positives respectively,  $0 \leq c_p,c_n \leq 1$ are the cost parameters for positive and negative classes, respectively, and we assume $c_p+c_n=1$.
\end{itemize}
The objectives can be equivalently reformulated into the following objective:
$$\sum_{y_t=+1}\rho\I_{(y_t\w\cdot\x_t<0)}+\sum_{y_t=-1}\I_{(y_t\w\cdot\x_t<0)}$$
where we set $\rho = \frac{\eta_p T_n}{\eta_n T_p}$ when maximizing the weighted sum, $T_p$ and $T_n$ are the number of positive and negative instances respectively; when minimizing the weighted misclassification cost, we instead set $\rho = \frac{c_p}{c_n}$. The objective is however non-convex, making it hard to optimize directly.
Instead of directly optimizing the non-convex objective, we attempt to optimize a convex surrogate. Specifically, we replace the indicator function $\mathbf{I}_{(\cdot)}$ by a convex surrogate, and attempt to optimize either one of the following modified hinge-loss functions at each online iteration:
$$\ell^I(\w;(\x,y))=\max(0,\rho*\I_{(y=1)}+\I_{(y=-1)}-y(\w\cdot \x))~~~$$
$$~~~~~~~\ell^{II}(\w;(\x,y))=(\rho*\I_{(y=1)}+\I_{(y=-1)})*\max(0,1-y(\w\cdot \x))$$
One can then derive cost-sensitive ODG (CSOGD) algorithms by applying OGD to optimize either one of the above loss functions. The detailed algorithms can be found in  \citep{wang2014cost}. Two recent works extend the problem setting to cost-sensitive classification of multi-class problem \citep{wang2016dealing,zhangonline}. Further there are efforts to do cost-sensitive online learning with kernels \citep{zhao2013costactive,hu2015kernelized}.

\paragraph{Online AUC Maximization}
Instead of optimizing accuracy, some online learning studies have attempted to directly optimize the Area Under the ROC curve (AUC), i.e.,
\bqs
&\text{AUC}(\w)=\frac{\sum_{i=1}^{T_+}\sum_{j=1}^{T_-}\mathbb{I}_{\w\cdot\x_i^+>\w\cdot\x_j^-}}{T_+T_-} =1-\frac{\sum_{i=1}^{T_+}\sum_{j=1}^{T_-}\mathbb{I}_{\w\cdot\x_i^+\leq\w\cdot\x_j^-}}{T_+T_-}
\eqs
where $\x^+$ is a positive instance, $\x^-$ is a negative instance, $T_+$ is the total number of positive instances and $T_-$ is the total number of negative instances. AUC measures the probability for a randomly drawn positive instance to have a higher decision value than a randomly sampled negative instance, and it is widely used in many applications. Optimizing AUC online is however very challenging.

First of all, in the objective, the term $\sum_{i=1}^{T_+}\sum_{j=1}^{T_-}\mathbb{I}_{\w\cdot\x_i^+\leq\w\cdot\x_j^-}$ is non-convex. A common way is to replace the indicator function by a convex surrogate, e.g., a hinge loss function
$$\ell(\w,\x_i^+-\x_j^-)=\max\{0,1-\w(\x_i^+-\x_j^-)\}$$
Consequently, the goal of online AUC maximization is equivalent to minimizing the accumulated loss $\mathcal{L}_t(\w)$ over all previous iterations, where the loss at the $t$-th iteration is 
\bqs\mathcal{L}_t(\w)=\mathbb{I}_{y_t=1}\sum_{\tau=1}^{t-1}\mathbb{I}_{y_\tau=-1}\ell(\w,\x_t-\x_\tau)+\mathbb{I}_{y_t=-1}\sum_{\tau=1}^{t-1}\mathbb{I}_{y_\tau=1}\ell(\w,\x_\tau-\x_t)\eqs
The above takes the sum of the pairwise hinge loss between the current instance $(\x_t,y_t)$ and all the received instances with the opposite class $-y_t$. Despite being convex, it is however impractical to directly optimize the above objective in online setting since one would need to store all the received instances and thus lead to growing computation and memory cost.


The Online AUC Maximization method in \citep{zhao2011online} proposed a novel idea of exploring {\it reservoir sampling} techniques to maintain two buffers, $B_+$ and $B_-$ of size $N_+$ and $N_-$, which aim to store a sketch of historical instances. Specifically, when receiving instance $(\x_t,y_t)$, it will be added to buffer $B_{y_t}$ whenever it is not full, i.e. $|B_{y_t}|<N_{y_t}$. Otherwise, $\x_t$  randomly replaces one instance in the buffer with probability $\frac{N_{y_t}}{N_{y_t}^{t+1}}$, where $N_{y_t}^{t+1}$ is the total number of instances with class $y_t$ received so far. Reservoir sampling is able to guarantee the instances in the buffers maintain an unbiased sampling of the original full dataset. As a result, the loss $\mathcal{L}_t(\w)$ can be approximated by only considering the instances in the buffers, and the classifier $\w$ can be updated by either OGD or PA algorithms.

{\it Others.} To improve the study in \citep{zhao2011online}, a number of following studies have attempted to make improvements from different aspects \citep{YiDing17}. For example, the study in \citep{wang2012generalization} generalized online AUC maximization as online learning with general pairwise loss functions, and offered new generalization bounds for online AUC maximization algorithms similar to \citep{zhao2011online}. The bounds were further improved by \citep{kar2013generalization} which employs a generic decoupling technique to provide Rademacher complexity-based generalization bounds. In addition, the work in \citep{gao2013one} overcomes the buffering storage cost by developing a regression-based algorithm which only needs to maintain the first and second-order statistics of training data in memory, making the resulting storage requirement independent from the training size. The very recent work in \citep{ding2015adaptive} presented a new second-order AUC maximization method by improving the convergence using the adaptive gradient algorithm. The stochastic online AUC maximization (SOLAM) algorithm \citep{ying2016stochastic} formulates the online AUC maximization as a stochastic saddle point problem and greatly reduces the memory cost.

\subsection{Online Multi-task Learning}

Multi-task Learning \citep{caruana1998multitask} is an approach that learns a group of related machine learning tasks together. By considering the relationship between different tasks, multi-task learning algorithms are expected to achieve better performance than algorithms that learn each task individually. Batch multi-task learning problems are usually solved by transfer learning methods \citep{pan2010survey} which transfer the knowledge learnt from one task to another similar tasks. In Online Multi-task Learning (OML) \citep{dekel2006online,li2010micro,li2014collaborative}, however, the tasks have to be solved in parallel with instances arriving sequentially, which makes the problem more challenging.

During time $t$, each of the task $i\in \{1,...K\}$ receives an instance $\x_{i,t}\in \R^{d_i}$, where $d_i$ is the feature dimension of task $i$. The algorithm then makes a prediction for each task $i$ based on the current model $\w_{i,t}$ as $\hat y_{i,t}=\text{sign}(\w_{i,t}^\top \x_{i,t})$. After making the prediction, the true labels $y_{i,t}$ are revealed and we get a loss function vector $\bm \ell_{i,t}\in \R_+^K$. Finally, the models are updated by considering the loss vector and task relationship.

A straightforward baseline algorithm is to parallel update all the classifiers $\w_i, i\in \{1,...,K\}$. OML algorithm should utilize the relationships between tasks to achieve higher accuracy compared with the baseline. The multitask Perceptron algorithm \citep{cavallanti2010linear} is a pioneering work of OML that considers the inter-task relationship. Assuming that a matrix $A\in\R^{K*K}$ is known and fixed, we can update the modePrasanthi Nairl $i$ when an instance $\x_{j,t}$ for task $j$ is received as follows:
\bqs
\w_{i,t+1}=\w_{i,t}+y_{j,t}A_{j,i}^{-1}\x_{j,t}
\eqs
Other approaches in \citep{saha2011online,wang2013online} learned to optimize the relationship matrix, which also offers the flexibility of using a time-varying relationship.

Apart from learning the relationship explicitly, another widely used approach in OML field is to add some structure regularization terms to the original objective function \citep{evgeniou2004regularized,yang2010online,kumar2012learning}. For example, we may assume that each model is made up of two parts, a shared part across all tasks $\w_0$ and an individual part $\mathbf v_i$, i.e.,
$ \w_i=\w_0+\mathbf v_i $ where the common part helps to take advantage of the task similarity. Now the regularized loss becomes
\bqs
\sum_{i=1}^K (\ell_{i,t}+||\mathbf v_i||_2^2)+\lambda||\w_0||_2^2
\eqs
It was also improved using more complex inter-task relationship \citep{murugesan2016adaptive}.

\subsection{Online Multi-view Learning}
Multi-view learning deals with problems where data are collected from diverse domains or obtained from various feature extractors. By exploring features from different views, multi-view learning algorithms are usually more effective than single-view learning. In literature, there many surveys that offer comprehensive summary of state-of-the-art methods in multi-view learning in batch setting \citep{sun2013survey,xu2013survey,li2016multi}, while few works tried to address this problem in online setting.

\paragraph{Two-view PA} We first introduce a seminal work, the two-view online passive aggressive learning (Two-view PA) algorithm \citep{nguyen2012two}, which is motivated by the famous single-view PA algorithm \citep{crammer2006online} and the two-view SVM algorithm \citep{farquhar2006two} in batch setting.

During each iteration $t$, the algorithm receives an instance $(\x_t^A,\x_t^B,y_t)$, where $\x_t^A\in \R^n$ is the feature vector in the first view, $\x_t^B\in \R^m$ is for the second view and $y_t\in\{1,-1\}$ is the label. The goal is to learn two classifiers $\w^A\in\R^n$ and $\w^B\in\R^m$, each for one view, and make accuracy prediction with their combination
$$\hat y_t=\text{sign}(\w^A \cdot\x^A+\w^B\cdot\x^B).$$
Thus the hinge loss at iteration $t$ is redefined as
$$\ell_t(\w_t^A,\w_t^B)=\max(0,1-\frac{1}{2}y_t (\w^A \cdot\x_t^A+\w^B\cdot\x_t^B))$$
In the single-view PA algorithm, the objective function in each iteration is a balance between two desires: minimizing the loss function at the current instance and minimizing the change made to the classifier. While to utilize the special information in the multi-view data, an additional term that measures the agreement between two terms is added. Thus, the optimization is as follows,
\bqs
&\hspace{-0.5in}(\w_{t+1}^A,\w_{t+1}^B)={\arg\min}_{\w^A,\w^B}\frac{1}{2}||\w^A-\w_t^A||_2^2+\frac{1}{2}||\w^B-\w_t^B||_2^2\\
&+C\ell_t(\w_t^A,\w_t^B)+\gamma|y_t\w^A\cdot\x_t^A-y_t\w^B\cdot\x_t^B|
\eqs
where $\gamma$ and $C$ are weight parameters. Fortunately, this optimization problem has a closed form solution.

\paragraph{Other related works:} Other than solving classification tasks, online multi-view learning has been explored for solving similarity learning or distance metric learning, such as Online multimodal deep similarity learning \citep{WHX+13} and online multi-modal distance metric learning \citep{wu2016online}.



\subsection{Online Transfer Learning}

Transfer learning aims to address the machine learning tasks of building models in a new target domain by taking advantage of information from another existing source domain through knowledge transfer. Transfer learning is important for many applications where training data in a new domain may be limited or too expensive to collect. There are two different problem settings, $homogeneous$ setting where the target domain shares the same feature space as the
old/source one, and $heterogeneous$ setting where the feature space of the target domain is different from that of the source domain. Although several surveys on transfer learning are available \citep{sousatransfer,pan2010survey}, most of the referred algorithms are in batch setting.

Online Transfer Learning (OTL) algorithms aim to learn a classifier $f:\R^{d}\rightarrow \R$ from a well-trained classifier $h: \R^{d'} \rightarrow \R$ in the source domain and a group of sequentially arriving instances $\x_t\in R^{d}, t=1,...,T$ in the target domain. For conciseness, we will use the previous notations for the online classification task. We first introduce a pioneer work of OTL \citep{zhao2014online,zhao2010otl}.

\paragraph{Homogeneous Setting} One key challenge of this task is to address
the concept drifting issue that often occurs in this scenario. The algorithm in homogeneous setting ($d=d'$) is based on the ensemble learning approach. At time $t$, an instance $\x_t$ is received. The algorithm makes a prediction based on the weighted average of the classifier in the source domain $h(\x_t)$ and the current classifier in the target domain $f_t(\x_t)$,
$$\hat{y_t}=\sign(w_{1,t} \Pi(h(\x_t))+w_{2,t}\Pi(f_t(\x_t))-\frac{1}{2}) $$
where $w_{1,t}>0, w_{2,t}>0$ are the weights for the two functions and need to be updated during each iteration. $\Pi$ is a normalization function, i.e. $\Pi(a)=\max(0,\min(1,\frac{a+1}{2}))$.

In addition to updating the function $f_t$ by using some online learning algorithms, the weights $w_{1,t}$ and $w_{2,t}$ should also be updated. One suggested scheme is
\bqs
\begin{aligned}
w_{1,t+1}= C_t w_{1,t} \exp(-\eta\ell^*(h)), \quad w_{2,t+1}= C_t w_{2,t} \exp(-\eta\ell^*(f_t))
\end{aligned}
\eqs
where $C_t$ is a normalization term to keep $w_{1,t+1}+w_{2,t+1}=1$ and $\ell^*(g)=(\Pi(g(\x_t))-\Pi(y_t))^2$.
\paragraph{Heterogeneous Setting}
Since heterogeneous OTL is generally very challenging, we consider one simpler case where the feature space of the source domain is a subset of that of the target domain. Without loss of generality, we assume the first $d'$ dimensions of $\x_t$ represent the old features, denoted as $\x_t^{(1)}\in\R^{d'}$. The other dimensions form a feature vector $\x_t^{(2)}\in\R^{d-d'}$. The key idea is to adopt a co-regularization principle of online learning two classifiers $f_t^{(1)}$ and $f^{(2)}_t$ simultaneously from the two views, and predict an unseen example on the target domain by $$\hat y_t =\sign\Bigg(\frac{1}{2}f_t^{(1)}(\x_t^{(1)})+\frac{1}{2}f_t^{(2)}(\x_t^{(2)})\Bigg)$$
The function from source domain $h(\x^{(1)})$ is used to initialize $f_t^{(1)}$. The update strategy at time $t$ is
\bqs\begin{aligned}&(f_{t+1}^{(1)},f_{t+1}^{(2)})=	&\arg\min_{f^{(1)},f^{(2)}} \frac{\gamma_1}{2}||f^{(1)}-f_t^{(1)}||^2_\H+\frac{\gamma_1}{2}||f^{(2)}-f_t^{(2)}||^2_\H+C\ell_t\end{aligned}\eqs
where $\gamma_1$, $\gamma_2$ and $C$ are positive parameters and $\ell_t$ is the hinge loss.

\paragraph{Other Related Work}
Multi-source Online Transfer Learning (MSOTL) \citep{ge2013oms,tommasi122012leveraging} solves a more challenging problem where $k$ classifiers $h_1, ...h_k$ are provided by $k$ sources. The goal is to learn the optimal combination of the $k$ classifiers and the online updated classifier $f_t$. A naive solution is to construct a new $d+k$ dimensional feature representation $\x'_t=[\x_t, h_1(\x_t), ..., h_k(\x_t)]$ and the online classifier in this new feature space. An extension of MSOTL \citep{ge2014handling} aims to deal with transfer learning problem under two disadvantageous assumptions, negative transfer where instead of improving performance, transfer learning from highly irrelevant sources degrades the performance on the target domain, and imbalanced distributions where examples in one class dominate. The Co-transfer Learning algorithm \citep{bhatt2014improving,bhatt2012matching} considers the transfer learning problem not only in multi-source setting but also in the scenario where a large group of instances are unlabeled.

\subsection{Online Metric Learning}
Distance metric learning (DML) \citep{yang2006distance} or similarity learning \citep{haoonline} is an important problem in machine learning, which enjoys many real-world applications, such as image retrieval, classification and clustering. The goal of classic DML is to seek a distance matrix $A\in \R^{d\times d}$ that defines the Mahalanobis distance between any two instances $\x_i\in R^d$ and $\x_j\in \R^d$
$$
d_A(\x_i,\x_j)=(\x_i-\x_j)^\top A(\x_i-\x_j) =||W\x_i-W\x_j||_2^2
$$
Typically, matrix $A\succeq 0$ is required to be symmetric positive semi-definite, i.e., there exist a matrix $W\in\R^{d\times d}$ such that $A=W^\top W$. 
It is often hard to collect training data with the exact true values of distances. Therefore, there are two types of problem settings for online DML: 1) Pairwise data, where at each round $t$ the learner receives a pair of instances $(\x_t^1,\x_t^2)$ and a label $y_t$ which is $+1$ if the pair is similar and $-1$ otherwise; 2) Triple data, where at each round $t$ the learner receives a triple $(\x_t,\x_t^+,\x_t^-)$, with the feedback that $d_A(\x_t,\x_t^+)>d_A(\x_t,\x_t^-)$. The goal of online learning is to minimize the accumulated loss during the whole learning process $\sum_{t=1}^T\ell_t(A)$, where $\ell_t$ is the loss suffered from imperfect prediction at round $t$. When evaluating the output model for online-to-batch-conversion, we may use the metric in information retrieval to evaluate the actual performance, such as mean average precision (mAP) or precision-at-top-$k$.

Below, we briefly introduce a few representative work for DML in online setting.
\paragraph{Pseudo-metric Online Learning (POLA)} The POLA algorithm \citep{shalev2004online} learns the distance matrix $A$ from a stream of pairwise data. The loss at time $t$ is an adaptation of the hinge loss
$$\ell_t(A,b)=\max\{0,1-y_t(b-d_A(\x_t^1,\x_t^2))\}$$
where $b$ is the adaptive threshold value for similarity and will be updated incrementally along with matrix $A$. We denote $(A,b)\in\R^{d^2+1}$ as the new variable to learn. The update strategy mainly follows the PA approach
\begin{equation*}
\begin{aligned}
(A_{t+\frac{1}{2}},b_{t+\frac{1}{2}})=\arg\min_{(A,b)}||(A,b)-(A_t,b_t)||_2^2 \quad s.t.~\ell_t(A,b)=0
\end{aligned}
\end{equation*}
The solution $(A_{t+\frac{1}{2}},b_{t+\frac{1}{2}})$ ensures correct prediction to current pair while makes the minimal change to the previous model. Then, the algorithm projects this solution to the feasible space $\{(A,b):A\succeq0, b\geq 1\}$ to obtain the updated model $(A_{t+1},b_{t+1})$. Like PA, one can generalize POLA to soft-margin variants to be robust to noise.

Another similar work named Online Regularized Metric Learning \citep{jin2009regularized} is simpler due to the adoption of fixed threshold, and adopts the following loss function
$$\ell_t(A)=\max(0,b-y_t(1-d_A(\x_t^1,\x_t^2)))$$
whose gradient is
$$\nabla \ell_t(A)=y_t(\x_t^1-\x_t^2)(\x_t^1-\x_t^2)^\top$$
At time $t$, if the prediction is incorrect, the algorithm updates the matrix $A$ by projecting the OGD updated matrix into the positive definite space.

\paragraph{Information Theoretic Metric Learning (ITML).} In the above algorithms, distances between two matrices $A_t$ and $A$ are usually defined using the Frobenius norm, i.e. $||A_t-A||_F^2$. The ITML algorithm \citep{davis2007information,jain2009online} adopts a different definition from an information theoretic perspective. Given a Mahalanobis distance parameterized by $A$, its corresponding multivariate Gaussian
distribution is $p(\x,A)=\frac{1}{Z}\exp(-\frac{1}{2}d_A(\x,\bm\mu))$. The difference between matrices is defined as the KL divergence between two distributions. Assuming all distributions have the same mean, the KL divergence can be calculated as
$$\text{KL}(p(\x;A),p(\x,A_t))=\text{tr}(AA_t^{-1})-\log\det(AA_t^{-1})-d $$
Similar to the PA update strategy, during time $t$ the matrix is updated by optimizing
$$A_{t+1}=\arg\min_{A\succeq0}\text{KL}(p(\x;A),p(\x,A_t))+\eta\ell_t(A)$$
where $\eta>0$ is a regularization parameter. This optimization has a closed-form solution.
\paragraph{Online Algorithm for Scalable Image Similarity Learning (OASIS)} The OASIS algorithm learns a bilinear similarity matrix $W\in\R^d$ from a stream of triplet data, where the bilinear similarity measure between two instances is defined as
$$S_W(\x_i,\x_j)=\x_i^\top W\x_j$$
During time $t$, one triplet $(\x_t,\x_t^+,\x_t^-)$ is received. Ideally, we expect the $\x_t$ is more similar to $\x_t^+$ than to $\x_t^-$, i.e. $S_W(\x_t,\x_t^+)>S_W(\x_t,\x_t^-)$. Similar to the PA algorithm, for a large margin, the loss function is defined as the hinge loss
$$\ell_t(W)=\max\{0,1-S_W(\x_t,\x_t^+)+S_W(\x_t,\x_t^-)\}$$
The optimization problem to solve for updating $W_t$ is
\begin{equation*}
\begin{aligned}
W_{t+1}&=\arg\min_W \frac{1}{2}||W-W_t||_F^2+C\xi s.t.~~~~\ell_t(W)\leq \xi~~and~~\xi\geq 0
\end{aligned}
\end{equation*}
where $C$ is the parameter controlling the trade-off. The OASIS algorithm differs from the previous work in several aspects. First, it does not require the similarity matrix $W$ to be positive semi-definite and thus saves computational cost for the projection step. The bilinear similarity matrix may be better than the Mahalanobis distance for some applications. Third, the triplet data may be easier to collect in some applications. 

Most of the previous methods assume a linear proximity function, \citep{xia2014online} overcomes the limitaiton using a kernelized approach for metric learning. Another approach in \citep{gao2014soml,gao2017sparse} performs sparse online metric learning for very high dimensional data. There is also some work that solves the online similarity learning in an active learning setting \citep{hao2015learning}, which significantly reduces the cost of collecting labeled data. Some work in \citep{chen2015simapp,chen2016mobile} also applied techniques of the online similarity learning for real-world applications of mobile application recommendation and tagging. The series of work in \citep{xia2013online,wu2016online} proposed the online multi-modal distance metric learning algorithms which learn distance metrics in multiple modalities, enabling multimedia information retrieval applications. 

\subsection{Online Collaborative Filtering}

Collaborative Filtering (CF) \citep{heckel2017sample} is an important learning technique for recommender systems. Different from content-based filtering techniques, CF algorithms usually require minimal knowledge about the features of items or users apart from the previous preferences. The fundamental assumption of CF is that if two users rate many items similarly, they expect to share common preference on some other items. Several survey papers in \citep{shi2014collaborative,su2009survey} gave detailed reviews of regular CF techniques. However, most of them assume batch settings. Below we introduce basics of CF and then review several popular online algorithms for CF tasks.

An online CF algorithm works on a sequence of observed ratings given by $n$ users to $m$ items.
At time $t\in\{1,2,...,T\}$, the algorithm receives the index of a user $u^{(t)}\in\{1,2,...n\}$ and the index of an item $i^{(t)}\in\{1,2,...m\}$ and makes a prediction of the rating $\hat r_{u,i}^{(t)}\in \R$ based on the knowledge of the previous ratings. Then the real rating $r_{u,i}^{(t)}\in\R$ is revealed and the algorithm updates the model based on the loss suffered from the imperfect prediction, denoted as $\ell(\hat r_{u,i}^{(t)},r_{u,i}^{(t)})$. The goal of online CF is to minimize the Root Mean Square Error (RMSE) or Mean Absolute Error(MAE) along the whole learning process, defined as follows:
$$\mathrm{RMSE}=\sqrt{\frac{1}{T}\sum_{t=1}^T{(r_{u,i}^{(t)} - \hat{r}_{u,i}^{(t)})^2}},\quad\mathrm{MAE}=\frac{1}{T}\sum_{t=1}^T|r_{u,i}^{(t)} - \hat{r}_{u,i}^{(t)}|$$

CF techniques are generally categorized into two types: memory-based methods and model-based methods. Below briefly introduces some popular algorithms in each category.

\paragraph{Memory-Based CF Methods.} This follows the instance-based learning paradigm:
 \begin{enumerate}
 \item Calculate the similarity score between any pairs of items. For example, the cosine similarity between item $i$ and item $j$ is defined as,
 $$S_{i,j}=\frac{\sum_{u\in \mathcal{U}_i\cap \mathcal{U}_j}r_{ui}\cdot r_{u,j}}{\sqrt{\sum_{u\in\mathcal{U}_i}r_{ui}^2\sum_{u\in\mathcal{U}_j}r_{u,j}^2}}$$
where $\mathcal{U}_i$ denotes the set of users that have rated item $i$.
\item For each item $i$, find its $k$ nearest neighbor set $\mathcal{N}_i$ based on the similarity score.
\item Predict the rating $r_{u,i}$ as the weighted average of ratings from user $u$ to the neighbors of item $j$, where the weight is proportional to the similarity.
\end{enumerate}
We name the above described algorithm as item-based CF, while similarly, the predictions may also be calculated as the weighted average of ratings from similar users, which is called user-based CF method.
Memory-based CF methods were used for some early generation recommendation systems, but very few is online learning approach. One reason is because of data sparsity, as the similarity score $S_{i,j}$ is only available when there is at least one common user that rates the two items $i$ and $j$, which might be unrealistic during the beginning stage. Another challenge is the large time consumption when updating the large number of similarity scores incrementally with the arrival of new ratings. The Online Evolutionary Collaborative Filtering \citep{liu2010online} algorithm provides an efficient similarity score updating method to address this problem.

\paragraph{Model-Based CF Methods} Memory-based online CF methods suffer two limitations, i.e., sensitivity due to data sparsity and inefficiency for similarity score update. To address these issues, extensive work hasbeen focused on model-based CF algorithms. One of the most successful approaches is the matrix factorization methodology \citep{koren2009matrix}, which assumes the rating by a user to an item is determined by $k$ potential features, $k\ll n,m$. Thus each user $u$  can be represented by a vector $\u_{u}\in \R^{k}$, and each item $i$ can be represented by a vector $\v_{i}\in \R^{k}$. The rating $r_{u,i}$ can then be approximated by the dot product of the corresponding user vector and item vector, i.e., $\hat r_{u,i}=\u_u^{\top}\v_i$. The CF problem can then be represented by the following optimization problem:
\begin{eqnarray*}
\label{eq:mf} \mathop{\arg\min}_{U\in\R^{k\times n},V\in
\R^{k \times m}}\hspace{-0.02in}\sum_{t=1}^T\ell(\u_{u}^{(t)}\cdot\v_{i}^{(t)},r_{u,i}^{(t)})
\end{eqnarray*}
where the loss function is defined to optimize certain evaluation metric:
$$\ell_{rmse}(\hat r_{u,i},r_{u,i})=(r_{u,i}-\hat r_{u,i})^2~~~\ell_{mae}(\hat r_{u,i},r_{u,i})=|r_{u,i}-\hat r_{u,i}|$$
The regularized loss at time $t$ is
$$\mathcal{L}_t=\lambda||\u_u^{(t)}||_2^2+\lambda||\v_i^{(t)}||_2^2+\ell(\u_{u}^{(t)}\cdot\v_{i}^{(t)},r_{u,i}^{(t)})$$
where $\lambda>0$ is the regularization parameter. A straightforward CF approach is to apply OGD on the regularized loss function \citep{abernethy2007online},
$$\u_u^{(t+1)}=(1-2\eta\lambda)\u_u^{(t)}-\eta \frac{\partial\ell(\u_{u}^{(t)}\cdot\v_{i}^{(t)},r_{u,i}^{(t)})}{\partial\u_u^{(t)}},\quad\v_i^{(t+1)}=(1-2\eta\lambda)\v_i^{(t)}-\eta \frac{\partial\ell(\u_{u}^{(t)}\cdot\v_{i}^{(t)},r_{u,i}^{(t)})}{\partial\v_i^{(t)}}$$
where $\eta>0$ is the learning rate. Later, several improved algorithms are proposed, such as Online Multi-Task Collaborative Filtering algorithm \citep{wang2013online}, Dual-Averaging Online Probabilistic Matrix Factorization algorithm \citep{yuan2014accelerated}, Adaptive Gradient Online Probabilistic Matrix Factorization algorithm \citep{yuan2014accelerated} and Second-order Online Collaborative Filtering algorithm \citep{lu2013second,liu2016onlinea}. These algorithms adopt more advanced update strategies beyond OGD and thus can achieve faster adaptation for rapid user preference changes in real-world recommendation tasks.

Besides the algorithms introduced, there are many online CF methods that explore other challenging tasks. First, in many applications, both features of users and items are available and thus need to be considered for better prediction. This generalized CF problem can be solved by using tensor product kernel functions. For example, the Online Low-rank with Features algorithm \citep{abernethy2007online} addresses this problem in online setting. However, it only adopts the linear kernel for efficiency. Perhaps, better performance might be achieved if online budget learning algorithms are adopted. Second, most CF algorithms are based on a regression model, which is mainly concerned with the accuracy of rating prediction, while there are some applications where ranking prediction might be much more important. Two algorithms based on OGD and Dual Averaging approaches are proposed to address this problem by replacing the regression-based loss with the ranking-based loss \citep{ling2012online}. Third, for very large-scale applications, when the model has to be learnt using parallel computing, conventional OGD update is not suitable because of the possible conflict in updating the user/item vectors. The Streaming Distributed Stochastic Gradient Descent algorithm \citep{ali2011parallel} provides an operable approach to addresses this problem. Finally, the CF methods for Google News recommendation \citep{das2007google} is a combination of memory-based and model-based algorithms.
Last but not least, to address the sparsity problem and imbalance of rating data, \citep{liu2017collaborative} incorporate content information via latent dirichlet allocation into online CF.

\subsection{Online Learning to Rank}

Learning to rank is an important family of machine learning techniques for information retrieval \citep{trotman2005learning,cao2007learning,hang2011short,zoghi2017online,shah2017online}. Different from classification problems where instances are classified as either ``relevant" or ``not relevant", learning to rank aims to produce a permutation of a group of unseen instances which is similar to the knowledge acquired from the previously seen rankings. To evaluate the performance of ranking algorithms, metrics for information retrieval such as Mean Average Precision (MAP), Normalized Discounted Cumulative Gain (NDCG) and Precision-At-Top-$k$ are most popular.


Unlike traditional learning to rank methods which are often based on batch learning \citep{hang2011short}, we mainly focus on reviewing existing learning to rank methods in online settings \citep{wang2015solar,wan2015online}, where instances are observed sequentially. Learning to rank techniques are generally categorized into two approaches: pointwise and pairwise. We will introduce some of the most representative algorithms in each category.

\paragraph{Pointwise Approach:} We first introduce a simple Perceptron-based algorithm, the Prank \citep{crammer2001pranking,crammer2005online}, which provides a straightforward view of the commonly used problem setting for pointwise learning to rank approaches.

To define the online learning to rank problem setting formally,
we have a finite set of ranks $\mathcal{Y}=\{1,...,k\}$ from which a rank $y\in\mathcal{Y}$ is assigned to an instance $\x\in\R^d$. During time $t$, an instance $\x_t$ is received and the algorithm makes a prediction $\hat y_t$ based on the current model $H_t: \R^d\rightarrow \mathcal{Y}$. Then the true rank $y_t$ is revealed and the model is updated based on the loss $\ell(\hat y_t,y_t)$. The loss, for instance, can be defined as $\ell(\hat y_t, y_t)= |\hat y_t, y_t|$. The goal of the online learning to rank task is to minimize the accumulated loss along the whole learning process $\sum_{t=1}^T \ell(\hat y_t, y_t)$.

The ranking rule of Prank algorithm consists of the combination of Perceptron weight $\w\in \R^d$ and a threshold vector $\mathbf{c}\in \{\R, \infty \}^d$, whose elements are in nondecreasing order i.e., $c^1\leq c^2\leq,...,\leq c^k=\infty$. Like the Perceptron algorithm, the rank prediction is determined by the value of the inner product $\w_t^\top\x_t$,
$$\hat y_t=\min_{r\in\{1,...,k\}}\{r: \w_t^\top\x_t<c^r_t\}$$
We can expand the target rank $y_t$ to a vector $\y_t=\{+1,...,+1,-1,...,-1\}\in\R^k$. For $r=1,...k$, $y_t^r=-1$ if $y_t<r$, and $y_t^r=1$ otherwise. Thus, for a correct prediction, $y^r_t(\w_t^\top\x_t-c_t^r)>0$ holds for all $r\in\mathcal{Y}$. When a mistake appears $\hat y_t\neq y_t$, there is subset $\mathcal{M}$ of $\mathcal{Y}$ where $y^r_t(\w_t^\top\x_t-c_t^r)>0$ does not hold. The update rule is to move the corresponding thresholds for ranks in $\mathcal{M}$ and the weight vector toward each other:
$$\w_{t+1}=\w_t+(\sum_{r\in\mathcal{M}}y_t^r)\x_t,~~~and ~~c_{t+1}^r=c_{t}^r-y_t^r, \forall r\in \mathcal{M}$$

In theory, the elements in threshold vector $\mathbf{c}$ are always in nondecreasing order and the total number of mistakes made during the learning process is bounded.

Online Aggregate Prank-Bayes Point Machine (OAP-BPM) \citep{harrington2003online} is an extension of the Prank algorithm by approximating the Bayes point. Specifically, the OAP-BPM algorithm generates $N$ diverse solutions of $\w$ and $\mathbf{c}$ during each iteration and combines them for a better final solution.
We denote $H_{j,t}$ as the $j$-th solution at time $t$. The algorithm samples $N$ Bernoulli variables $b_{j,t}\in\{0,1\}, j=\{1,...,N\}$ independently. If $b_{i,t}=1$, The $j$-th solution is updated using the Prank algorithm according to the current instance, $H_{j,t+1}=Prank(H_{j,t}, (\x_t,y_t))$. Otherwise, no update is conducted to the $j$-th solution. The solution $\w_{t+1}$ and $\mathbf{c}_{t+1}$ is the average over $N$ solutions. This work shows better generalization performance than the basic Prank algorithm.

\paragraph{Pairwise Approach:}
 One simple method is to address the ranking problem by transforming it to a classification problem \citep{herbrich1999support}. In a more challenging problem setting, where no accurate rank $y$ is available when collecting the data, only pairwise instances are provided. At time $t$, a pair of instances $(\x_t^1,\x_t^2)$ are received with the knowledge that $\x_t^1$ is ranked before $\x_t^2$ or the inverse case, and the aim is to find a function $f: \R^d\rightarrow \R$ that fits the instance pairs, i.e., $f(\x^1)>f(\x^2)$ when it is known $\x_t^1$ is ranked before $\x_t^2$ or otherwise $f(\x^1)<f(\x^2)$. When the function is linear, the problem can be rewritten as $\w^\top(\x^1-\x^2)>0$ when $\x^1$ is in front and otherwise $\w^\top(\x^1-\x^2)<0$, where $\w\in\R^d$ is the weight vector. This problem can easily be solved by using a variety of online classification algorithms (Online Gradient Descent \citep{chapelle2010efficient} for example).

\subsection{Distributed Online Learning}

Distributed online learning \citep{zhang2017projection} has become increasing popular due to the explosion in size and complexity of datasets. Similar to the mini-batch online learning, during each iteration, $K$ instances are received and processed simultaneously. Usually, each node processes one of the instances and updates its local model. These nodes communicate with each other to make their local model consistent.
When designing a distributed algorithm, besides computational time cost and accuracy, another important issue to consider is the communication load between nodes. This is because in real world systems with limited network capacity and large communication burden result in long latency.

Based on the network structure, distributed online learning algorithms can be classified into two groups: {\it centralized} and {\it decentralized} algorithms. A centralized network is made up of 1 master node and $K-1$ worker nodes, where the workers can only communicate with the master node. By gathering and distributing information across the network, it is not difficult for distributed algorithms to reach a global consensus \citep{boyd2011distributed}. In decentralized networks, however, there is no master and each node can only communicate with its neighbors \citep{agarwal2010distributed,mota2013d}. Although the algorithms are more complex, decentralized learning is more popular because of the robustness of network structure.

We can also group the distributed learning algorithms by {\it synchronized} and {\it asynchronized} working modes. Synchronized algorithms are easy to design and enjoy better theoretical bounds but the speed of the whole network is limited by the slowest node. Asynchronized learning algorithms, on the other hand, are complex and usually have worse theoretical bounds. The advantage is its faster processing speed \citep{smyth2009asynchronous}.

\subsection{Online Learning with Neural Networks}

In addition to kernel-based online learning approaches, another rich family of nonlinear online learning algorithms follows the general idea of neural network based learning approaches \citep{williams1989learning,platt1991resource,carpenter1991artmap,lecun2012efficient,liang2006fast,polikar2001learn++}. For example, the Perceptron algorithm could be viewed as the simplest form of online learning with neural networks (but it is not nonlinear due to its trivial network). Despite many extensive studies for online learning (or incremental learning) with neural networks, many of existing studies in this field fall short due to some critical drawbacks, including the lack of theoretical analysis for performance guarantee, heuristic algorithms without solid justification, and computational too expense to achieve efficient and scalable online learning. Due to the large body of related work, it is impossible to examine every piece of work in this area. In the following, we review several of the most popularly cited related papers and discuss their key ideas for online learning with neural networks.

A series of related work has explored online convex optimization methods for training classical neural network models \citep{bottou-98x}, such as the Multi-layer Perceptron (MLP). For example, online/stochastic gradient descent has been extensively studied for training neural networks in sequential/online learning settings, such as the efficient back-propagation algorithm using SGD \citep{lecun2012efficient}. These works are mainly motivated to accelerate the training of batch learning tasks instead of solving online learning tasks directly and seldom give theoretical analysis.

In addition to the above, we also briefly review other highly cited works that address online/incremental learning with neural networks. For example, the study in \citep{williams1989learning} presented a novel learning algorithm for training fully recurrent neural networks for temporal supervised learning which can continually run over time. However, the work is limited in lacking theoretical analysis and performance guarantee, and the solution could be quite computationally expensive. The work in \citep{platt1991resource} presented a Resource-Allocating Network (RAN) that learns a two-layer network by a strategy for allocating new units whenever an unusual pattern occurs and a learning rule for refining the network using gradient descent. Although the algorithm was claimed to run in online learning settings, it may suffer poor scalability as the model complexity would grow over time.
The study in \citep{carpenter1991artmap} proposed a new neural network architecture called ``ARTMAP" that autonomously learns to classify arbitrarily many, arbitrarily ordered vectors into recognition categories based on predictive success. This supervised learning system was built from a pair of ART modules that are capable of self organizing stable recognition categories in response to arbitrary sequences of input patterns. Although an online learning simulation has been done with ART (adaptive resonance theory), the solution is not optimized directly for online learning tasks and there is also no theoretical analysis.
The work in \citep{liang2006fast} proposed an online sequential extreme learning machine (OS-ELM) which explores an online/sequential learning algorithm for training single hidden layer feedforward networks (SLFNs) with additive or radial basis function (RBF) hidden nodes in a unified framework. The limitation also falls short in some heuristic approaches and lacking theoretical analysis. Last but not least, there are also quite many studies in the field which claim that they design neural network solutions to work online, but essentially they are not truly online learning. They just adapt some batch learning algorithms to work efficiently for seqential learning environments, such as the series of Learning++ algorithms and their variants~\citep{polikar2001learn++,elwell2011incremental}. Recently, Hedge Backpropagation \citep{sahoo2017online} was proposed to learn deep neural networks in the online setting with the aim to address slow convergence of deep networks through dynamic depth adaptation.

\subsection{Online Portfolio Selection}

On-line Portfolio Selection (OLPS) is a natural application of online learning for sequential decisions of selecting a portfolio of stocks for optimizing certain metrics, e.g. cumulative wealth, risk adjusted returns, etc. \citep{bianchi-2006-prediction,li2014online,li2015online,li2016olps}. Consider a financial market with $m$ assets, in which we have to allocate our wealth. At every time period (or iteration), the price of the $m$ stocks changes by a factor of $\x_t \in \R_+^m$. This vector is also called the price relative vector. $x_{t,i}$ denotes the ratio of the closing price of asset $i$ at time $t$ to the last closing price at time $t-1$. Thus, an investment in asset $i$ changes by a factor of $x_{t,i}$ in period $t$. At the beginning of time period $t$  the investment is specified by a portfolio vector $\b_t \in \delta_m$ where $\delta_m = {\b:\b \succeq 0, \b^\top \mathbf{1} = 1}$. The portfolio is updated in every time-period based on a specific strategy, and produces a sequence of mappings:
\bqs
	\b_1 = \frac{1}{m}, \quad \b_t: \R_+^{m(t-1)} \rightarrow \delta_m, \quad t= 2,3, \dots, T
\eqs
where $T$ is the maximum length of the investment horizon. To make a decision for constructing a portfolio at time $t$, the entire historical information from $\x_1, \dots, \x_{t-1}$ is available. The theoretical framework starts with a wealth of $S_0 = 1$, and at the end of every time period, the wealth changes as $S_t = S_{t-1} \times (\b_t^\top \x_t)$.

Most efforts in OLPS make a few (possibly unrealistic) assumptions, including no transaction costs, perfectly liquid market, and no impact cost (the portfolio selection strategy does not affect the market). Besides the traditional benchmarking approaches, the approaches for OLPS can be categorized into \emph{Follow-the-winner, Follow-the-loser, Pattern Matching}, and  \emph{Meta-Learning} approaches \citep{li2014online}. 

The benchmark approaches, as the name suggests, are simple baseline methods whose performance can be used to benchmark the performance of proposed algorithms. Common baselines are Buy and Hold (BAH) strategy, Best Stock and Constant Rebalanced Portfolio (CRP). The idea of BAH is
to start with a portfolio with equal investment in each asset, and never rebalance it. Best Stock is the performance of the asset with the highest returns at the end of the investment horizon. CRP \citep{kelly2011new} is a fixed portfolio allocation to which the portfolio is rebalanced to at the end of every period, and Best-CRP is the CRP which obtains the highest returns at the end of the investment horizon. It should be noted that Best Stock and Best-CRP strategies can only be executed in hindsight.

Follow-the-winner approaches adhere to the principle of increasing the relative portfolio allocation weight of the stocks that have performed well in the past. Many of the approaches are directly inspired by Convex Optimization theory,  including Universal Portfolios \citep{cover2011universal}, Exponential Gradient \citep{helmbold1998line}, Follow the Leader \citep{gaivoronski2000stochastic} and Follow the Regularized Leader\citep{agarwal2006algorithms}. In contrast to follow-the-winner, there is a set of approaches that aim to follow-the-loser, with the belief that asset prices have a tendency to revert back to a mean, i.e., if the asset price falls, it is likely to rise up in the next time-period. These are also called mean-reversion strategies. The early efforts in this category included Anti Correlation \citep{borodin2004can} which designed a strategy by betting making statistical bets on positive-lagged correlation and negative auto-correlation; and Passive-Aggressive Mean Reversion (PAMR) \citep{li2012pamr}, which extended the Online Passive Aggressive Algorithms \citep{crammer2006online} to update the portfolio to an optimal "loser" portfolio - by selecting a portfolio that would have made an optimal loss in the last observed time-period. A similar idea was used to extend confidence-weighted online learning to develop Confidence-Weighted Mean Reversion \citep{li2011confidence,li2013confidence}. The idea of PAMR was extended to consider multi-period asset returns, which led to the development of Online Moving Average Reversion (OLMAR) \citep{li2012line,li2015moving} and Robust Median Reversion (RMR) \citep{huang2013robust} strategies.

Another popular set of approaches is the Pattern-Matching approaches, which aim to find patterns (they may be able to exploit both follow-the-winner and follow-the-loser) for optimal sequential decision making. Most of these approaches are non-parametric. Exemplar approaches include \citep{gyorfi2003nonparametric,gyorfi2008nonparametric,li2011corn}.
Finally, meta-learning algorithms for portfolio selection aim to rebalance the portfolio on the basis of the expert advice. There are a set of experts that output a portfolio vector, and the meta-learner uses this information to obtain the optimal portfolio. In general, the meta-learner adheres to the follow-the-winner principle to identify the best performing experts. Popular approaches in this category include Aggregating Algorithms \citep{vovk1998universal}, Fast Universalization Algorithm \citep{akcoglu2004fast} and Follow the Leading History \citep{hazan2009efficient}. Besides these approaches, there are also efforts in portfolio selection with aims to optimize the returns accounting for transaction costs. The idea is to incorporate the given transaction cost into the optimization objective \citep{ormos2013performance,huang2015semi,li2017transaction}.

A closely related area to Online Portfolio Selection is Online Learning for Time Series Prediction.  Time series analysis and prediction~\citep{george1994time,clements1998forecasting,chatfield2000time,sapankevych2009time} is a classical problem in machine learning, statistics, and data mining. The typical problem setting of time series prediction is as follows: a learner receives a temporal sequence of observations, $x_1,\ldots,x_t$, and the goal of the learner is to predict the future observations (e.g., $x_{t+1}$ or onwards) as accurately as possible. In general, machine learning methods for time series prediction may also be divided into linear and non-linear, univariate and multivariate, and batch and online. Some time series prediction tasks may be resolved by adapting an existing batch learning algorithm using sliding window strategies.  Recently there have been some emerging studies for exploring online learning algorithms for time series prediction~\citep{anava2013online,liu2016online}.

\def \e {\mathbf{e}}
\def \g {\mathbf{g}}
\def \w {\mathbf{w}}
\def \x {\mathbf{x}}
\def \z {\mathbf{z}}

\def \R {\mathbb{R}}
\def \I {\mathbb{I}}
\def \E {\mathbb{E}}

\def \Pr {\mathrm{Pr}}

\def \bqs {\begin{eqnarray*}}
\def \eqs {\end{eqnarray*}}

\section{Bandit Online Learning}

\subsection{Overview}

Bandit online learning, a.k.a. the ``Multi-armed Bandit (MAB) problem~\citep{robbins1985some,katehakis1987multi,vermorel2005multi,bubeck2012regret,gittins2011multi}, is an important branch of online learning where a learner makes sequential decisions by receiving only partial feedback from the environment each time.

MAB problems are online learning tasks for sequential decisions with a trade-off between exploration and exploitation. Specifically, on each round, a player chooses one out of $K$ actions, the environment then reveals the payoff of the player's action, and the goal of the learner is to maximize the total payoff obtained during online learning process. A fundamental challenge of MAB is to address the exploration-exploitation tradeoff \citep{audibert2009exploration}, i.e., balancing between the {\it exploitation} of actions that gave highest payoffs in the past and the {\it exploration} of new actions that might give higher payoffs in the future.

MAB problems are collectively called ``bandit" problems. Historically, the name ``bandit" is referred to the scenario of playing slot machines in a casino, where a player faces a number of slot machines to insert coins (one slot machine is called one-armed bandit in American slang), the expected reward of each machine might be different, and the player's goal is to maximize the reward by repeatedly choosing where to insert the next coin.



MAB problems can be roughly divided into two major categories: \emph{stochastic MAB} and \emph{adversarial MAB}. The former assumes a stochastic environment where rewards (or ``losses" equivalently) are i.i.d. and independent from each other, while the later removes the stochastic assumption where the rewards (or losses) can be arbitrary or more formally ``adversarial" by adapting to the past decisions. Note that the notion of ``reward" and ``loss" are symmetric and can be translated from one to the other equivalently. To be consistent, if not mentioned specifically, we will use ``loss" instead of ``reward" for the rest discussion.

We now introduce the formal procedure of the MAB problem. Formally, a $K$-armed bandit problem takes place in a sequence of rounds with length $T\in\N$, where $T$ is often unknown at the beginning (typically a learner is called an ``anytime" algorithm if $T$ is unknown in advance). At the $t$-th round, the player chooses one out of $K$ actions $I_t\in[K]=\{1,\ldots,K\}$ using some strategy. After that, the environment reveals the loss $\ell_t(I_t)$ of the action to the forecaster. The goal is to minimize the total loss over the $T$ rounds. In theory, we are interested in analyzing the behavior of the learner, typically by comparing the performance of its actions against with some optimal strategy. More formally, we can define the ``regret" as the difference between the cumulative loss of the best fixed arm by an optimal strategy and that of the player after playing $T$ rounds $I_1,\ldots,I_T$ as
\bqs
R_T=\sum^T_{t=1}\ell_{t}(I_t) - \min_{i\in[K]}\sum^T_{t=1}\ell_t(i)
\eqs
As both loss $\ell_t(i)$ and action $I_t$ could be stochastic, we can define the expected regret as
\bqs
\E[R_T] = \E\left[\sum^T_{t=1} \ell_t(I_t) - \min_{i\in[K]}\sum^T_{t=1}\ell_t(i) \right]
\eqs
where the expectation is taken with respect to the random draw of both losses and learner's actions.

\subsection{Stochastic Bandits}
For stochastic MAB problem, each arm $i\in[k]$ corresponds to an unknown distribution $P_i$ on $[0,1]$, and the losses $\ell_t(i)$ are independently drawn from the distribution $P_i$ corresponding to the selected arm. Let us denote by $\mu_i$ the mean of the distribution $P_i$, and define
\bqs
\mu^*=\min_{i\in[k]}\mu_i\quad \rm{and}\quad i^*\in\arg\min_{i\in[k]}\mu_i
\eqs
The expected regret can be rewritten as
\bqs
\E[R_T] = \E[\sum^T_{t=1}\mu_{I_t}] - T\min_{i\in[K]}\mu_i = \E[\sum^T_{t=1}\mu_{I_t}] - T\mu^* = \E[\sum_{t=1}^T (\mu_{I_t}-\mu_*)]
\eqs
Further let $N_i(s)=\sum^s_{t=1}\I(I_t=i)$ denote the number of times the player selected arm $i$ on the first $s$ rounds and $\Delta_i=\mu_i-\mu_*$ be the suboptimality parameter of arm $i$, we can simplify the expected regret as follows:
\bqs
\E[R_T] =\E[\sum_{t=1}^T \Delta_{I_t}]=\sum^K_{i=1}\Delta_i\E[N_i(T)].
\eqs
\subsubsection{Stochastic Multi-armed Bandit}

In this section we mainly introduce two well-known algorithms for stochastic MAB.

\paragraph{$\epsilon$-Greedy.} The first simplest algorithm for stochastic MAB is called the $\epsilon$-Greedy rule \citep{barto1998reinforcement}. The idea is to with probability $1-\epsilon$ play the the current best arm of the highest average reward, and with probability $\epsilon$ play a random arm, where parameter $\epsilon>0$ is a constant value in $(0,1)$. Algorithm \ref{alg:epsilon-greedy} gives a summary of this algorithm.
\begin{algorithm}[htb]
\label{alg:epsilon-greedy}
\caption{$\epsilon$-Greedy}
\begin{algorithmic}
\STATE \textbf{INPUT:} parameter $\epsilon>0$ \\
\STATE \textbf{INIT}: empirical means  $\mu_i=0$, $\forall i \in [K]$\\
\FOR{ $t=1,2,\ldots,T$}
    \STATE with probability $1-\epsilon$ play the current best arm $i_t = \arg\min_{i\in[K]} \mu_i$
    \STATE with probability $\epsilon$ play a random arm
    \STATE receive $\ell_t(i_t)$
    \STATE update the empirical means $\mu_{i_t} = (\mu_{i_t}+\ell_t(i_t))/(N_{i_t}(t)+1)$
\ENDFOR
\end{algorithmic}
\end{algorithm}
However, the constant exploration probability $\epsilon$ results in a linear growth in the regret. One way to fix it is to decrease the value of $\epsilon$ over time and let it go to zero at a certain rate. For example, an improved $\epsilon_t$-greedy algorithm is to follow the epsilon-decreasing strategy by defining $\epsilon_t$ at round $t$ as
\bqs
\epsilon_t = \min\bigg\{1,\frac{cK}{d^2t}\bigg\}
\eqs
where $c>0$ and $d\in(0,1)$. When $0<d\leq\min_{i:\mu_i<\mu^*}\Delta_i<1$ and $T>\frac{cK}{d}$, this improved $\epsilon_t$-greedy algorithm can achieve the logarithm regret $O(\ln T)$.

\paragraph{UCB.}
Another well-known algorithm for stochastic MAB is the Upper Confidence Bound (UCB) algorithm~\citep{auer2002finite}, a strategy that simultaneously performs exploration and exploitation using a heuristic principle of {\it optimism in face of uncertainty}. The intuition is that, despite lacking knowledge about what actions are best, we will try to construct an optimistic guess as to how good the expected payoff/loss of each action is, and choose the action with the best guess. If our guess is correct, we will be able to exploit that action and incur little regret; but if our guess is wrong, then our optimistic guess will quickly decrease and we will then be compelled to switch to a different action, which therefore is able to balance the exploration-exploitation tradeoff.

Formally, the ``optimism" comes in the form of Upper Confidence Bound (UCB). In particular, the idea is to calculate the confidence intervals of the averages, which is a region around our estimates such that the true value falls within with high probability, and repeatedly shrink the confidence bounds such that the average will become more reliable. Algorithm \ref{alg:ucb1} gives a summary of the UCB algorithm.


\begin{algorithm}[ht]
\label{alg:ucb1}
\caption{UCB}
\begin{algorithmic}
\STATE \textbf{INPUT:} parameter $\epsilon>0$ \\
\STATE \textbf{INIT}: empirical means $\mu_i=0$, $\forall i \in [K]$\\
\FOR{ $t=1,2,\ldots,T$}
    \STATE play the arm $i_t=\arg\min_i (\mu_i - \sqrt{\frac{2\ln t}{N_i(t)}})$
    \STATE receive $\ell_t(i_t)$
    \STATE update the empirical means $\mu_{i_t} = (\mu_{i_t}+\ell_t(i_t))/(N_{i_t}(t)+1)$
\ENDFOR
\end{algorithmic}
\end{algorithm}

In theory, by running the UCB algorithm over $T$ rounds, the expected regret is
\bqs
\E[R_T] \leq 8\sum_{i:\mu_i<\mu^*}\Big(\frac{\ln T}{\Delta_i}\Big) + \Big(1+\frac{\pi^2}{3}\Big)\Big(\sum_{j=1}^K\Delta_j\Big)
\eqs
The above is a specific worst-case bound on the expected regret \citep{auer2002finite}. More concisely, one can show that the expected regret of UCB is at most $O(\sqrt{KT\ln T})$.
\cite{auer2002finite} also gave some variants of improved UCB algorithms. Improved algorithms and regret bound were also given in \citep{abbasi2011improved}.

\subsubsection{Bayesian Bandits}

Bayesian methods have been explored for studying bandit problems from the beginning of this field. One of the most well-known and classic algorithm in Bayesian Bandits is Thompson Sampling \citep{thompson1933likelihood}, which is considered one of the oldest algorithm to address the exploration-exploitation trade-off for bandit problems. Recent years have seen a lot of interests in analyzing both empirical performance \citep{chapelle2011empirical} and theoretical properties of Thompson Sampling for bandit problems \citep{agrawal2012analysis}. In the following, we introduce a Bayesian setting of Bernoulli bandit and then discusses the Thompson Sampling algorithm.

Consider a standard $K$-armed Bernoulli bandit, each action corresponds to the choice of an arm, and the reward of the $i$-th arm is either 0 or 1 which follows a Bernoulli distribution with mean $\mu_i$, i.e., the probability of success for arm $i$ (reward=1) is $\mu_i$.  The algorithm maintains Bayesian priors on the Bernoulli means $\mu_i$'s. At the beginning, the Thompson Sampling algorithm initializes each arm $i$ to have prior $Beta(1,1)$ on $\mu_i$, since $Beta(1,1)$ is the uniform distribution on $(0,1)$. On round $t$, after observing $S_i(t)$ successes (reward=1) and $F_i(t)$ failures (reward=0) on $k_i(t)=S_i(t)+F_i(t)$ times of playing arm $i$, the
algorithm updates the distribution on $\theta_i$ as $Beta(S_i(t)+1, F_i(t)+1)$. The algorithm then samples from these posterior distributions of the $\theta_i$'s, and plays an arm according to the probability of its
mean being the largest. Algorithm \ref{alg:ts} gives a summary of Thompson Sampling algorithm for $K$-armed Bernoulli bandits problems.

\begin{algorithm}[hptb]
\label{alg:ts}
\caption{Thompson Sampling}
\begin{algorithmic}
\STATE \textbf{INIT}: $S_i(1)=0, F_i(1) = 0, \forall i=1,\ldots,K$
\FOR{$t=1,2,\ldots,T$}
    \STATE For each arm $i\in[K]$, sample $\theta_i(t)$ from the $Beta(S_i+1,F_i+1)$ distribution.
    \STATE Play arm $i(t)=\arg\max_i \theta_i(t)$, and observe reward $r_t$
    \STATE If $r_t=1$, then $S_{i(t)}=S_{i(t)}+1$; else $F_{i(t)}=F_{i(t)}+1$.
\ENDFOR
\end{algorithmic}
\end{algorithm}
The above Thompson sampling algorithm can be easily extended to the general stochastic bandits setting, i.e., when the rewards for arm $i$ are generated from an arbitrary unknown distribution with support $[0,1]$ and mean $\mu_i$. It has also been extensively used for contextual bandit settings \citep{chapelle2011empirical,agrawal2013thompson}.

In theory, for the $K$-armed stochastic bandit problem, denoting $\Delta_i=\mu_1 - \mu_i$, the Thompson Sampling algorithm has the expected regret given in \citep{agrawal2012analysis}
\bqs
\mathbb{E}[R_T]\leq O\bigg(\big(\sum_{a=2}^K \frac{1}{\Delta_a^2}\big)^2 \ln T\bigg)
\eqs

{\it Other Related Work.} In addition to the classic Thompson Sampling algorithm, another notable variant of Bayesian Bandit is called the Bayes-UCB algorithm \citep{kaufmann2012bayesian}, which is a UCB-like algorithm, where the upper confidence bounds are based on the quantiles of Beta posterior distributions, and is able to achieve the lower bound of \cite{lai1985asymptotically} for Bernoulli rewards. More other extensive and recent studies of Bayesian Bandits can be found in \citep{scott2010modern,may2012optimistic,russo2016information}. 


\subsection{Adversarial Bandits}

In the previous setting of stochastic bandits, we generally assume that the rewards are i.i.d., which are drawn independently from some unknown but fixed distribution. We now relax such stochastic assumption on the rewards. We assume the reward distribution can be affected by the previous actions taken by the player, which is termed as ``Adversarial Bandits" problems \cite{auer1995gambling}. In the following, we review several classes of adversarial bandits, including some fundamentals of adversarial MAB and other active topics such as linear bandits and combinatorial bandits.

\subsubsection{Adversarial Multi-armed Bandit}

We consider a $K$-armed adversarial bandit problem where $K>1$ and the learner receives an arbitrary sequence of loss vectors $(\ell_1,\ldots,\ell_T)$ where $\ell_t\in [0,1]^K\forall t\in[T]$. On each round, the learner plays an action $I_t\in[K]$ and observes the loss $\ell_t(I_t)$. For adversarial bandits, a randomized policy is commonly used. In particular, given some policy $\pi$, the conditional distribution over the actions having observed $\Omega_{t-1}=\{(I_1,\ell_1),\ldots,(I_{t-1},\ell_{t-1})\}$ is $P_t=\pi(\cdot|\Omega_{t-1})\in\mathcal{P}_{K-1}$. The performance of a policy $\pi$ on the environment can be measured by the expected regret which is the expected loss of the policy relative to the best fixed action in hindsight:
\bqs
\E[R_T]=\E\bigg[\sum_{t=1}^T\ell_t(I_t)\bigg]-\min_{i\in[K]}\sum_{t=1}^T\ell_t(i)
\eqs

\paragraph{Exp3.}

The Exponential-weights for Exploration and Exploitation algorithm (Exp3) \citep{auer2002nonstochastic} is a popular algorithm for adversarial MAB. It follows the similar idea of prediction with expert advice and applies the Hedge (or Weighted-Majority) algorithm to the tradeoff of exploration and exploitation.
Specifically, we first define a probability vector $\mathbf p_t\in \R^K$ in which the $i-$th element $p_t(i)$ indicates the probability of drawing arm $i$ at time $t$. This vector is initialized uniformly and updated at each round. On each round, the learner plays an action by drawing $I_t\sim \mathbf p_t$ where $p_t$ is set as follows
\bqs
p_t(i) = (1-\gamma)\frac{w_t(i)}{\sum_{j=1}^K w_t(j)}  + \frac{\gamma}{K}, \forall i\in[K]
\eqs
where $w_t(i)$ is the importance weight of each arm $i$ learned by the Hedge algorithm, and $\gamma$ is a parameter for weighting the exploration term.
Algorithm \ref{alg:exp3} gives a summary of the Exp3 algorithm. By tuning the optimal parameter of $\gamma$, the Exp3 algorithm is able to achieve the regret $O(\sqrt{TK\ln K})$ in the adversarial setting.

\begin{algorithm}[ht]
\caption{Exp3}
\label{alg:exp3}
\begin{algorithmic}
\STATE \textbf{INPUT}: parameter $\gamma\in(0,1]$
\STATE \textbf{INIT}: $w_1(i)=1,\forall i\in[K]$\\
\FOR{ $t=1,2,\ldots,T$}
    \STATE Set $p_t(i) = (1-\gamma)\frac{w_t(i)}{\sum_{j=1}^K w_t(j)}  + \frac{\gamma}{K}, \forall i\in[K]$
    \STATE Play action by drawing $I_t\sim \mathbf p_t$
    \STATE Receive $\ell_t(I_t)\in[0,1]$,
    \STATE Update~$w_t(i) = w_t(i)e^{-\gamma\frac{\ell_t(i)}{p_t(i)}}$, $\textrm{if }i=I_t$.
\ENDFOR
\end{algorithmic}
\end{algorithm}

{\it Other Related Works.}  This is generally more challenging than the stochastic setting. A variety of algorithms have been explored in literature \citep{bubeck2012regret}. For example, the Exp3.P algorithm in \citep{auer2002nonstochastic} improves the loss estimation and probability update strategies to get a high probability bound. The Exp3.M algorithm in \citep{uchiya2010algorithms} explores the new problem setting of multiple plays.

\subsubsection{Linear Bandit and Combinatorial Bandit}

We first introduce the problem setting of the \emph{Linear Bandit} optimization problem \citep{auer2002using,rusmevichientong2010linearly,jun2017scalable}. During each iteration, the player makes its decision by choosing a vector from a finite set $\mathcal S \subseteq \R^d$ of elements $\mathbf v(i)$ for $i=1,...,k$. The chosen action at iteration $t$ is indexed as $I_t$. The environment chooses a loss vector $\bm\ell_t\in\R^d$ and returns the linear loss as $c_t(I_t)=\bm\ell_t^\top \mathbf v(I_t)$. Note that the player has no access to the full knowledge of loss vector and the only information revealed to the player is the loss of its own decision $c_t(I_t)$.
 Obviously, when setting $d=k$ and $\mathbf v(i)$ is the standard basis vector, this problem is identical to that in the previous section.

\paragraph{Combinatorial Bandit.} \emph{Combinatorial bandit} \citep{cesa2012combinatorial} is a special case of Linear Bandits, where $\mathcal S$ is a subset of binary hypercube $\{0,1\}^d$. The loss vector $\bm \ell_t$ may be generated from an unknown but fixed distribution, which is termed as stochastic combinatorial bandit, or chosen from some adversarial environment, which is termed as adversarial combinatorial bandit. The goal is to minimize the expected regret
\bqs
\bar{R}_T=\E[\sum_{t=1}^T c_t(I_t)]-\min_{i\in[K]}L_T(i)
\eqs
where $L_T(i)=\E \sum_{t=1}^T c_t(i)$ is the expected sum of loss for choosing action $i$ in all $T$ iterations, not a random variable in stochastic setting.
One example algorithm for Combinatorial Bandit is the COMBAND algorithm \citep{cesa2012combinatorial}. It first defines a sampling probability vector $\mathbf p_t\in \R^k$ for sampling $\mathbf v(I_t)$ from $\mathcal S$
\bqs
\mathbf p_t =(1-\gamma) \mathbf q_{t}+\gamma \bm\mu
\eqs
where $\mathbf q_{t}\in \R^k$ is the exploitation probability vector that is updated during all iterations to follow the best action, $\bm \mu \in \R^d$ is a fixed exploration probability, and $\gamma\in [0,1]$ is the weight to control the exploitation and exploration trade-off. The algorithm draws the action $I_t$ based on distribution $\mathbf p_t$ and gets the loss $c_t(I_t)$ from the environment. Second, an estimation of the loss vector $\bm \ell_t$ is calculated with the new information,
\bqs
\widetilde{\bm \ell_t}=c_t(I_t)P_t^+\mathbf v(I_t)
\eqs
where $P_t^+$ is the pseudo-inverse of the expected correlation matrix $\E_{\mathbf p_t}[\mathbf v \mathbf v^\top]$. Finally, the exploitation weights are scaled based on the estimated loss vector,
\bqs
\mathbf q_{t+1}(i)\propto \mathbf q_t(i)\exp(-\eta \widetilde{\bm \ell_t}^\top \mathbf v(i))
\eqs
where $\eta>0$ is a learning rate parameter and $\propto$ indicates that this scaling step is followed by a normalization step so that $\sum_{i=1}^k \mathbf q(i)=1$.
The COMBAND algorithm achieves a regret bound better than $O(\sqrt{T d \ln|\mathcal S|})$ for a variety of concrete choices of $\mathcal S$.

\paragraph{Other Related Works.} Recently, many studies also address linear bandits and combinatorial bandits in different settings. The ESCB algorithm  \citep{combes2015combinatorial} efficiently exploits the structure of the problem and gets a better regret bound of $O(\ln(T))$. The CUCB algorithm \citep{chen2016combinatorial} addresses the problem where the loss may be nonlinear. \citep{combes2015combinatorial} provided a useful survey for closely related works \citep{bubeck2012towards,cesa2012combinatorial} and gave a novel algorithm with promising bounds.

\subsection{Contextual Bandits}
Contextual Bandit is a widely used extension of MAB by associating contextual information with each arm \citep{zeng2016online,li2017provable}. For example, in personalized recommendation problem, the task is to select products that are most likely to be purchased by a user. In this case, each product corresponds to an arm and the features of each product are easy to acquire \citep{li2010contextual}.

In a contextual bandits problem, there is a set of policies $\mathcal F$, which may be finite or infinite. Each $f \in \mathcal F$ maps a context $\x\in\mathcal X \subseteq\R^d$ to an arm $i\in [k]$. Different from the previous setting where the regret is defined by competing with the arm with the highest expected reward, the regret here is defined by comparing the decision $I_t$ with the best policy $f^*=\arg\inf_{f\in\mathcal F}\ell_D(f)$, where $D$ is the data distribution.
\bqs
R_T(f)=\sum_{t=1}^T[\ell_{I_t,t}-\ell_t(f^*)]
\eqs

In literature, there are comprehensive surveys on contextual bandit algorithms in both stochastic and adversarial settings \citep{zhou2015survey,bubeck2012regret}. Below we focus on two settings of contextual bandits: multiclass classification and expert advice.

\subsubsection{The Multiclass Setting.}
In this setting, contextual bandit is regarded as a special case of online multi-class classification tasks in bandit setting. The goal is to learn a mapping from context space $\R^d$ to label space $\{1,...,k\}$ from a sequence of instances $\x_t \in \R^d$. Different from classic online multi-class classification problems where a class label $y_t\in\{1,...,k\}$ is revealed at the end of each iteration, in bandit setting, the learner only gets a partial feedback on whether $\hat y_t$ equals to $y_t$. In the following, we briefly review some representative works of contextual bandits for multi-class classification.

{\it Banditron} is the first bandit algorithm for online multiclass prediction~\citep{DBLP:conf/icml/KakadeST08}, which is a variant of the Perceptron. To efficiently make prediction and update the model, the Banditron algorithm keep a linear model $W^t$, which is initialized as $W^1=0\in\R^{k\times d}$. At the $t$-th iteration, after receiving the instance $\x_t\in\R^d$, it will first set
\bqs
\hat{y}_t=\arg\max_{r\in[k]}(W^t\x_t)_r
\eqs
where $(\z)_r$ denotes the $r$-th element of $\z$. Then the algorithm will define a distribution as
\bqs
\Pr(r)=(1-\gamma)\I(r=\hat{y}_t) + \gamma/k,\, \forall r\in[k]
\eqs
which roughly implies that the algorithm exploits with probability $1-\gamma$ and explores with the remaining probability by uniformly predicting  a random label from $[k]$. The parameter $\gamma$ controls the exploration-exploitation tradeoff. The algorithm then randomly sample $\tilde{y}_t$ according to the probability $\Pr$ and predicts it as the label of $\x_t$. After the prediction, the algorithm then receives the bandit feedback $\I(\tilde{y}_t=y_t)$. Then the algorithm uses this feedback to construct a matrix
\bqs
\tilde{U}^t_{r,j}= x_{t,j}\left(\I(\hat{y}_t=r)-\frac{\I(\tilde{y}_t=y_t)\I(\tilde{y}_t=r)}{\Pr(r)}\right)
\eqs
since its expectation satisfies $\E \tilde{U}^t_{r,j} = U^t_{r,j}= x_{t,j}\left(\I(\hat{y}_t=r)-\I(y_t=r)\right)$, where $U^t$ is actually a (sub)-gradient of the following hinge loss
\bqs
\ell(W; (\x_t, y_t)) = \max_{r\in[k]/\ \{y_t\}}[1- (W\x_t)_{y_t}+(W\x_{t})_r]_+
\eqs
where $[z]_+=\max(0, z)$. Then the algorithm will update the model by $W^{t+1}=W^t  - \tilde{U}^t.$
The Banditron algorithm is summarized in Algorithm~\ref{Banditron}.
\begin{algorithm}[hptb]
\caption{Banditron}\label{Banditron}
\begin{algorithmic}
\STATE \textbf{INIT}: $\mathbf{w}_{1,1}=0,..,\mathbf{w}_{k,1}=0 $\\
\FOR{ $t=1,2,\ldots,T$}
    \STATE Receive an incoming instance $\mathbf{x}_t$
    \STATE $P(r)=(1-\gamma)\bm 1[r=\arg\max_i \w_{i,t}^\top \x_t]+\frac{\gamma}{k}$.
    \STATE Sample $\hat y_t$ according to $P(r), r\in\{1,...,k\}$
    \STATE  $\u_r=\x_t(\frac{\bm1 [y_t=\hat y_t=r]}{P(r)}-\bm 1[r=\arg\max_i \w_{i,t}^\top \x_t])$;
    \STATE $\w_{r,t+1}=\w_{r,t}+\u_r$
\ENDFOR
\end{algorithmic}
\end{algorithm}

This algorithm achieves $O(\sqrt{T})$ in linear separable case and $O(T^{\frac{2}{3}})$ in inseparable case.

{\it Other Related Work.} Following the Banditron, many algorithms have been proposed. For example, Bandit Passive Aggressive (Bandit PA) follows the PA learning principle and adopts the framework of one against all others to make prediction and update the model~\citep{DBLP:conf/icdm/ChenCZCZ09}. In general, some update principles are based on first-order gradient descent \citep{wang2010potential}, while others adopt second order learning \citep{DBLP:conf/icml/CrammerG11,hazan2011newtron,zhang2016online,beygelzimer2017efficient}. Most of these algorithms explore the $k$ classes uniformly with probability $\gamma$, while \citep{DBLP:conf/icml/CrammerG11} sample the classes based on the Upper Confidence Bound.

\subsubsection{The Expert Setting.}

We now introduce a well-known algorithm called Exp4 \citep{auer2002nonstochastic} for contextual bandits in the expert settings. The Exp4 algorithm assumes that there are $N$ experts who will give advice on the distribution over arms during all iterations. $\xi_{i,t}^n$ indicates the probability of picking arm $i\in [K]$ recommended by expert $n\in [N]$ during time $t\in[T]$. Obviously, $\sum_{i=1}^k\xi_{i,t}^n=1$. During time $t$, the true reward vector is denoted by $\mathbf r_t\in[0,1]^K$. Thus the expected reward of expert $n$ is $\bm \xi_{t}^n \cdot \mathbf r_t$. The regret is defined by comparing with the expert with the highest expected cumulative reward.
\bqs
R_t=\max_{n\in[N]} \sum_{t=1}^T \bm \xi_{t}^n \cdot \mathbf r_t-\E \sum_{t=1}^T r_{t,I_t}
\eqs
The Exp4 algorithm first defines a weight vector $\w_t\in\R^N$ that indicates the weights for the $N$ experts. We set the weight as $\w_0=\bm 1$ and update it during each iteration.
During iteration $t$, we calculate the probability of picking arm $i$ as the weighted sum of advices from all $N$ experts,
\bqs
p_{i,t}=(1-\gamma) \frac{\sum_{n=1}^N w_{n,t}\xi_{i,t}^n}{\sum_{n=1}^N w_{n,t}}+\frac{\gamma}{K}
\eqs
where $\gamma\in[0,1]$ is a parameter to balance exploitation and exploration. We then draw an arm $I_t$ according to probability $p_{i,t}$ and calculate an unbiased estimator of 
$
\hat r_{i,t}=\frac{r_{i,t}}{p_{i,t}}\I_{i=I_t},
$
which will be used to calculate the expected reward. Finally the weight $\w_t$ is updated according to the expected reward of each arm. The Exp4 algorithm is able to achieve the regret bound $O(\sqrt{TK\ln N})$ as shown in \citep{auer2002nonstochastic} and tighter bounds were also given in \citep{mcmahan2009tighter}.


{\it Other Related Work.} Another general contextual bandit algorithm is the epoch-greedy algorithm in \citep{langford2008epoch} that is similar to $\epsilon$-greedy with shrinking $\epsilon$. This algorithm is computationally efficient given an oracle optimizer but has the weaker regret guarantee of $O(T^{2/3})$.  LinUCB \citep{chu2011contextual} is an extension of UCB to contextual bandit problem, by assuming that there is a feature vector $\x_{t,i}\in\R^d$ at time $t$ for each arm $i$. Similar to the UCB algorithm, a model is learnt to estimate the upper confidence bound of each arm $i\in[k]$ given the input of $\x_{t,i}$. The algorithm simply chooses the arm with the highest UCB. The LinREL algorithm \citep{auer2002using} is similar to LinUCB in that it adopts the same problem setting and same maximizing UCB strategy. While, a different regularization term is used which leads to a different calculation of the UCB.

\subsection{Other Bandit Variants}
In literature, there are many other studies addressing on various types of bandit variants. We refer readers for more comprehensive studies on bandit topics in \citep{bubeck2012regret}. Below we briefly introduce a few other major variants.

Other than stochastic bandits and adversarial bandits, another fundamental topic of multi-armed bandits is called ``{\it Markovian bandits}", which generally assumes the reward processes are neither i.i.d. (like in stochastic MAB) nor adversarial. Specifically, arms are associated with $K$ Markov processes, each with its own state space. On each round, an arm is chosen in some state, a stochastic reward is drawn from some probability distribution, and the state of the reward process for the arm changes in a Markovian fashion, based on an underlying stochastic transition matrix. Both reward and new state are revealed to the player. The seminal work of \cite{gittins1979bandit} gives an optimal greedy policy that can be computed efficiently. A special case of Markovian bandits is Bayesian bandits \citep{scott2010modern,gittins2011multi,kaufmann2012bayesian}, which are parametric stochastic bandits where the parameters of the reward distributions are assumed to be drawn from known priors, and the regret is computed by also averaging over the draw of parameters from the prior.

Another topic is to study {\it infinitely many-armed bandits} problems where the number of arms can be larger than the possible number of experiments or even infinite \citep{berry1997bandit,wang2009algorithms,bubeck2011x}. Among these studies, one niche sub-topic is {\it continuum-armed bandits} \citep{kleinberg2005nearly,KMAB17}, where the arms lie in some Euclidean (or metric) space and their mean-reward is a deterministic and smooth (e.g., Lipschitz) function of the arms, a.k.a. {\it Lipschitz Bandit} \citep{magureanu2014lipschitz}.


\def \q {\mathbf{q}}
\def \u {\mathbf{u}}
\def \w {\mathbf{w}}
\def \x {\mathbf{x}}

\def \I {\mathbb{I}}
\def \R {\mathbb{R}}
\def \E {\mathbb{E}}

\def \pr {\mathrm{Pr}}
\def \sign {\mathrm{sign}}

\section{Online Active Learning}

\subsection{Overview}

In a standard online learning task (e.g., online binary classification), the learner receives and makes prediction for a sequence of instances generated from some unknown distribution. At the end of every round, it always assumes the learner will receive the true label (feedback) from the environment. For many real-world applications, obtaining the labels could be very expensive, and sometimes it is not always necessary/informative to query the true labels of every instance, e.g., if an instance is correctly classified with a high confidence. Motivated to address this challenge, online active learning is a special class of online learner that observes a sequence of unlabeled instances each time deciding whether to query the label of the incoming instance; if the label is queried, then the learner can use the labelled instance to update the prediction model; otherwise, the model will be kept unchanged.

In literature, there are two major kinds of settings for online active learning. One is called the ``{\it selective sampling}" setting~\citep{NIPS1989SS,freund1997selective,DBLP:conf/colt/Cesa-BianchiCG03} by adapting classical online learning for active learning. The other is {\it online active learning with expert advice} by adapting the setting of prediction with expert advice for active learning \citep{helmbold1997some,DBLP:conf/uai/ZhaoHZ13}. Both operate in the similar problem settings where true label of an instance is only queried when some condition is satisfied, e.g., predictive confidence is below some threshold.



\subsection{Selective Sampling Algorithms}

In this section we review a family of popular Selective Sampling (SS) algorithms for online active learning tasks. In the following discussions, we use a typical online binary classification task as a running example. For notation, an example is a pair $(\mathbf{x},y)$, where $\mathbf{x}\in\mathbb{R}^d$ is an instance vector and $y\in\{-1,+1\}$ is the binary class label.
Assume the learning proceeds in a sequence of $T$ rounds, where $T$ may not be known in advance. On each round $t$, a learner observes an instance $\mathbf{x}_t$, then outputs a prediction $\hat{y}_t\in\{-1,+1\}$ as the label for the instance, and then decides whether or not to query the label $y_t$. Whenever $\hat{y}_t \neq y_t$, the learner's prediction outcome is considered as a mistake, no matter if it has decided to query the label or not. For notation, we denote $M_t=\I(\hat{y}_t \neq y_t)\in\{0,1\}$ as an indicator whether the learner makes a mistake at round $t$. For most cases, we also assume the leaner adopts a linear model to predict the class label using $\hat{y}_t=\sign(\hat{p}_t)$, where $\hat{p}_t=\w_t^\top\x_t$.

\subsubsection{First-order Selective Sampling Algorithms}


\paragraph{Selective-sampling Perceptron.}

\begin{algorithm}[htb]
\label{alg:ss-perceptron}
\caption{Selective-sampling Perceptron}
\begin{algorithmic}
\STATE \textbf{INPUT:} parameter $\delta>0$ \\
\STATE \textbf{INIT}: $\mathbf{w}_0=(0,\ldots,0)^{\top}$\\
\FOR{ $t=1,2,\ldots,T$}
    \STATE Observe an input instance $\mathbf{x}_t\in\mathbb{R}^d$
    \STATE Predict $\hat{y}_t=\sign(\hat{p}_t)$, where $\hat{p}_t=\w_t^\top\x_t$
    \STATE Draw a Bernoulli random variable $Z_t\in\{0,1\}$ of probability $\frac{\delta}{\delta+|\hat{p}_t|}$
    \STATE IF $Z_t=1$ THEN
    \STATE \quad\quad Query label $y_t\in\{-1,+1\}$ and Update $\mathbf{w}_t$ by Perceptron: $\w_{t+1}=\w_t + M_t y_t \x_t.$
\ENDFOR
\end{algorithmic}
\end{algorithm}

This algorithm~\citep{DBLP:journals/jmlr/Cesa-BianchiGZ06a} decides whether or not to query the label $y_t$ through a simple randomized rule: drawing a Bernoulli random variable $Z_t\in\{0, 1\}$ with probability
\bqs
\Pr(Z_t=1)=\frac{\delta}{\delta+|\hat{p}_t|}
\eqs
where $\delta>0$ is a smooth parameter that can be used to control the number of labels queried during the online active learning process. If $\delta$ increases, the number of queried labels increases. If $Z_t=1$, then the label $y_t$ of $\x_t$ will be queried, and the model will be updated using the Perceptron rule. Algorithm \ref{alg:ss-perceptron} gives a summary of the Selective-sampling Perceptron algorithm.
%

In theory, assuming $\|\x_t\|\le R$, for any $\w\in\R^d$, the expected number of mistakes of the Selective-sampling Perceptron algorithm can be bounded as:
\bqs
\E[\sum^T_{t=1}M_t]\leq (1+\frac{R^2}{2\delta})\frac{\bar{L}_{\gamma,T}(\w)}{\gamma} + \frac{\|\w\|^2(2\delta+R^2)^2}{8\delta\gamma^2}.
\eqs
where $\bar{L}_{\gamma,T}(\w)=\E[\sum^T_{t=1}Z_t M_t\ell_{\gamma,t}(\w)]$, and $\ell_{\gamma,t}(\w)=\max(0, \gamma- y_t\w^\top\x_t)$. Furthermore, the expected number of labels queried by the algorithm equals $\sum^T_{t=1}\E[\frac{\delta}{\delta+|\hat{p}_t|}]$. This bound depends on the value of the parameter $\delta$. By choosing the optimal value of $\delta$ as
\bqs
\delta = \frac{R^2}{2}\sqrt{1+\frac{4\gamma^2}{\|\w\|^2R^2}\frac{\bar{L}_{\gamma,T}(\w)}{\gamma}}
\eqs
the expected number of mistakes can be bounded
\bqs
\E[\sum^T_{t=1}M_t]\leq \frac{\bar{L}_{\gamma,T}(\w)}{\gamma}+\frac{\|\w\|^2 R^2}{2\gamma^2}+\frac{\|\w\|R}{\gamma}\sqrt{\frac{\bar{L}_{\gamma,T}(\w)}{\gamma}+\frac{\|\w\|^2R^2}{4\gamma^2}}
\eqs
This is an expectation version of the mistake bound for the standard Perceptron Algorithm. Especially, in the special case when the data is linearly separable, the optimal value of $\delta$ is $R^2/2$ and this bound becomes the familiar Perceptron bound $(\|\w\|R)^2/\gamma^2$.
Instead of using a fixed constant parameter, \citep{DBLP:journals/jmlr/Cesa-BianchiGZ06a} also proposed an adaptive parameter version of the selective sampling Perceptron algorithm as follows:
\bqs
\hspace{-0.3in}\Pr(Z_t=1) = \frac{\delta_t}{\delta_t+|\hat{p}_t|},\quad \textrm{s.t.}\quad \delta_t=\beta(R')^2\sqrt{1+\sum^{t-1}_{i=1}Z_iM_i},
\eqs
where $\beta>0$ is a predefined parameter, $R'=\max{R_{t-1},\|\x_t\|}$, $R_{t-1}=\max\{\|\x_i\| | Z_iM_i=1\}$.

\paragraph{Other first-order approaches.}

Instead of using Perceptron, the Passive-Aggressive Active learning algorithms in~\citep{DBLP:conf/acml/LuZhaoHoi14,lu2016online} are selective sampling algorithms by extending the framework of PA online learning algorithms. They also extended their algorithms for multi-class classification and cost-sensitive classification tasks. \citep{zhao2013costactive} proposed a cost-sensitive online active learning approach that directly optimizes cost-sensitive measures using PA-like algorithms for class-imbalanced classification tasks.

\subsubsection{Second-order Selective Sampling Algorithms}

\paragraph{Selective-sampling Second-order Perceptron.}

Instead of using the standard Perceptron algorithm, \citep{DBLP:journals/jmlr/Cesa-BianchiGZ06a} also proposed a selective-sampling algorithm based on the Second-order Perceptron.

\begin{algorithm}[htb]
\label{alg:ss-sop}
\caption{Selective-sampling Second-order Perceptron}
\begin{algorithmic}
\STATE \textbf{INPUT:} parameter $\delta>0$ \\
\STATE \textbf{INIT}:  $A_0=I$, $\mathbf{w}_0=(0,\ldots,0)^{\top}$\\
\FOR{ $t=1,2,\ldots,T$}
    \STATE Observe an input instance $\mathbf{x}_t\in\mathbb{R}^d$
    \STATE Computer $\hat{p}_t = [(A_t+\x_t\x_t^\top)^{-\frac{1}{2}}\u_t]^\top[(A_t+\x_t\x_t^\top)^{-\frac{1}{2}}\x_t] = \u_t^\top(A_t+\x_t\x_t^\top)^{-1}\x_t$
    \STATE Predict  $\hat{y}_t=\sign(\hat{p}_t)$
    \STATE Draw a Bernoulli random variable $Z_t\in\{0,1\}$ of probability $\frac{\delta}{\delta+|\hat{p}_t|}$
    \STATE IF $Z_t=1$ THEN
    \STATE \quad\quad Query label $y_t\in\{-1,+1\}$ and Update $\mathbf{w}_t$ by Second-order Perceptron:
    \STATE \quad\quad $\u_{t+1}=\u_t + M_t y_t \x_t,$~and~$A_{t+1} = A_t + M_t \x_t\x_t^\top$
\ENDFOR
\end{algorithmic}
\end{algorithm}
Let $\u_t$ denote the weight vector computed by standard Perceptron, and $A_t= I + \sum_{i\le t-1, Z_iM_i=1} \x_i\x_i^\top$ denote the correlation matrix over the mistaken trials plus an identity matrix $I$, then the second-order Perceptron predicts the label of current instance $\x_t$ as
\bqs\begin{aligned}
\hspace{-0.1in}\hat{y}_t =\sign(\hat{p}_t), \mathrm{where}~\hat{p}_t = [(A_t+\x_t\x_t^\top)^{-\frac{1}{2}}\u_t]^\top[(A_t+\x_t\x_t^\top)^{-\frac{1}{2}}\x_t] = \u_t^\top(A_t+\x_t\x_t^\top)^{-1}\x_t
\end{aligned}\eqs
The second-order algorithm differs from standard Perceptron in that, before each prediction, a linear transformation $(A_t+\x_t\x_t^\top)^{-1/2}$ is applied to both current Perceptron weight $\u_t$ and current instance $\x_t$. After prediction, the query strategy of this algorithm is the same with the previous selective sampling: draw a Bernoulli random variable $Z_t\in\{0, 1\}$ with
\bqs
\Pr(Z_t=1) = \frac{\delta}{\delta + |\hat{p}_t|}.
\eqs
Algorithm~\ref{alg:ss-sop} gives a summary of the Selective-sampling Second-order Perceptron algorithm.
In theory, if the algorithm runs on a sequence of $T$ rounds, for any $\w$, the expected number of mistakes made by the algorithm is bounded:
\bqs\begin{aligned}
\E[\sum^T_{t=1}M_t]	\le \frac{\bar{L}_{\gamma,T}(\w)}{\gamma}+\frac{\delta}{2\gamma^2}\w^\top\E[A_T]\w + \frac{1}{2\delta}\sum^d_{i=1}\E\ln(1+\lambda_i)
\end{aligned}\eqs
where $\bar{L}_{\gamma,T}(\w)=\E[\sum^T_{t=1}Z_tM_t\ell_{\gamma,t}(\w)]$ with $\ell_{\gamma,t}(\w)=\max(0, \gamma- y_t\w^\top\x_t)$, $\lambda_1,\ldots,\lambda_d$ are the eigenvalues of the random correlation matrix $\sum^T_{t=1}Z_tM_t\x_t\x_t^\top$ and $A_T=I+\sum^T_{t=1}M_tZ_t\x_t\x_t^\top$. Moreover, the expected number of queries by the algorithm equals $\sum^T_{t=1}\E[\frac{\delta}{\delta+|\hat{p}_t|}]$. Furthermore, by setting $\delta = \gamma\sqrt{\frac{\sum^d_{i=1}\E\ln(1+\lambda_i)}{\w^\top\E[A_T]\w}}$, it leads to the optimal bound
\bqs
\hspace{-0.3in}\E[\sum^T_{t=1}M_t]\le\frac{\bar{L}_{\gamma,T}(\w)}{\gamma} + \frac{1}{\gamma}\sqrt{(\w^\top\E[A_T]\w)\sum^d_{i=1}\E\ln(1+\lambda_i)}
\eqs
\citep{DBLP:journals/jmlr/Cesa-BianchiGZ06a} also proposed an improved selective-sampling algorithm based on second-order Perceptron, which modifies the sampling probability by incorporating the second-order information, i.e.,
with
\bqs
\Pr(Z_t=1) = \frac{\delta}{\delta + |\hat{p}_t| + \frac{1}{2}\hat{p}_t^2(1+\x_t^\top A_t^{-1}\x_t)},
\eqs

\paragraph{Other second-order approaches.}
\cite{DBLP:conf/colt/Cesa-BianchiCG03} proposed a margin-based selective sampling algorithm which also exploits second-order information in the model:
\bqs
\hat{y}_t=\sign(p_t),~~\textrm{where}\quad p_t=\w_t^\top\x_t,~~\w_t= A_t^{-1}\u_t,~~\u_t=\sum^{t-1}_{i=1}Z_i y_i\x_i^\top,~~A_t=(I+\sum^{t-1}_{i=1}Z_i \x_i\x_i^\top)
\eqs
But the query strategy is a margin-based sampling approach without explicitly exploiting the second-order information:
\bqs
\Pr(Z_{t+1}=1)=\I(p_t\le\frac{4\ln t}{\sum^{t-1}_{i=1}Z_i})
\eqs
\cite{hao2016soal,hao2017second} proposed second-order online active learning algorithms by fully exploiting both the first-order and second-order information for online active learning tasks and also gave cost-sensitive extensions for class-imbalanced tasks.

\subsubsection{Other Selective Sampling Approaches}
There are also a few other selective sampling approaches in which the base classifier is based on the Regularized Least Squares (RLS).
In particular, on each round $t$, the linear classification model can be updated by the RLS estimate
\bqs
\mathbf{w}_t = (I+S_{t-1}S_{t-1}^{\top}+\mathbf{x}_t\mathbf{x}^{\top})^{-1}S_{t-1}Y_{t-1}
\eqs
where matrix $S_{t-1}=[\mathbf{x}_1',\ldots,\mathbf{x}_{N_{t-1}}']$ is the collection of $N_{t-1}$ instances queried up to time $t-1$, and the vector $Y_{t-1}=(Y_1',\ldots,Y_{N_{t-1}}')$ is the set of queried labels for the instances. The selective sampling algorithms that follow this paradigm include the Bound on Bias Query (BBQ) algorithms \citep{DBLP:conf/icml/Cesa-BianchiGO09,DBLP:conf/icml/OrabonaC11} and their improved variants \cite{DBLP:conf/colt/DekelGS10,DBLP:conf/icml/OrabonaC11}.
A major drawback of these methods is that the RLS-based base learner is more like a fashion of batch learner instead of truly online learning, and thus the overall learning scheme might be inefficient or non-scalable if the number of queried labeled examples can be large.

\subsection{Online Active Learning with Expert Advice}

The idea of online active learning with expert advice dates back to classical Query by Committee (QBC) in \citep{seung1992query,freund1997selective}, where the idea is to query the label of an instance based on the principle of \textit{maximal disagreement} among a set of experts, i.e., the confidence criteria in this case is how much the expert hypotheses disagree on their evaluation of instance predictions. QBC bounds from below the average information gain provided by each requested label. \cite{baram2004online} considers the setting of how to online combine an ensemble of active learners, which is executed based on a maximum entropy criterion. Another perhaps more dominating line of studies in \citep{helmbold1997some,cesa2005minimizing,DBLP:conf/uai/ZhaoHZ13} explore the exponentiated weighted average forecaster for online active learning tasks, where an instance is stochastically queried based on the available feedback on the importance of each expert in the pool.

Next we describe in detail one of the most recent approaches for online active learning with expert advice in \citep{DBLP:conf/uai/ZhaoHZ13}. Consider an unknown sequence of instances $\x_1,\ldots,\x_T \in\mathbb{R}^d$, a ``forecaster" aims to predict the class labels of every incoming instance $\x_t$. The forecaster sequentially computes its predictions based on the predictions from a set of $N$  ``experts". Specifically, at the $t$-th round, after receiving an instance $\x_t$, the forecaster first accesses the predictions  of the  experts $\{f_{i,t}: \mathbb{R}^d \rightarrow [0,1]| i=1,\ldots,N\}$, and then computes its own prediction $p_t\in[0,1]$ based on the predictions of the $N$ experts. After $p_t$ is computed, the true outcome $y_t\in\{0,1\}$ is disclosed. To solve this problem, the ``Exponentially Weighted Average Forecaster" (EWAF) makes the following prediction:
\begin{eqnarray}
p_t = \frac{\sum^N_{i=1}\exp(-\eta L_{i,t-1})f_i(\x_t)}{\sum^N_{i=1}\exp(-\eta L_{i,t-1})},
\end{eqnarray}
where $\eta$ is a learning rate, $L_{i,t} = \sum_{j=1}^t \ell(f_i(\x_j), y_j),\quad L_{t} = \sum_{j=1}^t \ell\left(p_j, y_j\right)$ with  $\ell(p_t, y_t) = |p_t - y_t|$.
Unlike the above regular learning, in an active learning with expert advice task, the outcome of an incoming instance is {\it only} revealed whenever the learner  requests the label from the environment/oracle. To solve this problem, binary variables $z_s \in \{0, 1\}, s=1, \ldots, t$ are introduced to indicate if an active forecaster has requested the  label of an instance at $s$-th trial. $\widehat{L}_{i,t}$ is used to denote the loss function experienced by the active learner w.r.t. the $i^{th}$ expert, i.e., $\widehat{L}_{i,t} = \sum_{s=1}^t \ell(f_i(\x_s), y_s) z_s$. For this problem setting, \cite{DBLP:conf/uai/ZhaoHZ13} proposed a general framework of active forecasters, as shown in Algorithm \ref{alg:oalea}.
\begin{algorithm}[hptb]
\label{alg:oalea}
\caption{Online Active Learning with Expert Advice}
\begin{algorithmic}
\STATE \textbf{INPUT:} a pool of experts $f_i,\ i=1,\ldots, N$.\\
\STATE \textbf{ INIT:} tolerance threshold $\delta$ and $\widehat{L}_{i,t}=0,\ i\in[N]$. \\
\FOR{ $t=1,2,\ldots,T$}
    \STATE Receive $\x_t$ and compute $f_i(\x_t)$, $i\in[N]$;
    \STATE Compute  $\hat{p}_t=\frac{\sum^N_{i=1}\exp(-\eta \widehat{L}_{i,t-1})f_i(\x_t))}{\sum^N_{i=1}\exp(-\eta \widehat{L}_{i,t-1})}$;
    \STATE If a {\it confidence condition} is not satisfied\\
    ~~~~\indent request label $y_t$ and update $\widehat{L}_{i,t}=\widehat{L}_{i,t-1}+\ell(f_i(\x_t), y_t),\ i\in[N]$;\\
\ENDFOR
\end{algorithmic}
\end{algorithm}

At each round, after receiving an instance $\x_t$, we compute the prediction of class label for the instance by aggregating the prediction of each expert in the pool, i.e., $f_i(\x_t)$. Then, we examine if a confidence condition is satisfied. If so, we will skip the label request; otherwise, the learner will request the class label for this instance from the environment. To decide when to request the class label or not, the key idea is to seek a confidence condition by estimating the difference between $p_t$ and $\hat{p}_t$. Intuitively, the smaller the difference, the more confident we have for the prediction made by the forecaster. More specifically, the work in \citep{DBLP:conf/uai/ZhaoHZ13} proved that for a small constant $\delta>0$, $\max_{1\le i, j\le N} |f_i(\x_t)-f_j(\x_t)|\le \delta$ implies $|p_t-\widehat{p}_t|\le \delta$. This roughly means that, if any two experts do not disagree with each other too much on the instance, then we can skip requiring its label.

In addition to the above work, there are also a few other active learning strategies for online learning with expert advices, for example the active greedy forecaster~\citep{DBLP:conf/uai/ZhaoHZ13}. Online active learning with expert advice can be applied in some real-world applications, e.g., crowdsourcing tasks \citep{hao2015active} where  the learner attempts to address both the diverse quality of annotators' performance with expert learning and efficient annotation in seeking informative data using active learning.



\section{Online Semi-supervised Learning}
\subsection{Overview}

Semi-Supervised Learning (SSL) has been an important class of machine learning tasks and techniques, which aims to make use of unlabeled data for learning tasks. It has been extensively studied mostly in the settings of batch learning and some comprehensive surveys can be found in \citep{zhu2006semi,zhu2009introduction}. When online learning meets semi-supervised learning, there are two major branches of research. One major branch of studies is to turn traditional batch semi-supervised learning methods into online algorithms such that they can work from data streams of both labeled and unlabeled data, which we call this setting as ``Online Semi-supervised Learning" and we will review a popular framework of ``online manifold regularization". The other branch of studies is to study classical online learning tasks in transductive learning settings (e.g., by assuming unlabeled data can be made available before online learning tasks), which we call this setting as ``Transductive Online Learning". We note that online active learning as introduced previously can be generally viewed as a special type of online semi-supervised learning where an online learner has to deal with both labeled and unlabeled data.

\subsection{Online Manifold Regularization}

In the area of semi-supervised learning, one major framework for semi-supervised learning is based on manifold regularization \citep{belkin2006manifold}, where the learner not only minimizes the loss on the labeled data, but also minimizes the difference of predictions on the unlabeled instances which are similar on the manifold. Specifically, consider instances $(\x_t,y_t)$, $t\in \{1,...T\}$, the idea is to minimize the following objective function
\bqs
&&J(f)=\frac{1}{l}\sum_{t=1}^T\delta(y_t)\ell(f(\x_t),y_t)+\frac{\lambda_1}{2}||f||^2+\frac{\lambda_2}{2T}\sum_{s,t=1}^T(f(\x_s)-f(\x_t))w_{st}
\eqs
where the first term is the loss on labeled instances where $\delta(y_t)=1$ if and only if $y_t$ exists and $l$ is the number of labeled data, the second term is a classic regularization term for supervised learning, and the last term is the manifold regularization on unlabeled data.

In literature, online manifold regularization has been explored \citep{goldberg2008online}, which attempts to turn batch manifold regularization algorithms into online/incremental algorithms. Specifically, the above objective can be returned online for each instance:
\bqs
&&J_t(f)=\frac{T}{l}\delta(y_t)\ell(f(\x_t),y_t)+\frac{\lambda_1}{2}||f||^2+\lambda_2\sum_{i=1}^{t-1}(f(\x_i)-f(\x_t))w_{it}
\eqs
It can be solved using Online Gradient Descent in $O(T^2)$ time. Unfortunately, such straightforward solution is expensive in both time and space, since the calculation of the last term requires to store all instances and measure the similarity $w_{it}$ between the incoming instances and all existing ones.

To address this problem, the authors offer two sparse approximations of the objective function. The first solution is not to keep all instances but to keep only the newest $\tau$ ones, where $\tau$ is the buffer size. This strategy is simple but not very efficient since the discarded old instances may contain important information. The second solution adopts a random projection tree to find $s$ cluster centers during online learning. Finally, instead of calculating the similarity between $\x_t$ and all existing instances, the algorithm only consider the $s$ cluster centers as the most representative instances.

In addition to the above work, \citep{valko2010online} proposed a fast approximate algorithm for online semi-supervised learning. which leverages the incremental k-center quantization method to group neighboring points so as to yield a set of reduced representative points, and as a result an approximate
similarity graph can be constructed to find the harmonic solution in semi-supervised learning \citep{zhu2003semi}.

Finally, there were some related efforts on online active semi-supervised learning \citep{goldberg2011oasis}, which extends active learning in the online semi-supervised learning settings. For example, following such kind of setting, \citep{goldberg2011oasis} developed the OASIS algorithm by using a general online Bayesian learning framework.



\subsection{Transductive Online Learning}

Transductive online learning \citep{cesa2013efficient,ben1997online} is a niche class of online learning tasks, where we want to learn from an arbitrary sequence of
labeled examples $(\mathbf{x}_1,y_1),\ldots,(\mathbf{x}_T,y_T)$ by making the assumption that the set of unlabeled instances $(\mathbf{x}_1,\ldots,\mathbf{x}_T)$ can be given in advance to the learner before an online learning task begins. In particular, the work \citep{cesa2013efficient} proposed an efficient algorithm based on the principle of prediction with expert advice by combining ``random playout" and ``randomized rounding" of loss subgradients. We note that this niche topic has received very few attention, possibly because of their assumption of obtaining unlabeled data in advance, which may be unrealistic in many applications. However, the studies in this niche family of studies may provide some theory insights
about the linkage between online learning and batch learning as demonstrated in \citep{cesa2013efficient}.

\def \q {\mathbf{q}}
\def \u {\mathbf{u}}
\def \w {\mathbf{w}}
\def \x {\mathbf{x}}

\def \I {\mathbb{I}}
\def \R {\mathbb{R}}
\def \E {\mathbb{E}}

\def \pr {\mathrm{Pr}}
\def \sign {\mathrm{sign}}

\section{Online Unsupervised Learning}

\subsection{Overview}

In this section we briefly review some key work in the literature of online unsupervised learning, where models are learned from unlabeled data streams where no explicit feedback is available. Broadly, we categorize the existing work into four major groups: Online Clustering, Online Dimension Reduction, Online Anomaly Detection, and Online Density Estimation. Due to the vast number of ways in which unsupervised learning in online settings have been explored in literature, and numerous applications for which algorithms are designed, it is almost impossible to make a comprehensive treatment on this topic in this survey. Instead, we try to focus on the key areas and give a general overview of the main ideas in each area which are closely related to online learning.

\subsection{Online Clustering}

Clustering is an unsupervised learning process of grouping unlabeled data instances such that instances in the same group are similar, and instances between groups are dissimilar. It gives an effective mechanism to summarize the data, and does not require labels of the instances in order to perform the clustering. For batch learning settings, clustering is usually classified into following categories: partition based clustering, hierarchical clustering, density-based clustering, and grid-based clustering \citep{berkhin2006survey}.
For online settings \citep{aggarwal2013survey}, partition-based and density-based clustering have been studied more extensively. In the following we briefly review some of online learning approaches for clustering on streaming data especially for these two categories.

\emph{Partitioning Based} clustering methods split the instances into partitions where each partition represents a cluster. The partitions are designed on the basis of some distance measures (e.g. Euclidean distance). The number of clusters is usually pre-defined by the user. The most popular algorithms in this category are those based on \emph{k-MEANS} and \emph{k-MEDOIDS} algorithm. The k-MEANS algorithm is one of the oldest and most popular clustering methods, where the idea is to identify $k$ centroids, where each centroid corresponds to one out of $k$ clusters by minimizing the sum of square errors between each instance to their corresponding centroids.
Sequential algorithms performing k-MEDOID or k-MEDIAN clustering usually try to break the stream of instances into chunks where the size of each chunk is set based on some pre-specified memory budget. Given a data stream $D$, it is broken into several chunks denoted by $D_1, D_2, \dots, D_t, \dots$ where each chunk contains at most $m$ instances, where $m$ is the budget of the chunks. In such a case, k-MEDIANS can be directly applied to each chunk. This framework is called the STREAM framework. \citep{guha2000clustering, o2002streaming,guha2003clustering}. There are also sampling approaches designed for clustering when the data streams are extremely large \citep{kaufman2008clustering}. Another method is the StreamKM++ \citep{ackermann2012streamkm}. In this approach, first, an adaptive non-uniform sampling approach is used to obtain small coresets from the data streams. The coreset construction is done by the utilization of coreset tree proposed in this paper which helps in significant speed up.

\emph{Density-based clustering}. Most clustering techniques suffer from several drawbacks. First, many of them (e.g. k-MEANS) are designed for only spherical clusters and can not adapt to arbitrary cluster shapes. In addition, the value of $k$, or the number of clusters has to be known a priori. Lastly, these methods are susceptible to outliers. Density based clustering algorithms (most popularly DBSCAN and its variants) are able to address all these challenges. Density based approaches cluster dense regions which are separated by sparse regions. A cluster based on density can take on arbitrary shapes, does not require prior knowledge of the number of clusters, and is robust to outliers in the data. However, performing density based clustering on streaming data in an online manner is plagued with several challenges including dynamic evolution of the clusters, limited memory space, etc. Following \citep{aggarwal2013survey,amini2014density}, we categorize the online density-based clustering algorithms into \emph{Micro-clustering Algorithms} and \emph{Grid-based clustering Algorithms}. The micro-clustering algorithms aim to summarize a data in an online manner, and the clustering is performed using these summaries \citep{cao2006density,tasoulis2007visualising,ruiz2009c,li2009three,ren2009density,ntoutsi2012density}. Grid-based methods, divide the entire instance space into grids, and each instance upon arrival is assigned a grid, and then the clustering is then done based on the density of the grids \citep{gao2005incremental,chen2007density,jia2008grid,tu2009stream,wan2009density,ren2011clustering,amini2012dengris,bhatnagar2014clustering}.

\emph{Other Clustering Methods}. \emph{Hierarchical Clustering} is a paradigm in which either a bottom-up approach or a top-down approach is used to gradually agglomerate the data points together. This results in a tree of clusters, which is also called a dendogram. Among the earliest approaches to incremental hierarchical clustering was CobWeb \citep{fisher1987knowledge}, which determines how to insert a new data point into the tree structure based on a category utility criteria. Recent hierarchical clustering algorithms include the ClusTree \citep{kranen2011clustree}, which offers a compact self adapting index structure for storing stream summaries in addition to giving more importance to recent data, and Perch \citep{kobren2017online}, which allows the clustering to scale to a large number of data points and clusters. \citep{tu2012density} propose an incremental approach to do Hierarchical clustering of the data, in addition to accounting for variance and density of the data. Some techniques have been developed for online clustering for very high dimensional data where the data sparsity makes it very hard to perform clustering as many instances tend to be equidistant from one another. {HPStream}\citep{aggarwal2004framework} introduces a concept of projected clustering to data streams. There are online clustering algorithms for other specific scenarios such as clustering of discrete and categorical streams, text streams, uncertain data streams, graph streams as well as distributed clustering \citep{aggarwal2013survey}.

\subsection{Online Dimension Reduction}

When the feature dimensions are very large, Dimension Reduction (DR) techniques can be used to improve learning efficiency, compress original data, visualize data better, and improve its applicability to real-world applications. Consider instance $\x_t \in \R^d$, the goal of dimension reduction is to learn a new instance $\hat \x_t \in \R^k$ where $k \le d$ by following some principle of unsupervised learning. There have been several approaches to unsupervised dimensional reduction. We broadly categorize them into two major groups of studies: subspace learning and manifold learning More comprehensive surveys of classic dimension reduction techniques can be found in \citep{burges2010dimension}.

\paragraph{Subspace Learning.} This class of DR methods aims to find an optimal linear mapping of input data in high-dimensional space to a lower-dimensional space. In general, there are two major types of approaches: linear methods and nonlinear methods. Popular linear subspace methods include Principal Component Analysis (PCA) and Independent Component Analysis (ICA), etc. Nonlinear methods often extend the linear subspace learning methods using kernel tricks. Examples include Kernel PCA, Kernel ICA, etc.

For online dimension reduction tasks, more popular efforts have been focused on addressing online PCA for unsupervised learning on streaming data settings in literature \citep{warmuth2008randomized,arora2013stochastic,Arora2012,mitliagkas2013memory,feng2013online}, while there are also a few studies for online ICA \citep{li2016online,wang2017scaling}.
For nonlinear space learning methods, online Kernel-PCA has also received some research interests \citep{kuzmin2007online,honeine2012online}.

\paragraph{Manifold learning.} This class of DR methods generally belongs to nonlinear DR techniques. Manifold learning assumes that input data lie on an embedded non-linear manifold within the high-dimensional space. DR by manifold learning aims to find a low-dimensional representation by preserving some properties of the manifold. For example, some methods preserving global properties include Multi-dimensional scaling (MDS), and IsoMap \citep{tenenbaum2000global}, while some preserving local properties including Locally Linear Embedding (LLE) \citep{roweis2000nonlinear} and Laplacian Eigenmaps.

For online manifold learning settings, the are some efforts for achieving the incremental approaches of manifold learning in literature. For example, \cite{law2006incremental} proposed an incremental learning algorithm for ISOMAP and \cite{schuon2008truly} presented an online approach for LLE.

\subsection{Online Density Estimation}

Online density estimation refers to constructing an estimate of an underlying unobservable probability density function based on observed data streams \citep{silverman2018density}. In literature, there are many different approaches to perform density estimation, e.g., histograms, naive estimator, nearest neighbour methods, Parzen windows, etc. Among various approaches, Kernel Density Estimation (KDE) is probably one of the most extensively explored topics in density estimation, which is a non-parametric way to estimate the probability density function of a target random variable \citep{scott2015multivariate}. Here, we briefly review and categorize some of the commonly used approaches for kernel density estimation in online-learning settings.
Given a sequence of instances $\mathcal{D} = \{\x_1, \dots, \x_T\} $, where $\x_t \in \R^d$, KDE estimates the density at a point $\x$ as
\begin{equation*}
\f(\x) = \frac{1}{T}\sum_{t=1}^T \kappa (\x, \x_t) = \frac{1}{Th} \sum_{t=1}^{T} \kappa \big(\frac{\x - \x_t}{h}\big)
\end{equation*}
where the kernel $\kappa(\x, \x_t)$ is a radially symmetric unimodal function that integrates to 1 and $h$ is a smoothing parameter called the bandwidth. Like in the case of Online Learning with Kernels \citep{kivinen2004online,Lu2015}, this problem suffers from the curse of kernelization, which means to estimate the density at any point $\x$, it requires computing the kernel function with respect to all the data points observed in the data stream so far.

There have been several attempts to overcome this curse of kernelization, and can be grouped into \textit{Merging} and \textit{Sampling} approaches. Merging approaches require a pre-specified budget on how many instances or kernels can be stored in memory. A newly arriving sample will (typically) be stored in memory as a kernel, unless the budget is exceeded. If the budget is exceeded, two or more similar kernels get merged. The merging criteria depends on some objective function. Some efforts in this direction include \citep{zhou2003m,boedihardjo2008framework,kristan2011multivariate}, which usually differ in how they select the bandwidth values.
Another approach in \citep{cao2012somke} performs clustering using self-organizing maps, and leverages this to perform kernel merging. Sampling approaches randomly select points to be kept in memory, but attempts to maintain a certain level of accuracy \citep{zheng2013quality}. Recent approaches \citep{qahtan2017kde} try to perform online density estimation with efficient methods for bandwidth selection and also to capture changes in the data distribution. Finally, online KDE techniques can be applied and integrated with real-world applications, such as real-time visual tracking \citep{han2008sequential}.

\subsection{Online Anomaly Detection}

Anomaly Detection (AD), also known as ``outlier detection" or ``novelty detection", is the process of detecting abnormal behavior in the data. The definition of abnormal behaviors can be very subjective, and the notion of ``anomaly" varies from domain to domain. Anomaly detection research is abundant in literature due to its wide applications. Example applications include but not limited to intrusion detection, fraud detection, medical anomaly detection, industrial damage detection, amongst others. Anomaly detection has been extensively studied by many communities in a wide range of diverse settings, ranging from supervised to unsupervised and semi-supervised learning, and batch learning to online learning settings. More comprehensive surveys of classic anomaly detection studies can be found in \citep{chandola2009anomaly,gupta2014outlier}. In this survey, we will focus online anomaly detection in unsupervised learning settings, which we believe it is one of the most popular and dominating scenarios in many real-world applications.

According to the literature surveys, unsupervised anomaly detection can be grouped into several major categories, including Distance based, Density based, Clustering based, Statistical methods, and others (such as subspace and one-class learning, etc). In the following, we briefly review some of popular work by focusing on online learning settings.

In literature, distance based online AD algorithms have been extensively studied in the context of unsupervised learning over data streams using distance-based methods \citep{angiulli2007detecting,yang2009neighbor,bu2009efficient,kloft2012security}. Some typical strategies of distance-based online AD approaches is to apply the sliding window
model where distance-based anomalies/outliers can be detected in the current window. In addition to distance-based methods, there are also some other studies that explore different methods for online AD in data streams, such as using one-class anomaly detector \citep{tan2011fast} or online clustering based approaches \citep{spinosa2009novelty}.
We note that, despite extensive and diverse studies in the field of online anomaly detection, from a machine learning perspective, many of theses approaches (e.g., sliding windows based) not purely learn in an online-learning fashion and many are not designed in machine learning based manners. We therefore keep the review of this part brief and concise.

\section{Related Areas and Other Terminologies}

\subsection{Overview}
In this section, we discuss the relationship of online learning with other related areas and terminologies which sometimes may be confused. We note that some of the following remarks may be somewhat subjective, and their meanings may vary in diverse contexts whereas some terms and notions may be used interchangeably.

\subsection{Incremental Learning}

\index{Incremental learning} Incremental learning, or decremental learning, represents a family of machine learning techniques \citep{michalski1986multi,poggio2001incremental,read2012batch}, which are particularly suitable for learning from data streams. There are various definitions of incremental learning/decremental learning. The basic idea of incremental learning is to learn some models from a stream of training instances with limited space and computational costs, often attempting to approximate a traditional batch machine learning counterpart as much as possible. For example, incremental SVM~\citep{poggio2001incremental} aims to train an SVM classifier the same as a batch SVM in an incremental manner where one training instance is added for updating the model each time (and similarly a training instance can be removed by updating the model decrementally).

Incremental learning can work either in online learning or batch learning manners~\citep{read2012batch}. For the incremental online learning~\citep{poggio2001incremental}, only one example is presented for updating the model at one time, while for the incremental batch learning~\citep{wang2003mining}, a batch of multiple training examples are used for updating the model each time. Incremental learning (or decremental learning) methods are often natural extensions of existing supervised learning or unsupervised learning techniques for addressing efficiency and scalability when dealing with real-world data particularly arriving in stream-based settings. Generally speaking, incremental learning can be viewed as a branch of online learning and extensions for adapting traditional offline learning counterparts in data-stream settings. 


\subsection{Sequential Learning}

\index{Sequential learning} Sequential learning is mainly concerned with learning from sequential training data~\citep{dietterich2002machine}, formulated as follows: a learner trains a model from a collection of $N$ training data pairs $\{(\mathbf{x}^{(i)},\mathbf{y}^{(i)}),i=1,\ldots,N\}$ where $\mathbf{x}^{(i)}=(x^{i}_1,x^{i}_2,\ldots,x^{i}_{N_i})$ is an $N_i$-dimensional instance vector and $\mathbf{y}^{(i)}=(y^{i}_1,y^{i}_2,\ldots,y^{i}_{N_i})$ is an $N_i$-dimensional label vector. It can be viewed as a special type of supervised learning, known as structured prediction or structured (output) learning~\citep{bakir2007predicting}, where the goal is to predict structured objects (e.g., sequence or graphs), rather than simple scalar discrete (``classification") or real values (``regression"). Unlike traditional supervised learning that often assume data is independently and identically distributed, sequential learning attempts to exploit significant sequential correlation of sequential data when training the predictive models. Some classical methods of sequential learning include sliding window methods, recurrent sliding windows, hidden Markov models, conditional random fields, and graph transformer networks, etc. There are also many recent studies for structured prediction with application to sequential learning~\citep{bakir2007predicting,roth2009sequential}. In general, sequential learning can be solved by either batch or online learning algorithms. Finally, it is worth mentioning another closely related learning, i.e., ``sequence classification", whose goal is to predict a single class output for a whole input ``sequence" instance. Sequence classification is a special case of sequential learning with the target class vector reduced to a single variable. It is generally simpler than regular sequential learning, and can be solved by either batch or online learning algorithms.


\subsection{Stochastic Learning}

\index{Stochastic Learning} Stochastic learning refers to a family of machine learning algorithms by following the theory and principles of stochastic optimization~\citep{bottou2004stochastic,zhang2004solving,bottou2010large}, which have achieved great successes for solving large-scale machine learning tasks in practice~\citep{bousquet2008tradeoffs}. Stochastic learning is closely related to online learning. Typically, stochastic learning algorithms are motivated to accelerate the training speed of some existing batch machine learning methods for large-scale machine learning tasks, which may be often solved by batch gradient descent algorithms. Stochastic learning algorithms, e.g., Stochastic Gradient Descent (SGD) or a.k.a Online Gradient Descent (OGD) in online learning terminology, often operate sequentially by processing one training instance (randomly chosen) each time in an online learning manner, which thus are computationally more efficient and scalable than the batch GD algorithms for large-scale applications. Rather than processing a single training instance each time, a more commonly used stochastic learning technique in practice is the mini-batch SGD algorithm~\citep{bousquet2008tradeoffs,shalev2011pegasos}, which processes a small batch of training instances each time. Thus, stochastic learning can be viewed as a special family of online learning algorithms and extensions, while online learning may explore more other topics and challenges beyond stochastic learning/optimizations.

\subsection{Adaptive Learning}

\index{Adaptive learning} This term is occasionally used in the machine learning and neural networks fields. There is no a very formal definition about what exactly is adaptive learning in literature. In literature, there are quite a lot of different studies more or less concerned with adaptive learning~\citep{carpenter1991artmap,carpenter1992fuzzy}, which attempt to adapt a learning model/system (e.g., neural networks) for dynamically changing environments over time. In general, these existing works are similar to online learning in that the environment is often changing and evolving dynamically. But they are different in that they are not necessarily purely based on online learning theory and algorithms. Some of these works are based on heuristic adaptation/modification of existing batch learning algorithms for updating the models with respect to the environment changes. Last but not least, most of these existing works are motivated by different kinds of heuristics, generally lack solid theoretical analysis and thus can seldom give performance guarantee in theory.


\subsection{Interactive Learning}

\index{Interactive learning} Traditional machine learning mostly works in a fully automated process where training data are collected and prepared typically with the aid of domain experts. By contrast, interactive (machine) learning aims to make the machine learning procedure interactive by engaging human (users or domain experts) in the loop~\citep{ware2001interactive,johnson2003interactive}. The advantages of interactive learning include the natural integration of domain knowledge in the learning process, effective communication and continuous improvements for learning efficacy through the interaction between learning systems and users/experts. Online learning often plays an important role in an interactive learning system, in which active (online) learning can be used in finding the most informative instances to save labeling costs, incremental (online) learning algorithms could be applied for updating the models sequentially, and/or bandit online learning algorithms may be used for decision-making to trade off exploration and exploitation in some scenarios.


\subsection{Reinforcement Learning}

\index{Reinforcement learning} Reinforcement Learning (RL)~\citep{kaelbling1996reinforcement,barto1998reinforcement} is a branch of machine learning inspired by behaviorist psychology, which is often concerned with how software agents should take actions in an environment to maximize cumulative rewards. The goal of an agent in RL is to find a good policy and state-update function by attempting to maximize the the expected sum of discounted rewards. RL is different from supervised learning~\citep{dietterich2004} in that the goal of supervised learning is to reconstruct an unknown function $f$ that can assign the desired output values $y$ to input data $x$; while the goal of RL is to find the input (policy/action) $x$ that gives the maximum reward $R(x)$.
In general, RL can work either in batch or online learning manners. In practice, RL methods are commonly applied to problems involving sequential dynamics and optimization of some objectives, typically with online exploration of the effects of actions. RL is closely related to bandit online learning with the similar goal of finding a good policy that has to balance the tradeoff between exploration (of uncertainty) and exploitation (of known knowledge). Many RL solutions follow the same ideas of multi-armed bandits, and some bandit algorithms were also inspired by the field of RL studies too. However, RL can be more general when learning to interact with more complex scenarios and environments.



\subsection{Continual Learning}

\textit{Continual Learning}, also called ``Lifelong Learning" \citep{ruvolo2013ella,silver2013lifelong,parisi2018continual} is a field of machine learning inspired by human ability to learn new tasks throughout their lifespan. When new tasks arrive, humans are able to leverage existing knowledge, and more effectively learn the new tasks, and at the same time, they do not forget how to perform the old tasks. In formal settings, the tasks arrive sequentially, but instances for each task arrive as a batch, and thus each task is still learned in batch settings. While older methods used linear models for lifelong learning \citep{ruvolo2013ella}, recent efforts have been focused on continual learning with neural networks \citep{parisi2018continual}, in which one of key challenges is to address the {\it catastrophic forgetting}, a problem which traditional machine learning including neural networks is often susceptible to, but humans are immune to. When new tasks are learned, traditional machine learning tends to forget how to perform older tasks, and a major research direction in continual learning is to develop algorithms that can address this catastrophic forgetting. Although continual learning is closely related to online learning, most existing studies still follow the paradigm of batch training, which are not considered as online learning algorithms.

\section{Conclusions}

\subsection{Concluding Remarks}

This paper gave a comprehensive survey of existing online learning works and reviewed ongoing trends of online learning research. In theory, online learning methodologies are founded primarily based on learning theory, optimization theory, and game theory. According to the type of feedback to the learner, the existing online learning methods can be roughly grouped into the following three major categories:
\begin{itemize}
\item{\bf Supervised online learning} is concerned with the online learning tasks where full feedback information is always revealed to the learner, which can be further divided into three groups: (i) ```linear online learning" that aims to learn a linear predictive model, (ii) ``nonlinear online learning" that aims to learn a nonlinear predictive model, and (iii) non-traditional online learning that addresses a variety of supervised online learning tasks which are different from traditional supervised prediction models for classification and regression.
\item{\bf Online learning with limited feedback} is concerned with the online learning tasks where the online learner receives partial feedback information from the environment during the learning process. The learner often has to make online predictions or decisions by achieving a tradeoff between the exploitation of disclosed knowledge and the exploration of unknown information.
\item{\bf Unsupervised online learning} is concerned with the online learning tasks where the online learner only receives the sequence of data instances without any additional feedback (e.g., true class label) during the online learning tasks. Examples of unsupervised online learning include online clustering, online representation learning, and online anomaly detection tasks, etc.
\end{itemize}
In this survey, we have focused more on the first category of work because it has received more research attention than the other two categories in literature. This is mainly because supervised online learning is a natural extension of traditional batch supervised learning, and thus an online supervised learning technique could be directly applied to a wide range of real-world applications especially for real-time machine learning from data streams where conventional batch supervised learning techniques may suffer critical limitations. However, we do note that in contrast to supervised online learning, the problems of online learning with limited feedback or unsupervised online learning are generally much more challenging, and thus should attract more research attentions and efforts in the future.

\subsection{Future Directions}
Despite the extensive studies in literature, there are still many open issues and challenges, which have not been fully solved by the existing works and need to be further explored by the community efforts in the future work. In the following, we highlight a few important and emerging research directions for researchers who are interested in online learning.


First of all, although supervised online learning has been extensively studied, learning from non-stationary data streams remain an open challenge. In particular, one critical challenge with supervised online learning is to address ``concept drifting" where the target concepts to be predicted may change over time in unforeseeable ways. Although many online learning studies have attempted to address concept drifting by a variety of approaches, they are fairly limited in that they often make some restricted assumptions for addressing certain types of concept drifting patterns. In general, there still lacks of formal theoretical frameworks or principled ways for resolving all types of concept drifting issues, particularly for non-stationary settings where target concepts may drift over time in arbitrary ways.

Second, an important growing trend of online learning research is to explore large-scale online learning for real-time big data analytics. Although online learning has huge advantages over batch learning in efficiency and scalability, it remains a non-trivial task when dealing with real-world big data analytics with extremely high volume and high velocity. Despite extensive research in large-scale batch machine learning, more future research efforts should address parallel online learning and distributed online learning by exploiting various computational resources, such as high-performance computing machines, cloud computing infrastructures, and perhaps low-cost IoT computing environments.

Third, another challenge of online learning is to address the ``variety" in online data analytics tasks. Most existing online learning studies are often focused on handling single-source structured data typically by vector space representations. In many real-world data analytis applications, data may come from multiple diverse sources and could contain different types of data (such as structured, semi-structured, and unstructured data). Some existing studies, such as the series of online multiple kernel learning works, have attempted to address some of these issues, but certainly have not yet fully resolved all the challenges of variety. In the future, more research efforts should address the ``variety" challenges, such as multi-source online learning, multi-modal online learning, etc.


Fourth, existing online learning works seldom address the data ``veracity" issue, that is, the quality of data, which can considerably affect the efficacy of online learning. Conventional online learning studies often implicitly assume data and feedback are given in perfect quality, which is not always true for many real-world applications, particularly for real-time data analytics tasks where data arriving on-the-fly may be contaminated with noise or may have missing values or incomplete data without applying advanced pre-processing. More future research efforts should address the data veracity issue by improving the robustness of online learning algorithms particularly when dealing with real data of poor quality.

Fifth, due to the remarkable successes and impact of deep learning techniques for various applications in recent years, another emerging and increasingly important topic is ``online deep learning" \citep{sahoo2017online}, i.e., learning deep neural networks from data streams on the fly in an online fashion. Despite some preliminary research, we note there are still many research challenges in this field, e.g., how to balance the tradeoff between learning accuracy, computational efficiency, learning scalability and model complexity.

Last but not least, we believe it can be valuable to explore ``online continual learning" by extending traditional continual learning methods for pure online-learning settings, which is more natural for many real-world applications where data for either existing and novel tasks are often arriving in a streaming and continuous fashion. Some research progress in online deep learning might be applied here, but the key challenge of continual online learning is to resolve the catastrophic forgetting problem across tasks during the online learning process.



\bibliography{references}

\end{document}